\documentclass[journal]{IEEEtran}
\PassOptionsToPackage{numbers,sort&compress}{natbib}
\usepackage{multirow}
\usepackage{booktabs}
\usepackage{graphicx}
\usepackage[normalem]{ulem}
\useunder{\uline}{\ul}{}

\usepackage{amsmath,amsfonts}
\usepackage{algorithm}
\usepackage{array}
\usepackage[caption=false,font=scriptsize,labelfont=rm,textfont=rm]{subfig}
\usepackage{textcomp}
\usepackage{stfloats}
\usepackage{url}
\usepackage{verbatim}
\usepackage{graphicx}
\usepackage{cite}
\usepackage{amssymb}
\usepackage{xcolor}
\usepackage{pifont}
\usepackage{algpseudocode}
\usepackage{bbding}
\usepackage{colortbl}
\usepackage{makecell}
\usepackage{caption}
\usepackage{cuted}
\definecolor{cvprblue}{rgb}{0.21,0.49,0.74}
\usepackage[pagebackref,breaklinks,colorlinks,allcolors=cvprblue]{hyperref}
\begin{document}

\def\xc{\textcolor{red}}

\def\hx{\textcolor{blue}}

\def\etal{\textit{et al.}}
\def\etc{\textit{etc}}
\def\ie{\textit{i.e.}}
\def\eg{\textit{e.g.}}
\definecolor{mygray}{gray}{.9}

\title{{Beyond Inference Intervention: Identity-Decoupled Diffusion for Face Anonymization}

\author{Haoxin~Yang,
        Yihong~Lin,
        Jingdan~Kang, 
        Xuemiao~Xu,~\IEEEmembership{Member,~IEEE,}\\
        Yue~Li,
        Cheng~Xu,
        Shengfeng He,~\IEEEmembership{Senior Member,~IEEE}
        }

\thanks{Haoxin~Yang, Yihong~Lin, Xuemiao~Xu and Yue~Li are with the School of Computer Science and Engineering, South China University of Technology, Guangzhou, China. Xuemiao Xu is also with the State Key Laboratory of Subtropical Building Science, Ministry of Education Key Laboratory of Big Data and Intelligent Robot, and Guangdong Provincial Key Lab of Computational Intelligence and Cyberspace Information, Guangzhou 510640, China. E-mail: harxis@outlook.com; amcsyihonglin@foxmail.com; xuemx@scut.edu.cn; liyue@scut.edu.cn.}
\thanks{Jingdan~Kang is with the School of Future Technology, South China University of Technology, Guangzhou, China. E-mail: jingdankang6@gmail.com.}
\thanks{Cheng Xu and Shengfeng He are with the School of Computing and Information Systems, Singapore Management University, Singapore. Email: cschengxu@gmail.com; shengfenghe@smu.edu.sg.}}

\markboth{Preprint}%
{Yang \MakeLowercase{\textit{et al.}}: Beyond Inference Intervention: Identity-Decoupled Diffusion for Face Anonymization}


\maketitle

\begin{abstract}

   Face anonymization aims to conceal identity information while preserving non-identity attributes. Mainstream diffusion models rely on inference-time interventions such as negative guidance or energy-based optimization, which are applied post-training to suppress identity features. These interventions often introduce distribution shifts and entangle identity with non-identity attributes, degrading visual fidelity and data utility. To address this, we propose \textbf{ID\textsuperscript{2}Face}, a training-centric anonymization framework that removes the need for inference-time optimization. The rationale of our method is to learn a structured latent space where identity and non-identity information are explicitly disentangled, enabling direct and controllable anonymization at inference. To this end, we design a conditional diffusion model with an identity-masked learning scheme. An Identity-Decoupled Latent Recomposer uses an Identity Variational Autoencoder to model identity features, while non-identity attributes are extracted from same-identity pairs and aligned through bidirectional latent alignment. An Identity-Guided Latent Harmonizer then fuses these representations via soft-gating conditioned on noisy feature prediction. The model is trained with a recomposition-based reconstruction loss to enforce disentanglement. At inference, anonymization is achieved by sampling a random identity vector from the learned identity space. To further suppress identity leakage, we introduce an Orthogonal Identity Mapping strategy that enforces orthogonality between sampled and source identity vectors. Experiments demonstrate that ID\textsuperscript{2}Face outperforms existing methods in visual quality, identity suppression, and utility preservation.


\end{abstract} 
\section{Introduction}

The rapid growth of visual data across digital platforms has raised serious concerns over biometric privacy. Facial imagery inherently encodes persistent and traceable identity information, and is continuously captured and shared across social media and surveillance systems. Its potential misuse creates substantial privacy risks, making reliable face anonymization a critical research challenge. Face anonymization seeks to conceal identity information while preserving non-identity attributes such as expression, pose, and background~\cite{neustaedter2006blur, vishwamitra2017blur,hukkelaas2019deepprivacy, maximov2020ciagan, kuang2021effective,RiDDLE,yang2024g, AIDPro,barattin2023attribute}. It protects privacy without compromising downstream data utility, enabling applications in privacy-preserving face recognition~\cite{laishram2025toward}, video surveillance~\cite{proencca2021uu,ye2024securereid}, and secure content sharing~\cite{ciftci2023my}.

\begin{figure}[t!]
    \centering
    \includegraphics[width=\linewidth]{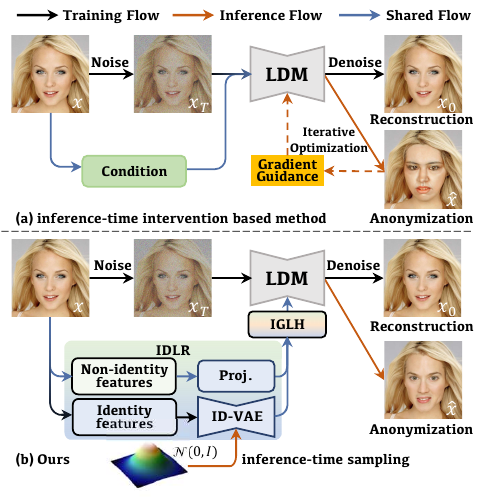}
    \caption{(a) Existing methods rely on inference-time intervention to erase identity, often resulting in suboptimal anonymization and distortion of non-identity features. (b) ID\textsuperscript{2}Face introduces an inference-time-intervention-free framework that disentangles and harmonizes identity and non-identity features, achieving superior anonymization while preserving identity-irrelevant attributes.
    }
    \label{fig:teaser}
 \end{figure}


Classical anonymization techniques such as blurring, masking, and pixelation effectively obscure identity~\cite{neustaedter2006blur, vishwamitra2017blur}, but severely degrade visual quality and utility. To mitigate these limitations, GAN-based methods synthesize anonymized faces that preserve non-identity features~\cite{hukkelaas2019deepprivacy, maximov2020ciagan, kuang2021effective,yang2024g,RiDDLE,AIDPro,barattin2023attribute}. However, GANs remain constrained by mode collapse, unstable training, and limited visual fidelity. Diffusion models have recently emerged as powerful generative learners~\cite{DDPM,DDIM,ldm}, enabling impressive anonymization performance~\cite{DiffPrivacy,FAMS,NullFace} through accurate facial synthesis.
\looseness=-1

Notwithstanding the demonstrated success, current diffusion-based approaches primarily obscure identity via inference-time interventions, as illustrated in Fig.~\ref{fig:teaser}(a). Typical strategies include negative guidance~\cite{FAMS,NullFace} and energy-based optimization~\cite{DiffPrivacy}, which externally influence the sampling trajectory to suppress identity cues. These interventions introduce two key limitations. First, post-hoc optimization alters the sampling distribution, resulting in visual artifacts and degraded anonymization performance. Second, as identity and non-identity attributes remain entangled in the latent space learned during training, forcing identity manipulation at inference often distorts non-identity features, diminishing downstream utility. This challenge motivates a fundamental question: how can we achieve high-fidelity anonymization without relying on inference-time intervention and without compromising non-identity information?
\looseness=-1

One promising direction is to incorporate anonymization objectives directly into diffusion model training. However, diffusion reconstruction inherently promotes identity preservation, which conflicts with identity removal and prevents naive end-to-end learning of anonymization. Overcoming this optimization conflict requires a principled strategy that separates identity from other facial attributes within the diffusion process.
\looseness=-1

To address the aforementioned challenges, we propose to construct an identity-decoupled diffusion space that enables selective manipulation of identity attributes while preserving non-identity consistency. This disentangled representation supports effective and flexible anonymization at inference by modifying only identity-related latent codes, without altering utility-relevant features. To this end, we introduce \emph{ID\textsuperscript{2}Face}, an inference-intervention-free framework for face anonymization, illustrated in Fig.~\ref{fig:teaser}(b). Built on a conditional denoising diffusion paradigm, ID\textsuperscript{2}Face incorporates an identity-masked learning scheme that encourages the model to internalize identity and non-identity information into separable latent subspaces. A key novelty of ID\textsuperscript{2}Face lies in its \emph{recomposition-driven disentanglement}: instead of learning to suppress identity cues through global objectives or post-hoc guidance, the model is trained to explicitly factor identity and non-identity features from paired inputs and reconstruct a coherent image from their controlled fusion. This structured approach is implemented through two main components:

\emph{(i) Identity-Decoupled Latent Recomposer (IDLR).} Given two facial images of the same identity, IDLR isolates identity features using an Identity Variational Autoencoder (ID-VAE), while extracting non-identity cues from variations across the pair. A bidirectional alignment mechanism ensures semantic and structural consistency between the two feature streams, promoting a well-separated latent representation.

\emph{(ii) Identity-Guided Latent Harmonizer (IGLH).} IGLH adaptively integrates the disentangled features through a region-aware, scale-sensitive gating mechanism conditioned on noisy latent predictions. This enables fine-grained control over identity content while preserving local appearance and global structure in the generated output.

To further reduce identity leakage, we introduce an \emph{Orthogonal Identity Mapping (OIM)} strategy at inference, which enforces orthogonality between the sampled identity vector and the source identity representation. By explicitly disentangling and recomposing identity and non-identity attributes during training, ID\textsuperscript{2}Face enables efficient anonymization at inference through simple identity sampling, with no optimization or external intervention required. Extensive experiments show that our method achieves state-of-the-art performance in identity suppression while preserving high visual fidelity and downstream utility.

In summary, our contributions are fourfold:
\begin{itemize}
    \item We resolve the conflict between reconstruction and anonymization in diffusion-based face anonymization by reformulating the task as a unified reconstruction problem within an identity-decoupled diffusion framework. To the best of our knowledge, this is the first diffusion-based approach that eliminates inference-time optimization, enabling precise and high-fidelity identity obfuscation.
    \item We introduce an identity-masked diffusion learning paradigm that explicitly disentangles identity and non-identity representations through recomposition-based reconstruction, enabling accurate and controllable identity manipulation at inference.
    \item We design an Orthogonal Identity Mapping strategy that enforces latent orthogonality between the source and anonymized identities, maximizing anonymization effectiveness while preserving image quality.
    \item Extensive experiments demonstrate that our method achieves state-of-the-art face anonymization performance, producing high-fidelity outputs and preserving non-identity attributes critical for downstream utility.
\end{itemize}

\section{Related Work}

\subsection{Diffusion Models}
Diffusion models have recently emerged as a powerful class of generative models, known for their ability to synthesize high-quality images through iterative denoising. Foundational works such as denoising diffusion probabilistic models (DDPM)~\cite{DDPM}, denoising diffusion implicit models (DDIM)~\cite{DDIM}, and latent diffusion models (LDM)~\cite{ldm} have demonstrated that diffusion models can outperform GAN-based methods~\cite{goodfellow2020generative}, especially in complex image generation tasks. Unlike adversarial training, diffusion models avoid issues such as mode collapse and training instability, making them more robust and scalable. These advances have enabled applications across diverse domains, including text-to-image generation~\cite{mou2024t2i, yu2024beyond, kang2025sita}, portrait synthesis~\cite{gal2022image, ruiz2023dreambooth, peng2024portraitbooth, ye2023ip}, and image editing~\cite{kawar2023imagic, huang2025diffusion}.
Building on these advances, we explore diffusion models for face anonymization, where the goal is not only to generate realistic faces but also to ensure identity obfuscation and utility preservation. Our work extends existing diffusion frameworks by introducing mechanisms for learning identity-disentangled representations, tailored specifically for anonymization.

\subsection{GAN-based Face Anonymization}
Earlier work in face anonymization primarily relied on Generative Adversarial Networks (GANs)~\cite{goodfellow2020generative}, which can be broadly categorized into two types. The first trains conditional GANs from scratch to synthesize anonymized faces by modifying identity attributes~\cite{hukkelaas2019deepprivacy, maximov2020ciagan, wen2023divide, yuan2022pro, yuan2024pro, AIDPro}. While flexible, these models often suffer from limited visual quality due to unstable training. The second type builds on pre-trained StyleGAN models~\cite{karras2019style}, modifying latent codes~\cite{RiDDLE, barattin2023attribute} or applying conditional editing~\cite{yang2024g} to obscure identity while leveraging high-quality priors. However, all GAN-based approaches are constrained by fundamental issues such as mode collapse and adversarial instability, leading to inconsistent anonymization quality.
In contrast, our method avoids unstable adversarial training and leverages the powerful generative capability of diffusion models to achieve more stable and consistent anonymization, with a design that supports explicit control over identity and non-identity information.

\subsection{Diffusion-based Face Anonymization}
With the success of diffusion models in image generation, several methods have recently adapted them for face anonymization~\cite{DiffPrivacy, FAMS, NullFace}. DiffPrivacy~\cite{DiffPrivacy} introduces identity suppression through inference-time energy optimization. FAMS~\cite{FAMS} conditions a U-Net on identity features and modifies internal representations to alter facial identity. NullFace~\cite{NullFace} proposes a training-free method that dynamically adjusts guidance weights during inference. While effective in certain settings, these approaches share a common reliance on inference-time intervention, which introduces distribution shifts and lacks explicit identity disentanglement. As a result, they often produce artifacts or inadvertently degrade non-identity features, limiting their utility.
The proposed ID\textsuperscript{2}Face addresses these limitations by integrating identity control directly into the training process. Rather than relying on post-hoc intervention, our method is designed to learn an identity-disentangled representation space that supports targeted identity manipulation while preserving non-identity features. This training-centric design improves generation fidelity and anonymization consistency, offering a principled alternative to intervention-based approaches.

\section{Proposed Method}

\subsection{Problem Formulation}

Face anonymization aims to conceal identity-specific facial cues while faithfully preserving identity-irrelevant attributes such as hairstyle, facial expression, and background, thereby ensuring high utility of the anonymized data for downstream applications. In this work, we adopt the LDMs~\cite{ldm} as our backbone due to its powerful generative capacity.
Let $\mathcal{M}$ denote an LDM consisting of three components:
\begin{itemize}
    \item {Encoder} ($E_{\rm diff}$): a diffusion VAE encoder that maps the input image from pixel space to latent space;
    \item {Denoiser} ($\epsilon_\theta$): a noise prediction network that estimates the injected Gaussian noise during diffusion;
    \item {Decoder} ($D_{\rm diff}$): a diffusion VAE decoder that reconstructs the image from its latent representation.
\end{itemize}

Given an input face image $x$ with identity embedding $e^{\text{x}}_{\text{id}}$, we aim to synthesize an anonymized image $\hat x$ such that:
\begin{equation}
    \hat x = \mathcal{M}(x, e^{\text{ctrl}}_{\text{id}}), 
    \quad \text{s.t.} \  \text{id}(\hat x) = e^{\text{ctrl}}_{\text{id}}, \  \text{id}(\hat x) \neq e^{\text{x}}_{\text{id}}.
\end{equation}
Here, $e^{\text{ctrl}}_{\text{id}}$ is a randomly sampled identity embedding. The generated image $\hat x$ must preserve all non-identity aspects of $x$ while ensuring that the original identity information is effectively suppressed.

\subsection{Overview}

To construct a unified, inference-intervention-free diffusion framework for face anonymization, we propose {ID\textsuperscript{2}Face, a conditional latent diffusion model that explicitly disentangles identity and non-identity attributes during training. By internalizing anonymization into the learning process, the model eliminates the need for post-hoc identity suppression at inference, avoiding distributional shifts and attribute entanglement.

As illustrated in Fig.~\ref{fig:framework}, the architecture comprises two primary components trained under an \emph{identity-masked diffusion learning} framework:
(i) Identity-Decoupled Latent Recomposer (IDLR). The IDLR disentangles identity-related and non-identity representations from input facial images by leveraging paired samples of the same identity. Under the identity-masked learning strategy, an Identity Variational Autoencoder (ID-VAE) is used to encode identity vectors from the input image. These vectors are sampled from a learned identity prior during inference to enable anonymization. Meanwhile, non-identity features are extracted from intra-identity variations and aligned using a bidirectional latent alignment mechanism to ensure semantic and structural consistency. The disentangled identity and non-identity representations are transformed into conditional control signals that guide the diffusion process, promoting controllable identity manipulation and high-fidelity generation.
(ii) Identity-Guided Latent Harmonizer (IGLH). To enhance identity control and improve fusion quality, the IGLH extends the UNet’s standard attention layers into dual-branch conditional attention blocks equipped with a learnable gating mechanism. This design enables region-aware, scale-sensitive modulation between identity-relevant and identity-irrelevant features, facilitating fine-grained spatial control and coherent visual synthesis.
In addition, we introduce an Orthogonal Identity Mapping (OIM) strategy to further suppress identity leakage. During inference, anonymized vectors are sampled from the learned identity space and constrained to be orthogonal to the source image’s identity representation. This encourages latent disentanglement and maximizes privacy preservation without degrading visual quality.

\begin{figure*}
    \centering
    \includegraphics[width=1.0\textwidth]{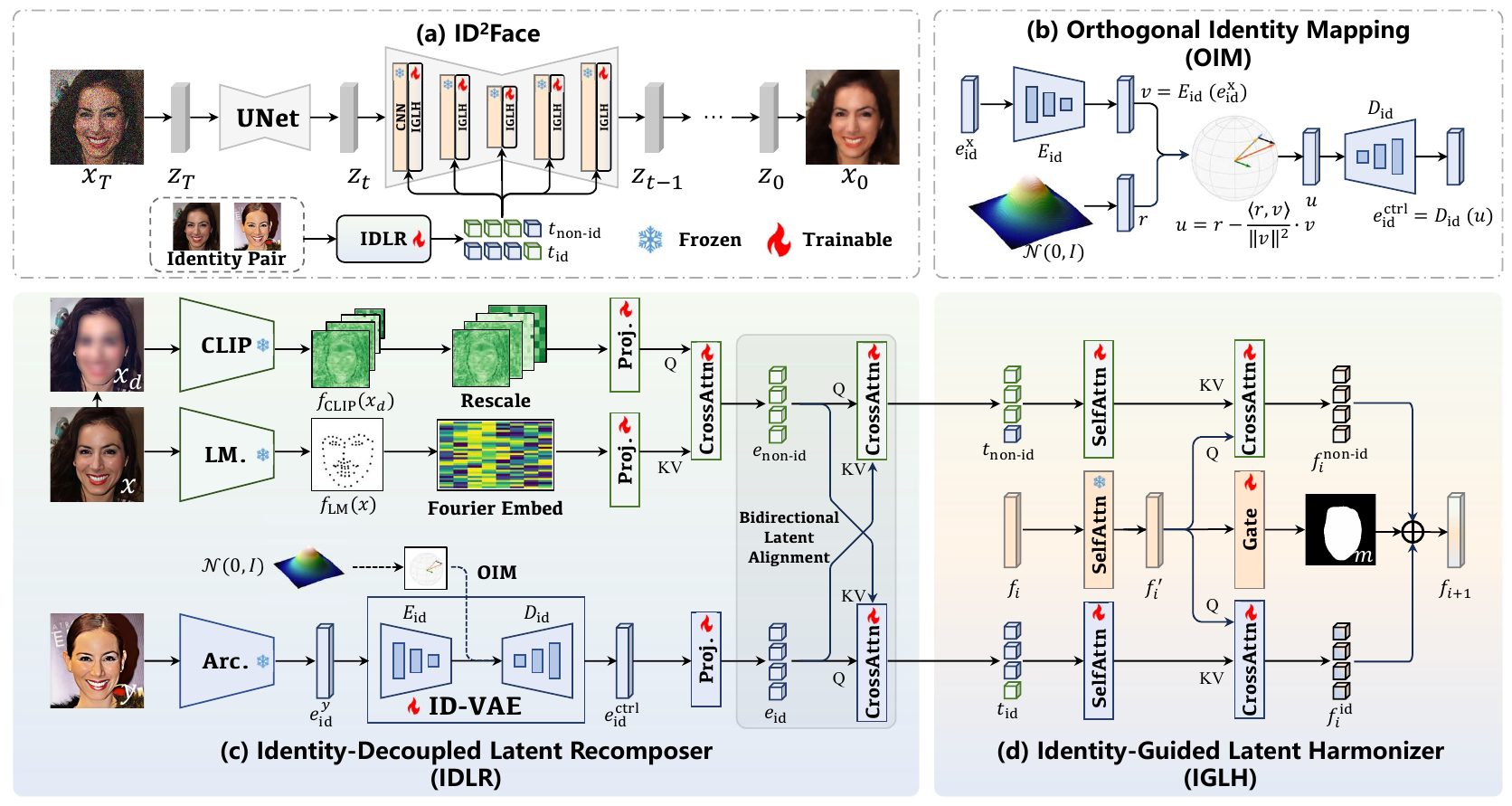}
    \caption{Overview of the proposed ID\textsuperscript{2}Face framework. The model learns an identity-decoupled latent space via identity-masked diffusion training, enabling anonymization without inference-time intervention. The Identity-Decoupled Latent Recomposer (IDLR) extracts identity vectors using an ID-VAE and recomposes them with non-identity cues from paired inputs with bidirectional alignment. The Identity-Guided Latent Harmonizer (IGLH) fuses these recomposed features via identity-guided soft-gating for fine-grained, spatially-aware control. At inference, a random identity vector is sampled from the learned space and constrained via Orthogonal Identity Mapping (OIM) to suppress identity leakage and maximize anonymization effectiveness.
    }
    \label{fig:framework}
\end{figure*}

\subsection{Identity-Masked Diffusion Learning Framework}
The proposed identity-masked diffusion learning framework explicitly disentangles identity and non-identity representations through a recomposition-based reconstruction in an end-to-end manner. 
This design enables accurate and controllable identity learning during training, while allowing effective identity manipulation and removal during inference. 
The framework comprises two key modules: the \emph{Identity-Decoupled Latent Recomposer} and the \emph{Identity-Guided Latent Harmonizer}.

\subsubsection{Identity-Decoupled Latent Recomposer}
During face anonymization, the objective is to eliminate identity-related information while preserving identity-independent visual attributes. To this end, we propose the IDLR, which is specifically designed to achieve identity removal with structural and appearance consistency. 
The IDLR operates in three stages. First, it explicitly disentangles non-identity features from identity features, ensuring that the latent representation is purged of identifiable cues. Second, it models the controllable identity information via an {ID-VAE}, which learns a compact and disentangled identity embedding. Finally, the non-identity and identity representations, extracted from different source images, are semantically aligned and adaptively recomposed using a {bidirectional cross-attention block}. This fusion produces a coherent conditional representation that effectively guides the subsequent diffusion-based image generation process, enabling identity-anonymized synthesis with preserved structural and perceptual realism.

\textbf{Identity-Masked Latent Decoupling.}
In the task of face anonymization, the goal is to remove identity-specific information from a given facial image $x$ while faithfully preserving identity-irrelevant attributes. 
To maintain high fidelity in these attributes, the original image $x$ is typically used as a conditioning signal. 
However, directly conditioning on $x$ introduces a critical limitation: without proper preprocessing, the model may exploit identity cues present in the full image, leading to identity leakage.

To address this issue, we propose an \emph{identity-masked learning} scheme that explicitly disentangles non-identity features from the input image. 
Specifically, we employ a facial parsing network~\cite{zheng2022general} to isolate the facial region from $x$, followed by a stochastic degradation process~\cite{wang2021towards} to obtain a partially corrupted version $x_d$. 
Unlike full occlusion, this degradation preserves semantically meaningful yet identity-agnostic details, effectively suppressing identity information while retaining contextual integrity.
To enrich the representation of non-identity semantics, we extract multi-scale spatial features from $x_d$ using a pretrained CLIP model~\cite{radford2021learning} implemented with a convolutional backbone~\cite{liu2022convnet}. 
These features, denoted as $f_{\text{CLIP}}(x_d)$, capture high-level contextual cues that promote faithful reconstruction of identity-irrelevant content during diffusion.

Furthermore, to preserve dynamic facial attributes such as expression and geometry, we extract facial landmarks from the original image $x$ and encode them via a Fourier embedding module~\cite{li2023gligen}, yielding geometric features $f_{\text{LM}}(x)$. 
The semantic and geometric features are subsequently fused through a cross-attention mechanism to construct an identity-agnostic spatial embedding $e_{\text{non-id}}$, formulated as:
\begin{equation}
    \begin{aligned}
        e_{\text{non-id}} = \text{CrossAttn}\Bigl(
        & q = \text{Proj}\bigl(f_{\text{CLIP}}(x_d)\bigr), \\
        & k = v = \text{Proj}\bigl(f_{\text{LM}}(x)\bigr)
        \Bigr),
    \end{aligned}
\end{equation}
where $\text{Proj}(\cdot)$ denotes a projection function implemented as a multilayer perceptron (MLP). 
The resulting spatial embedding $e_{\text{non-id}}$ serves as an identity-agnostic conditioning signal for the diffusion model, guiding it to generate anonymized facial images that retain structural, geometric, and appearance consistency while effectively suppressing identity information.

\textbf{ID-VAE for Identity Control.}
To construct a disentangled identity space during training and enable random identity sampling at inference, we employ an ID-VAE (Identity Variational Autoencoder) to explicitly model and separate identity information prior to diffusion. 
During training, the ID-VAE learns the distribution of identity representations by encoding and reconstructing identity control vectors derived from the input images. 
This joint learning process equips the system with two complementary capabilities: the ID-VAE generates controllable identity embeddings, while the diffusion model learns to condition on these embeddings for effective identity manipulation.
At inference, identity vectors are randomly sampled from the learned latent distribution and decoded into identity embeddings, which serve as conditional inputs to the diffusion model. 
Because the diffusion model is trained under explicit identity control, it can consistently synthesize controllable outputs that preserve non-identity attributes and ensure robust identity obfuscation by randomly sampling identity during the inference stage.

To further enhance identity controllability, we deivse a \emph{pairwise identity-guided training} strategy. Specifically, for each training instance, we construct an image pair $(x, y)$ from the same subject. An identity embedding $e^{y}_{\text{id}}$ is extracted from $y$ using a pretrained face recognition model~\cite{deng2019arcface}. This embedding is encoded and decoded via the ID-VAE to produce a controllable identity vector $e^{\text{ctrl}}_{\text{id}}$, which is used to guide the synthesis of $\hat x$. The resulting vector is projected into a format suitable for conditioning the diffusion model, yielding the identity embedding $e_{\text{id}}$:
\begin{equation}
e_{\text{id}} = \text{Proj}(e^{\text{ctrl}}_{\text{id}}), \quad
e^{\text{ctrl}}_{\text{id}} = D_{\text{id}}\bigl(E_{\text{id}}(e^{y}_{\text{id}})\bigr),
\end{equation}
where $E_{\text{id}}$ and $D_{\text{id}}$ denote the ID-VAE encoder and decoder, respectively.
This pairwise supervision allows our ID\textsuperscript{2}Face to learn more disentangled and precise control over facial identity. By explicitly linking identity embeddings to consistent semantic sources, the model generalizes more effectively across diverse visual conditions, while maintaining high-quality anonymization performance.

\textbf{Identity-Decoupled Latent Recomposing.}
Simply concatenating $e_{\text{non-id}}$ and $e_{\text{id}}$ and feeding them directly into the UNet’s cross-attention block as conditioning inputs leads to suboptimal results, where the generated images fail to preserve identity-independent low-level details, as illustrated in Fig.~\ref{fig:compose}(b). This degradation arises because the semantic distributions of $e_{\text{non-id}}$ and $e_{\text{id}}$ differ substantially, making it difficult for a single concatenation operation and a single cross-attention layer to effectively fuse such heterogeneous information. 

\begin{figure}[t]
    \centering
    \subfloat[Input]{
        \begin{minipage}{0.3\linewidth}
            \includegraphics[height=\linewidth]{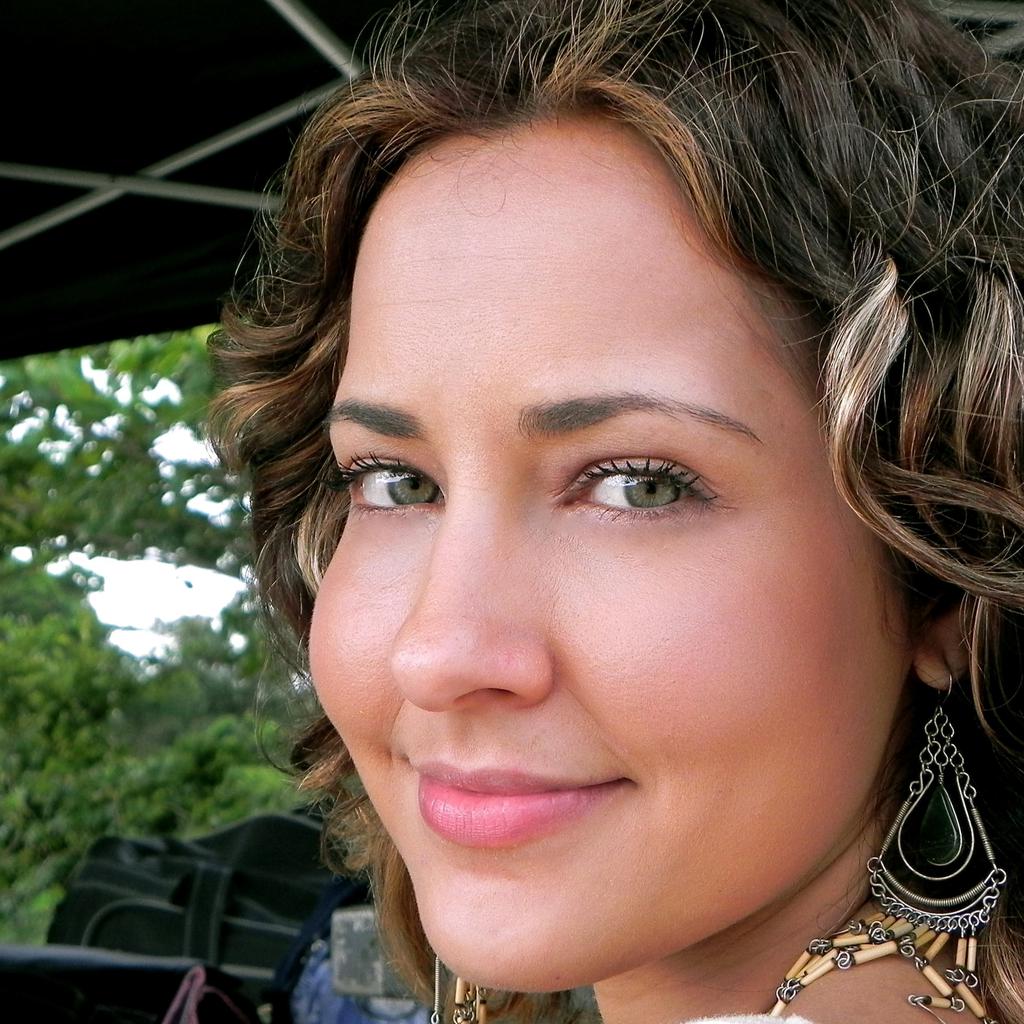} \\ \vspace{-9pt}
        \end{minipage}
    }
    \hspace{-8pt}
    \subfloat[Concat.]{
        \begin{minipage}{0.3\linewidth}
            \includegraphics[height=\linewidth]{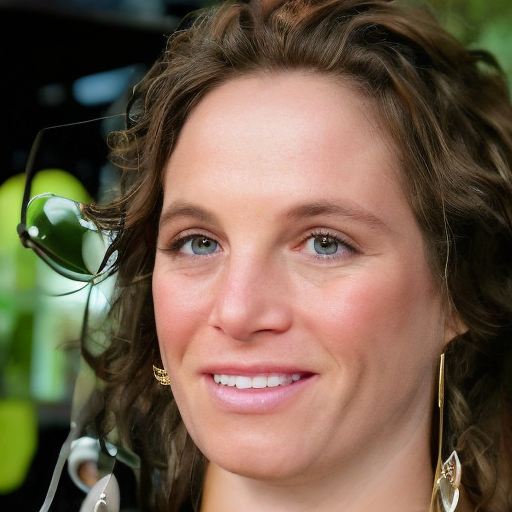} \\ \vspace{-9pt}
        \end{minipage}
    }
    \hspace{-8pt}
    \subfloat[Ours]{
        \begin{minipage}{0.3\linewidth}
            \includegraphics[height=\linewidth]{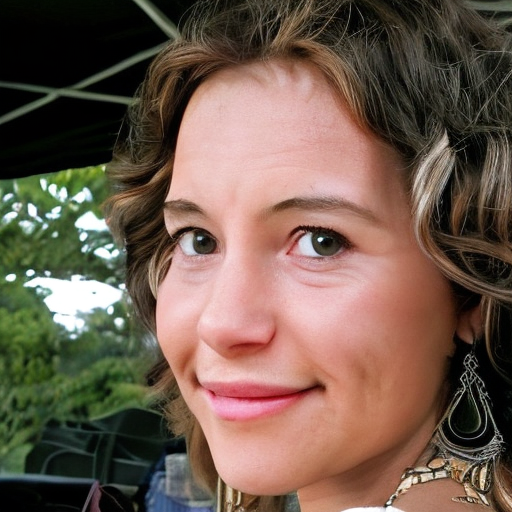} \\ \vspace{-9pt}
        \end{minipage}
    }
    \caption{Effectiveness of bidirectional latent alignment. (a) is the input image. (b) is the result of directly concatenating the non-id non-identity embedding $e_{\text{non-id}}$ and identity embedding $e_{\text{id}}$ to guide the diffusion model for face anonymization. (c) is our result.
    Simply concatenating $e_{\text{non-id}}$ and $e_{\text{id}}$ fail to preserve identity-independent low-level details.}
     \label{fig:compose}
\end{figure}

To tackle this issue, we first propose an alignment and recomposition strategy that harmonizes the semantics of these two representations before synthesis. Specifically, since $e_{\text{non-id}}$ and $e_{\text{id}}$ are extracted from different source images, they may encode distinct synthesis constraints (e.g., variations in pose or viewpoint). To ensure coherent semantic correspondence, we introduce a \emph{bidirectional latent alignment module} designed to perform semantic alignment and identity-decoupled latent recomposition. Within this cross-attenton based module, each semantic feature set acts as the \emph{query} while attending to the other as \emph{keys} and \emph{values}, enabling content-aware retrieval and reassembly in the latent space. 
This bidirectional alignment mechanism establishes an explicit and learnable pathway for mapping identity attributes onto the appropriate spatial regions, while simultaneously allowing the non-identity features to selectively attend to identity cues consistent with the given viewpoint. As a result, the model yields a harmonized latent representation in which structural layout and identity details are disentangled yet mutually consistent. Such a representation facilitates more effective downstream denoising and image synthesis, producing results that preserve both identity fidelity and geometric coherence.

Formally, the interaction between the identity condition embedding $e_{\text{id}}$ and the spatial non-identity semantic embedding $e_{\text{non-id}}$ is realized through a bidirectional cross-attention process defined as:

\begin{equation}
\begin{aligned}
& t_{\text{non-id}} = \text{CrossAttn}(\text{q}=e_{\text{non-id}},\ \text{k} = \text{v} = e_{\text{id}}), \\
& t_{\text{id}} = \text{CrossAttn}(\text{q}=e_{\text{id}},\ \text{k} = \text{v} = e_{\text{non-id}}).
\end{aligned}
\end{equation}

The resulting fused tokens, $t_{\text{non-id}}$ and $t_{\text{id}}$, encapsulate both spatial and identity-aware features. These are subsequently injected into the LDM, enabling fine-grained control over identity attributes while preserving spatial structure, semantic consistency, and photorealistic fidelity in the generated output.

\subsubsection{Identity-Guided Latent Harmonizer}

Once the non-identity token $t_{\text{non-id}}$ and the identity token $t_{\text{id}}$ are aligned, we introduce the IGLH, which replaces the conventional cross-attention layers with the dual-branch conditional diffusion blocks in the UNet architecture. The IGLH performs spatially-aware and scale-adaptive fusion between identity and non-identity representations by learning dynamic modulation masks that regulate the relative contributions of $t_{\text{id}}$ and $t_{\text{non-id}}$ across spatial locations and network depths. 
Through this mechanism, the model achieves fine-grained, region-aware control over feature integration, enabling it to selectively emphasize identity-relevant cues or suppress them when focusing on identity-irrelevant structures. Such adaptive harmonization ensures that the synthesized results maintain both identity consistency and spatial coherence throughout the generation process.

Formally, let $f_i$ denote the input feature map at the $i$-th UNet layer. We first apply the standard self-attention operation to capture global contextual dependencies:

\begin{equation}
f_i' = \text{SelfAttn}(f_i),
\end{equation}
where $f_i'$ represents a globally contextualized feature representation. Next, we generate a spatial mask $m_i \in [0, 1]$ via a lightweight gating mechanism:

\begin{equation}
m_i = \sigma\bigl( \text{Gate}(f_i') \bigr),
\end{equation}
where $\sigma(\cdot)$ is the sigmoid activation function, and $\text{Gate}(\cdot)$ denotes a shallow MLP applied across spatial locations.

Meanwhile, the tokens $t_{\text{non-id}}$ and $t_{\text{id}}$ are first processed independently via self-attention to capture intra-token dependencies at each scale. These refined tokens are then used as key-value pairs in two parallel cross-attention branches, both using $f_i'$ as the query:
\begin{equation}
\begin{aligned}
f_i^{\text{non-id}} &= \text{CrossAttn}\bigl(\text{q} = f_i',\ \text{k} = \text{v} = \text{SelfAttn}(t_{\text{non-id}})\bigr), \\
f_i^{\text{id}} &= \text{CrossAttn}\bigl(\text{q} = f_i',\ \text{k} = \text{v} = \text{SelfAttn}(t_{\text{id}})\bigr).
\end{aligned}
\end{equation}

The outputs of the two attention branches are then fused using the spatial mask $M_i$, allowing for region-wise selection between identity-relevant and identity-agnostic features:

\begin{equation}
f_{i+1} = m_i \cdot f_i^{\text{id}} + (1 - m_i) \cdot f_i^{\text{non-id}}.
\end{equation}

This spatially-adaptive fusion empowers the model to selectively emphasize identity or non-identity features across different image regions and UNet depths. By replacing standard attention layers with IGLH across multiple scales of the denoising UNet, we enable fine-grained, context-aware disentanglement and fusion. This results in improved identity controllability and enhanced visual fidelity in the final outputs.

\subsection{Orthogonal Identity Mapping}

As illustrated in Fig.~\ref{fig:orthogonal_sampling}, conventional random sampling suffers from variability due to the uncontrolled geometric relationship between the identity latent vector $\mathbf{v}$ and a Gaussian-sampled vector $\mathbf{r}$. Depending on their alignment, $\mathbf{r}$ may form an obtuse angle with $\mathbf{v}$ (Fig.~\ref{fig:orthogonal_sampling}(a)) or be more aligned (Fig.~\ref{fig:orthogonal_sampling}(b)), leading to inconsistent anonymization.

To overcome this, we introduce an Orthogonal Identity Mapping (OIM) strategy within the learned identity space of the ID-VAE during inference. Unlike conventional random sampling, our method ensures that the generated identity control vector is always orthogonal to the source identity embedding, thereby guaranteeing consistent anonymization. Specifically, given the original identity embedding $e^{\text{x}}_{\text{id}}$ extracted by ArcFace~\cite{deng2019arcface}, we encode it into the ID-VAE latent space as $\mathbf{v} = E_{\text{id}}(e^{\text{x}}_{\text{id}})$. A random latent vector $\mathbf{r} \sim \mathcal{N}(\mathbf{0}, \mathbf{I})$ is then projected onto the subspace orthogonal to $\mathbf{v}$, and the resulting component is decoded to produce the anonymized identity vector:

\begin{equation}
    e^{\text{ctrl}}_{\text{id}} = D_\text{id} \left( \mathbf{r} - \frac{\langle \mathbf{r}, \mathbf{v} \rangle}{\|\mathbf{v}\|^2} \cdot \mathbf{v} \right).
\end{equation}

Here, let $\mathbf{u} = \mathbf{r} - \frac{\langle \mathbf{r}, \mathbf{v} \rangle}{\|\mathbf{v}\|^2} \mathbf{v}$, which denotes the orthogonal projection of $\mathbf{r}$ onto the complement of $\mathbf{v}$. 
It is straightforward to verify that $\mathbf{u}$ lies in the orthogonal subspace since
\begin{equation}
    \mathbf{u} \cdot \mathbf{v} = \mathbf{r} \cdot \mathbf{v} - \frac{\langle \mathbf{r} , \mathbf{v} \rangle}{\|\mathbf{v}\|^2} (\mathbf{v} \cdot \mathbf{v}) = 0.
\end{equation}

\begin{figure}[t]
    \centering
    \includegraphics[width=\linewidth]{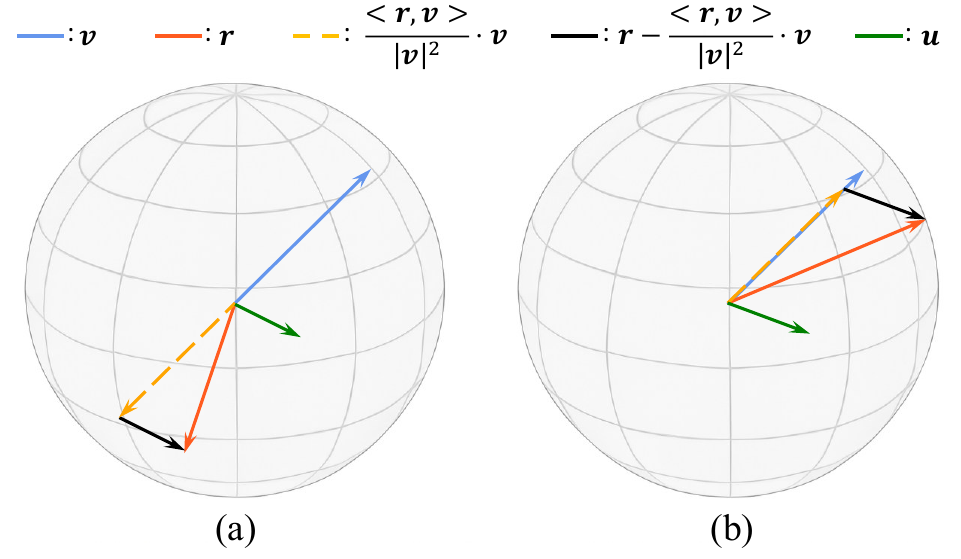}
    \caption{Orthogonal sampling in the latent space of the ID-VAE. (a) is an obtuse angle case, while (b) is an acute angle case. Our method ensures that the sampled identity vector is always orthogonal to the original identity embedding, eliminating uncertainty from random alignment.
    }
    \label{fig:orthogonal_sampling}
\end{figure}

This construction guarantees that the anonymized identity vector is fully decorrelated from the original embedding, eliminating the uncertainty introduced by random alignment. At the same time, since $\mathbf{r}$ is still sampled from a Gaussian distribution, the diversity of anonymized identities is preserved.
By enforcing strict orthogonality, our method provides a principled mechanism for generating identity vectors that are both diverse and fully anonymous, thereby significantly enhancing the robustness of identity anonymization.

\subsection{Loss Functions}

To enable precise and controllable manipulation of identity information while preserving identity-irrelevant attributes, we design a unified training objective composed of three complementary loss components: the \textit{Diffusion Loss}, the \textit{Identity-Related Loss}, and the \textit{ID-VAE Loss}. Together, these losses provide a principled optimization framework that jointly promotes visual fidelity, identity controllability, and semantic disentanglement.

\subsubsection{Diffusion Loss}
To ensure high-quality image generation, we employ a two-part diffusion loss comprising a noise prediction term and a latent reconstruction term.
The first component follows the standard noise prediction loss used in DDPM~\cite{DDPM}. At each timestep $t$, the model learns to predict the noise $\epsilon$ added to a clean latent $z_0$ to produce the noisy latent $z_t$:
\begin{equation}
 \mathcal{L}_{\text{diff-noise}} = \mathbb{E}_{z, t, \epsilon} \left[ \left\| \epsilon - \epsilon_\theta(z_t, t) \right\|^2 \right],
\end{equation}
\noindent where $\epsilon_\theta(z_t, t)$ is the model's prediction of the added noise. This objective encourages accurate denoising across the diffusion process.

The second component improves training stability and reconstruction consistency by explicitly recovering the original latent $z_0$ from the predicted noise~\cite{DDPM}:
\begin{equation}
 \hat{z}_0 = \frac{z_t - \sqrt{1 - \bar{\alpha}_t} \cdot \epsilon_\theta(z_t, t)}{\sqrt{\bar{\alpha}_t}},
\label{eq:z0}
\end{equation}
\noindent where $\bar{\alpha}_t$ denotes the cumulative noise schedule. The corresponding reconstruction loss is given by:
\begin{equation}
 \mathcal{L}_{\text{diff-recon}} = \mathbb{E}_{z_0, t} \left[ \left\| z_0 - \hat{z}_0 \right\|^2 \right].
  \label{eq:reconstruction}
\end{equation}

As the diffusion timestep $t$ increases, the predicted $\hat{z}_0$ becomes increasingly noisy and less reliable. To mitigate this, we introduce a time-dependent weighting factor $\bar{\alpha}_t$ (E.q.~\eqref{eq:z0}) to downweight the influence of reconstructions from later, noisier timesteps. The overall diffusion denoising loss is then defined as:
\begin{equation}
 \mathcal{L}_{\text{diff}} = \mathcal{L}_{\text{diff-noise}} + \lambda_1 \cdot \bar{\alpha}_t \cdot \mathcal{L}_{\text{diff-recon}},
 \label{eq:diffusion-loss}
\end{equation}
\noindent where $\lambda_1$ balances the contributions of the two terms. This formulation encourages accurate noise estimation and robust latent reconstruction, enhancing the realism and consistency of the generated outputs.

\subsubsection{Identity-Related Loss}
To guide the model in modulating identity-specific attributes while preserving non-identity characteristics, we introduce two complementary loss terms: an identity similarity loss and a multi-scale identity-region prediction loss.

The \textit{identity similarity loss} aligns the identity of the generated image with a given control vector. Specifically, we measure the cosine similarity between the identity embedding of the generated image and the reference identity vector extracted by the ArcFace model~\cite{deng2019arcface}:
\begin{equation}
 \mathcal{L}_{\text{id-sim}} = 1 - \cos\Bigl(\text{ArcFace}\bigl(D_{\rm diff}(\hat{z}_0)\bigr), e_{\text{id}}^{\text{ctrl}}\Bigr),
  \label{eq:id-sim}
\end{equation}
\noindent where $D_{\rm diff}$ is the diffusion VAE decoder, $\text{ArcFace}(\cdot)$ extracts identity embeddings, and $e_{\text{id}}^{\text{ctrl}}$ denotes the control identity vector, $\cos$ is the cosine similarity function. 

To further disentangle the multi-scale identity-related features spatially, we incorporate a \textit{multi-scale identity-region prediction loss}. This encourages the model to localize identity-relevant regions across multiple spatial resolutions. Formally:
\begin{equation}
 \mathcal{L}_{\text{id-region}} = \frac{1}{J} \sum_{j=1}^{J} \mathcal{L}_{\text{BCE}}^{(j)},
\end{equation}
\noindent where $J$ denotes the number of scales, and the binary cross-entropy loss at each scale $j$ is given by:
\begin{equation}
 \mathcal{L}_{\text{BCE}}^{(j)} = - \frac{1}{N} \sum_{i=1}^{N} m_i \log\bigl(\sigma(\hat{m}_i^{(j)})\bigr) + (1 - m_i) \log\bigl(1 - \sigma(\hat{m}_i^{(j)})\bigr),
\end{equation}
\noindent where $N$ is the number of spatial locations, $m_i$ is the ground-truth binary label from a facial parsing model, $\hat{m}_i^{(j)}$ is the predicted score, and $\sigma(\cdot)$ is the sigmoid function.

As in Eq.~\eqref{eq:diffusion-loss}, we apply the time-dependent weight $\bar{\alpha}_t$ to the identity similarity term to account for increasing uncertainty at later timesteps. The complete identity-related loss is:
\begin{equation}
 \mathcal{L}_{\text{id}} = \bar{\alpha}_t \cdot \mathcal{L}_{\text{id-sim}} + \lambda_2 \cdot \mathcal{L}_{\text{id-region}},
 \label{eq:id-loss}
\end{equation}
\noindent where $\lambda_2$ adjusts the influence of the region prediction term. This dual formulation enables semantic control over identity features and improves spatial disentanglement of identity-relevant content.

\subsubsection{ID-VAE Loss}
To ensure a well-structured latent space for identity representation, we adopt a standard VAE loss composed of an identity embedding reconstruction term and a Kullback–Leibler (KL) divergence~\cite{kingma2013auto}.
The reconstruction loss encourages the decoder to preserve identity semantics by minimizing the discrepancy between the original and reconstructed identity embeddings:
\begin{equation}
 \mathcal{L}_{\text{VAE-recon}} = \| e_{\text{id}}^{y} - e_{\text{id}}^{\text{ctrl}} \|_2^2,
\end{equation}
\noindent where $e_{\text{id}}^{y}$ is the reference embedding and $e_{\text{id}}^{\text{ctrl}}$ is its reconstruction from the latent representation.

To regularize the latent space, the KL divergence term aligns the posterior with a standard Gaussian prior:
\begin{equation}
 \mathcal{L}_{\text{KL}} = -\frac{1}{2} \sum_{k=1}^{d} \bigl(1 + \log(\sigma_k^2) - \mu_k^2 - \sigma_k^2 \bigr),
\end{equation}
\noindent where $\mu_k$ and $\sigma_k$ are the $k$-th components of the predicted mean and variance vectors. This encourages smoothness, diversity, and sampleability of the latent identity space, supporting robust identity randomization during inference.

Thus, the overall ID-VAE loss is defined as:
\begin{equation}
 \mathcal{L}_{\text{VAE}} = \mathcal{L}_{\text{VAE-recon}} + \lambda_3 \cdot \mathcal{L}_{\text{KL}},
\end{equation}
\noindent where $\lambda_3$ controls the regularization strength.

\subsubsection{Total Loss}
The overall training objective for our ID\textsuperscript{2}Face framework is defined as:
\begin{equation}
 \mathcal{L}_{\text{total}} = \mathcal{L}_{\text{diff}} + \mathcal{L}_{\text{id}} + \mathcal{L}_{\text{VAE}},
\end{equation}

\section{Experiments}

\subsection{Settings}

\subsubsection{Implementation Details} 
Our method is implemented in PyTorch and trained on four NVIDIA RTX 4090 GPUs. We adopt the Stable Diffusion v2.1 base model~\cite{ldm} as the backbone of our framework. Model optimization is performed using the AdamW~\cite{kingma2014adam} optimizer with a fixed learning rate of $1 \times 10^{-4}$.
Training is conducted over a total of 200{,}000 iterations, divided into two distinct stages. During the first 100{,}000 iterations, we pre-train the model using images of resolution $256 \times 256$, with a batch size of 8 per GPU. In the second stage, we increase the resolution to $512 \times 512$ to improve image generation fidelity, and reduce the batch size to 2 per GPU to accommodate the increased computational requirements.
The loss weights are empirically set as $\lambda_1 = 0.1$, $\lambda_2 = 0.1$, and $\lambda_3 = 1 \times 10^{-5}$. For inference, we employ the DDIM sampler~\cite{DDIM} with 40 denoising steps, striking a balance between image quality and computational efficiency.

\subsubsection{Datasets} 
For training, we utilize the VGGFace2-HQ~\cite{chen2023simswap} dataset and the first 65{,}000 images drawn from FFHQ~\cite{FFHQ} datasets. Specifically, we randomly select two images of the same identity from VGGFace2-HQ to form identity pairs. For FFHQ, we pair each image with itself to ensure consistency in identity representation.
For evaluation, we employ two datasets: the last 5{,}000 images from FFHQ and 30{,}000 images from CelebA-HQ~\cite{CelebAHQ}. These diverse datasets allow for a comprehensive assessment of both identity anonymization and attribute preservation.

\begin{figure*}[t!]
    \centering
    \subfloat[Input]{
        \begin{minipage}{0.120\linewidth}
            \includegraphics[height=\linewidth]{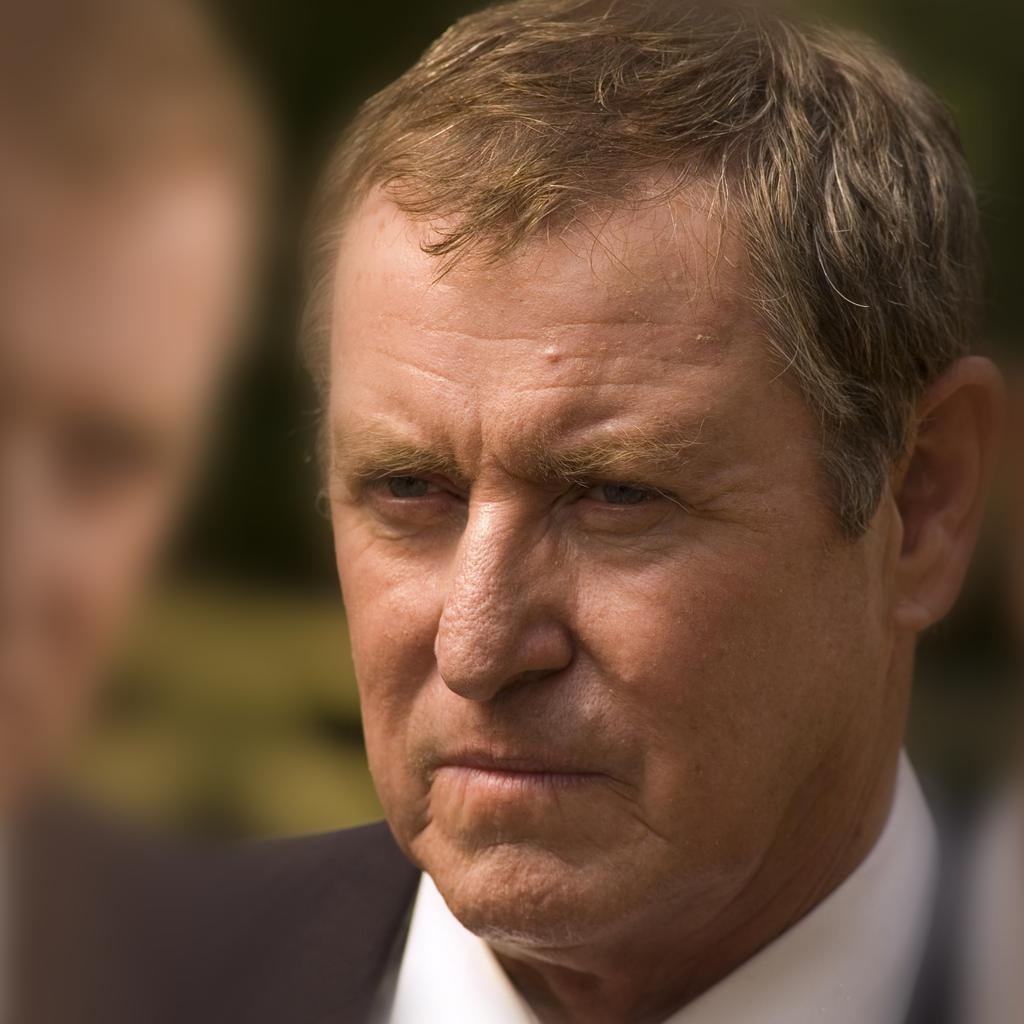} \\ \vspace{-10pt}
            \includegraphics[height=\linewidth]{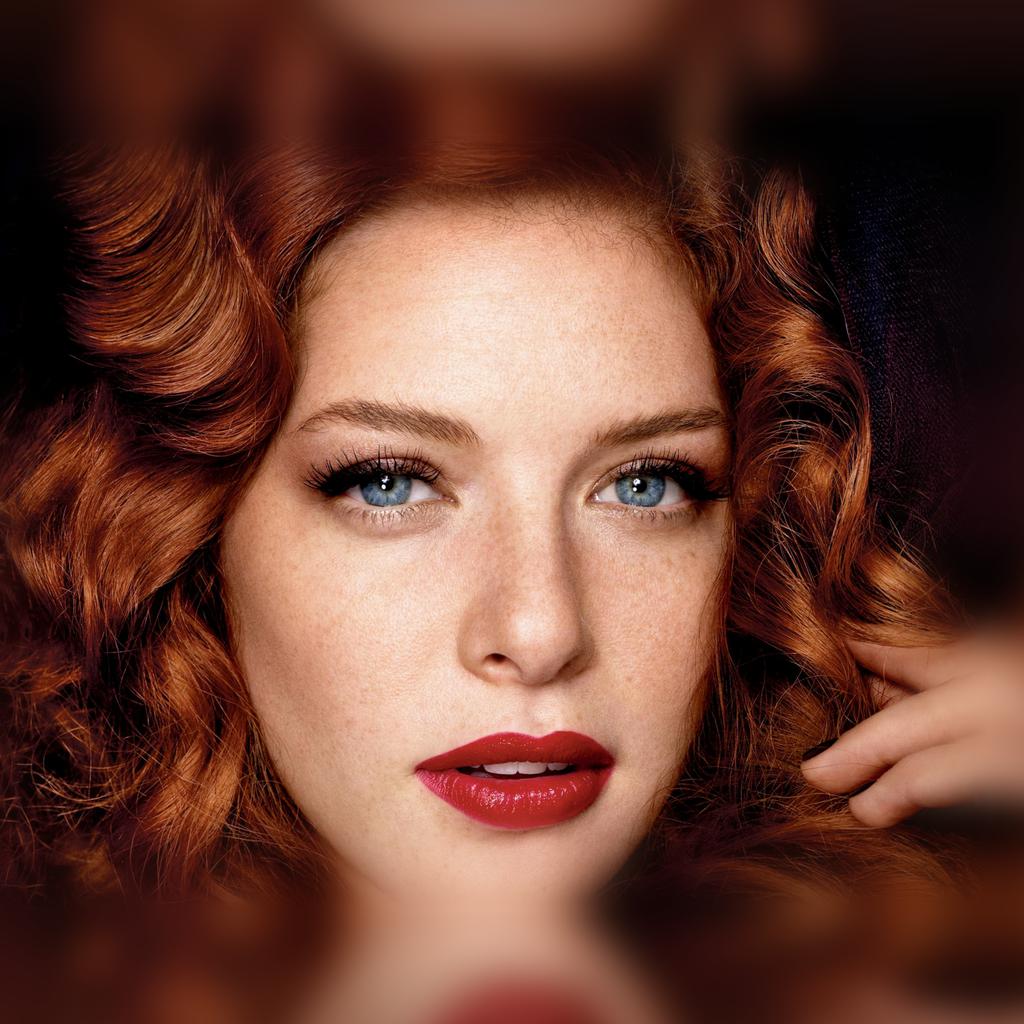} \\ \vspace{-10pt}
            \includegraphics[height=\linewidth]{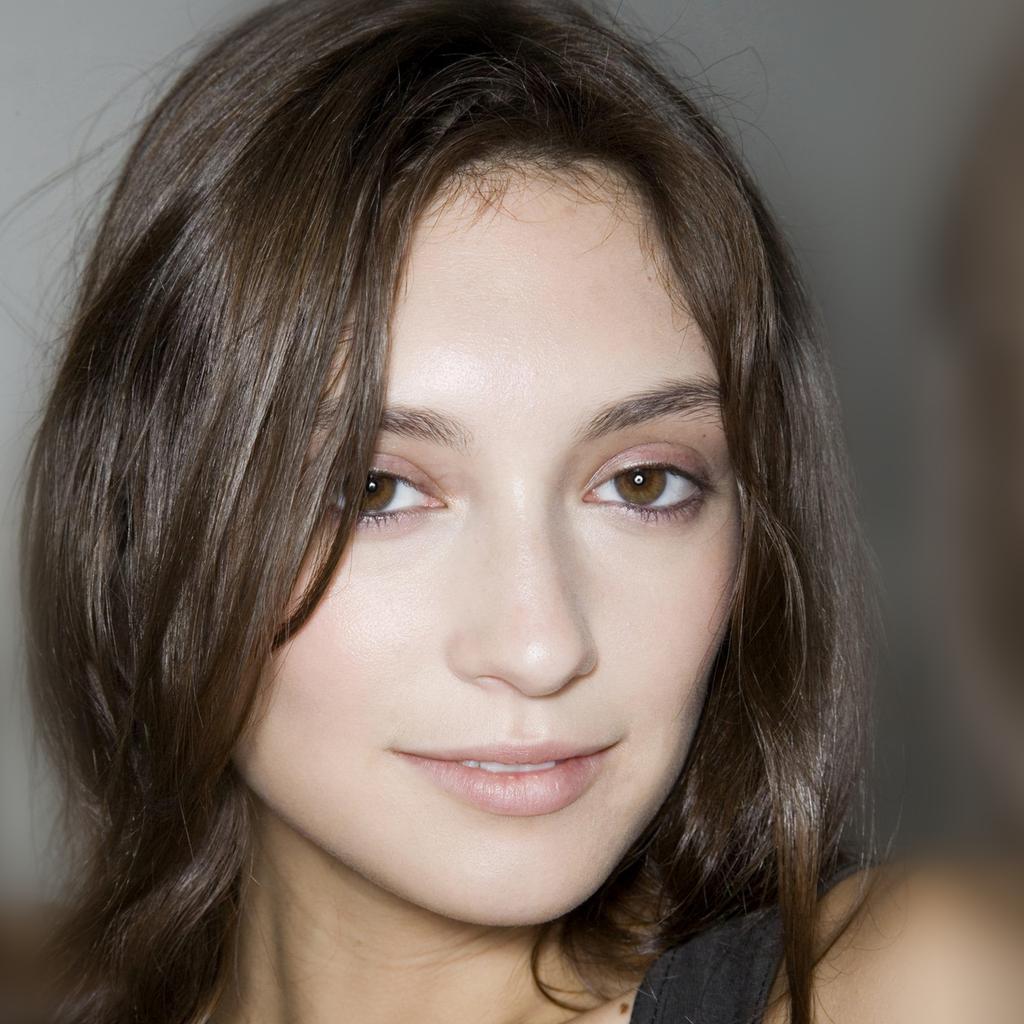} \\ \vspace{-10pt}
            \includegraphics[height=\linewidth]{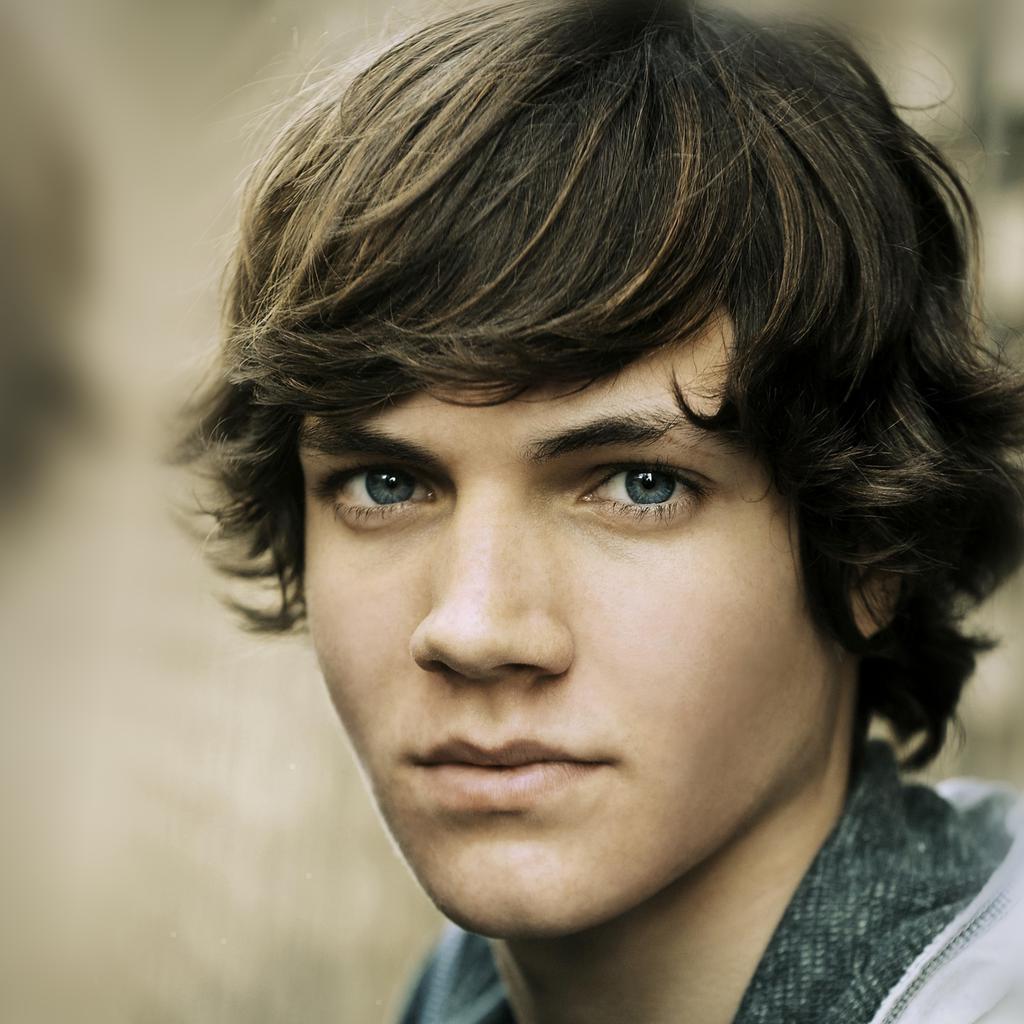} \\ \vspace{-10pt}
            \includegraphics[height=\linewidth]{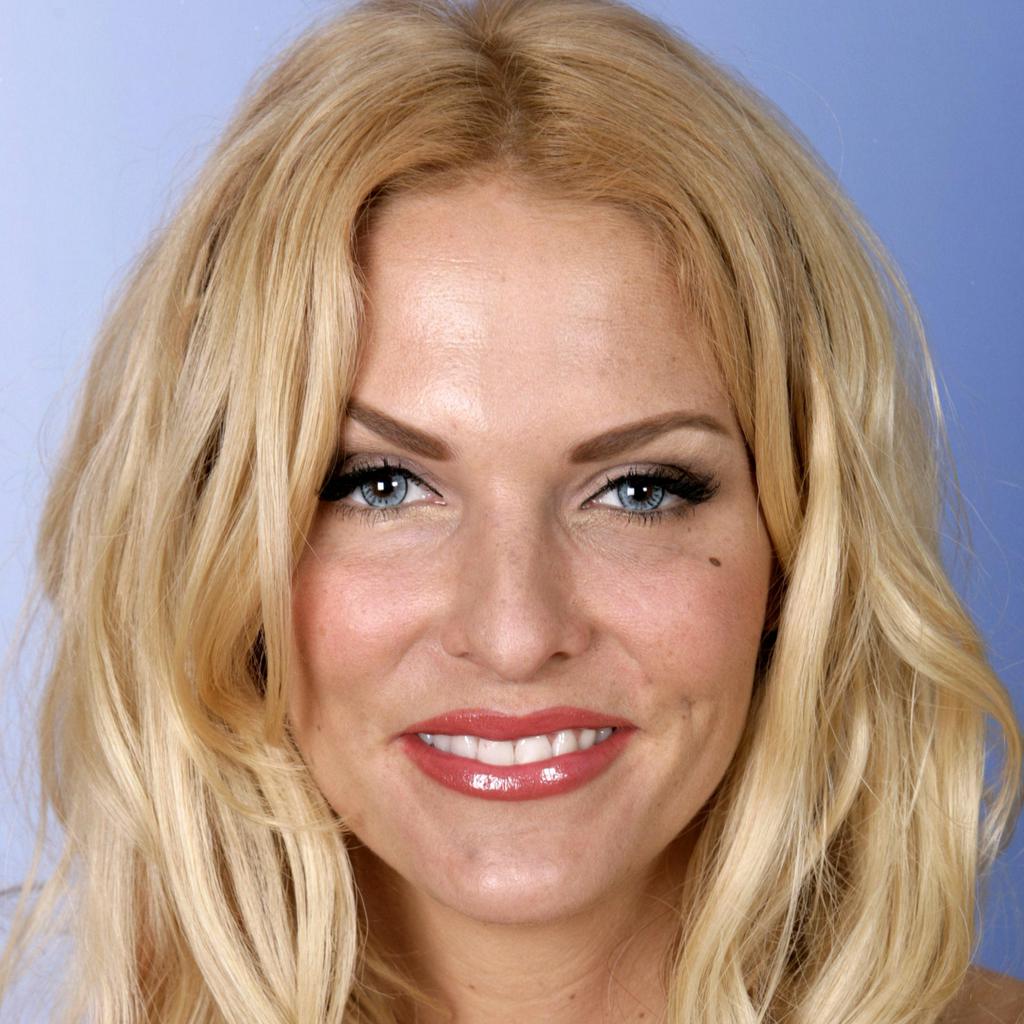} \\ \vspace{-10pt}
        \end{minipage}
    }
    \hspace{-9pt}
    \subfloat[RiDDLE~\cite{RiDDLE}]{
        \begin{minipage}{0.120\linewidth}
            \includegraphics[height=\linewidth]{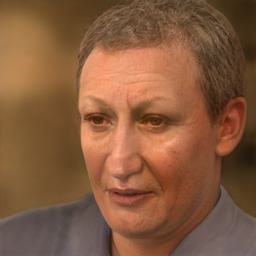} \\ \vspace{-10pt}
            \includegraphics[height=\linewidth]{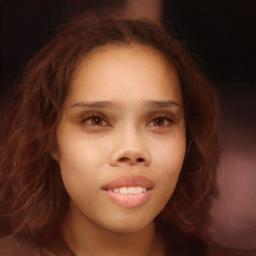} \\ \vspace{-10pt}
            \includegraphics[height=\linewidth]{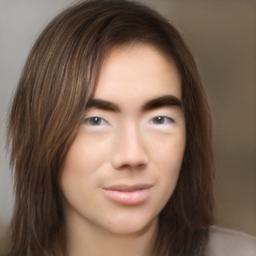} \\ \vspace{-10pt}
            \includegraphics[height=\linewidth]{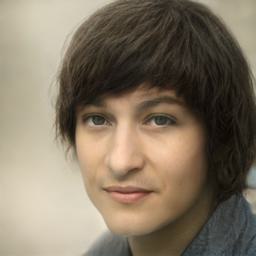} \\ \vspace{-10pt}
            \includegraphics[height=\linewidth]{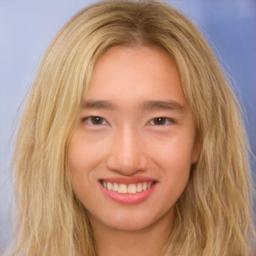} \\ \vspace{-10pt}
        \end{minipage}
    }
    \hspace{-9pt}
    \subfloat[G$^2$Face~\cite{yang2024g}]{
        \begin{minipage}{0.120\linewidth}
            \includegraphics[height=\linewidth]{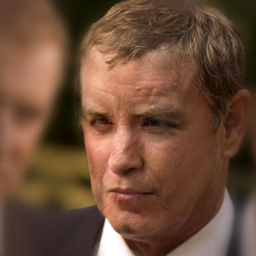} \\ \vspace{-10pt}
            \includegraphics[height=\linewidth]{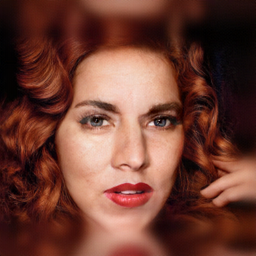} \\ \vspace{-10pt}
            \includegraphics[height=\linewidth]{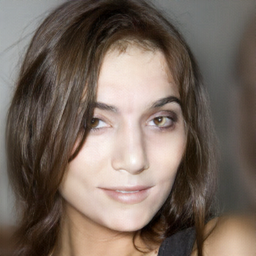} \\ \vspace{-10pt}
            \includegraphics[height=\linewidth]{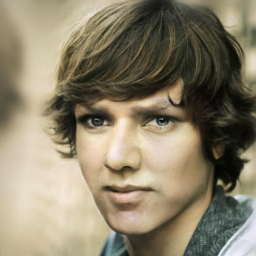} \\ \vspace{-10pt}
            \includegraphics[height=\linewidth]{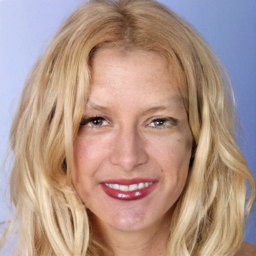} \\ \vspace{-10pt}
        \end{minipage}
    }
    \hspace{-9pt}
    \subfloat[AIDPro~\cite{AIDPro}]{
        \begin{minipage}{0.120\linewidth}
            \includegraphics[height=\linewidth]{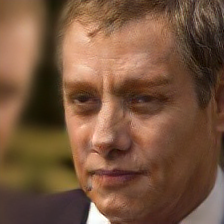} \\ \vspace{-10pt}
            \includegraphics[height=\linewidth]{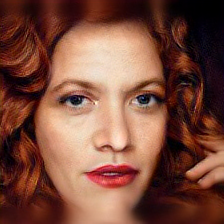} \\ \vspace{-10pt}
            \includegraphics[height=\linewidth]{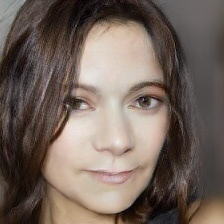} \\ \vspace{-10pt}
            \includegraphics[height=\linewidth]{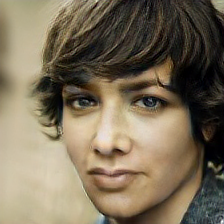} \\ \vspace{-10pt}
            \includegraphics[height=\linewidth]{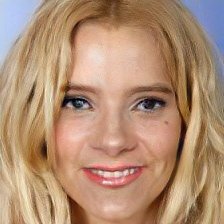} \\ \vspace{-10pt}
        \end{minipage}
    }
    \hspace{-9pt}
    \subfloat[DiffPrivacy~\cite{DiffPrivacy}]{
        \begin{minipage}{0.120\linewidth}
            \includegraphics[height=\linewidth]{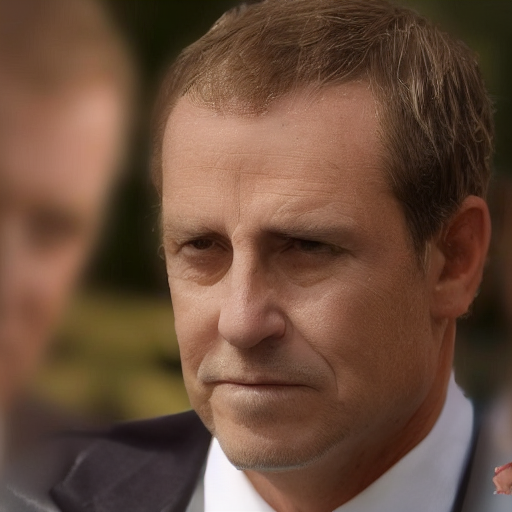} \\ \vspace{-10pt}
            \includegraphics[height=\linewidth]{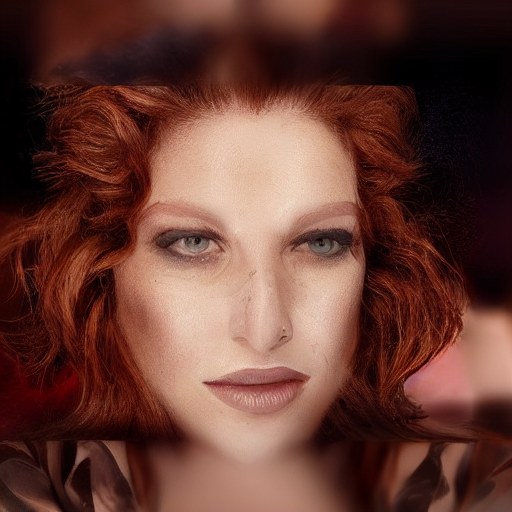} \\ \vspace{-10pt}
            \includegraphics[height=\linewidth]{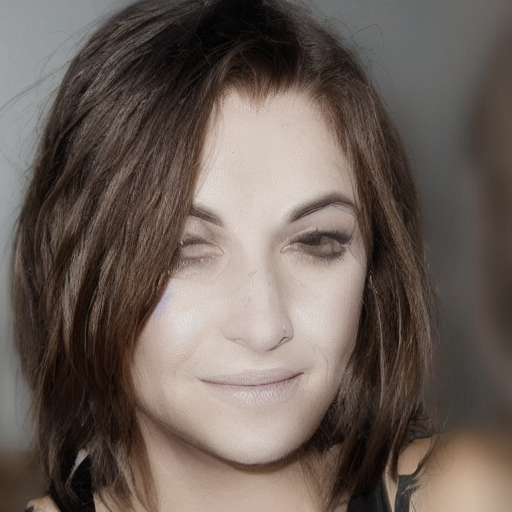} \\ \vspace{-10pt}
            \includegraphics[height=\linewidth]{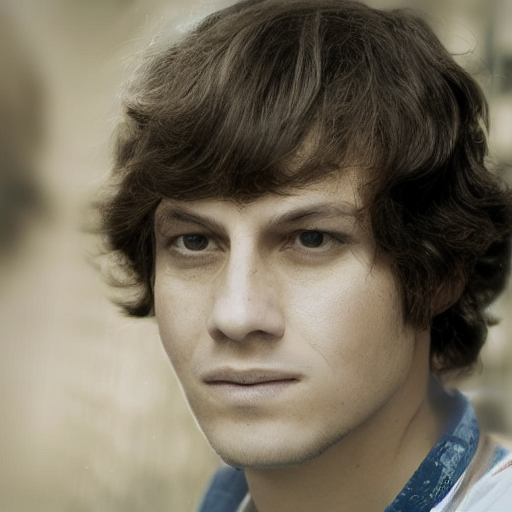} \\ \vspace{-10pt}
            \includegraphics[height=\linewidth]{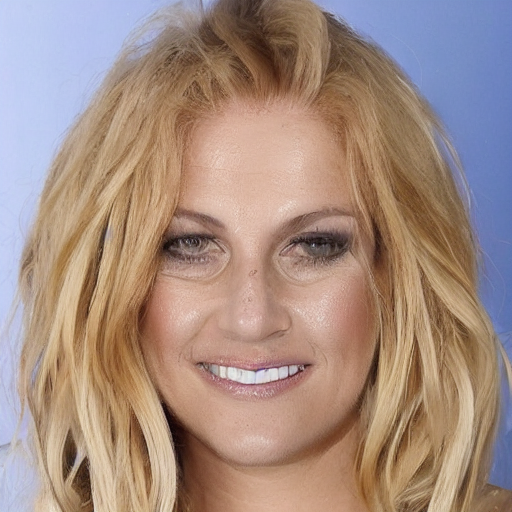} \\ \vspace{-10pt}
        \end{minipage}
    }
    \hspace{-9pt}
    \subfloat[FAMS~\cite{FAMS}]{
        \begin{minipage}{0.120\linewidth}
            \includegraphics[height=\linewidth]{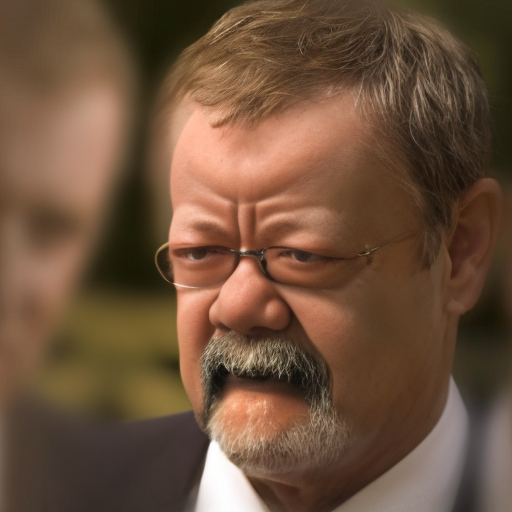} \\ \vspace{-10pt}
            \includegraphics[height=\linewidth]{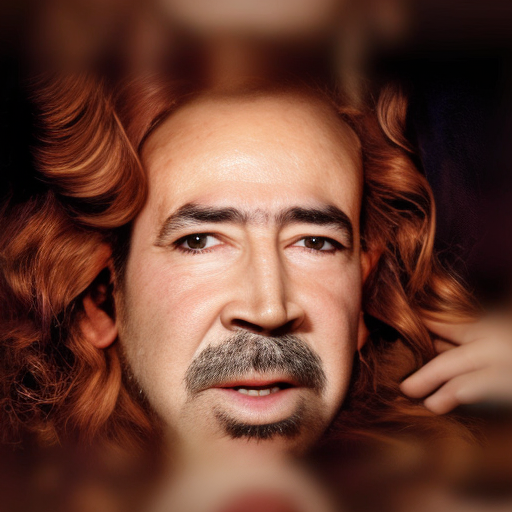} \\ \vspace{-10pt}
            \includegraphics[height=\linewidth]{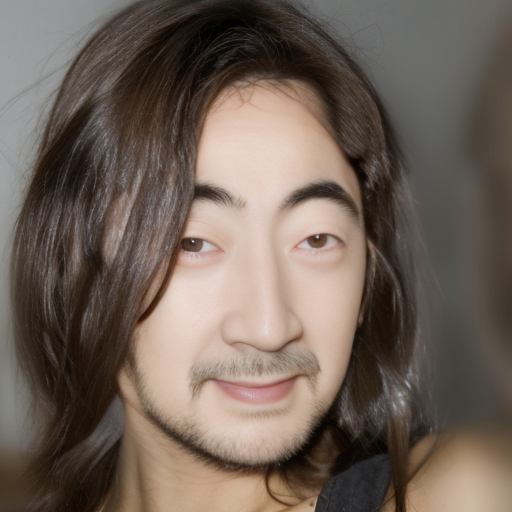} \\ \vspace{-10pt}
            \includegraphics[height=\linewidth]{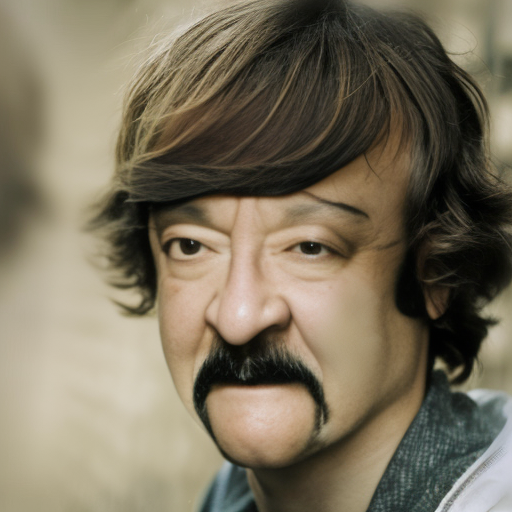} \\ \vspace{-10pt}
            \includegraphics[height=\linewidth]{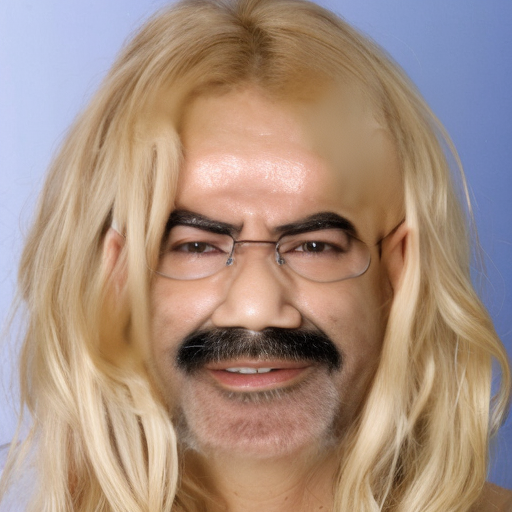} \\ \vspace{-10pt}
        \end{minipage}
    }
    \hspace{-9pt}
    \subfloat[NullFace~\cite{NullFace}]{
        \begin{minipage}{0.120\linewidth}
            \includegraphics[height=\linewidth]{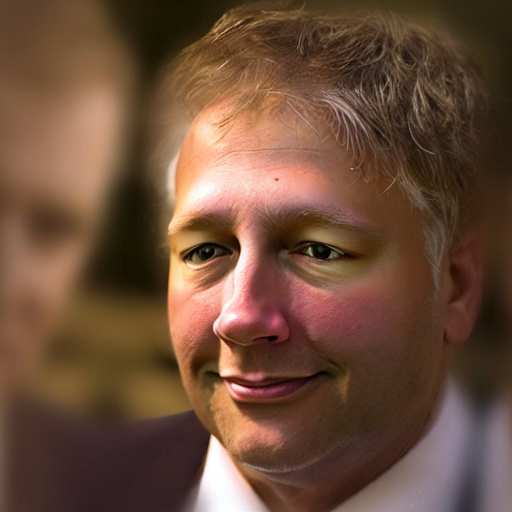} \\ \vspace{-10pt}
            \includegraphics[height=\linewidth]{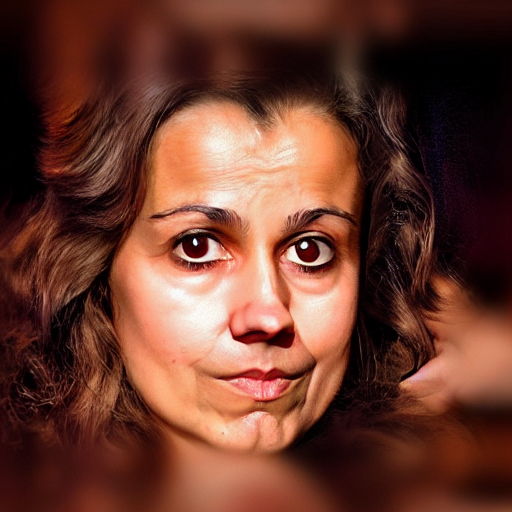} \\ \vspace{-10pt}
            \includegraphics[height=\linewidth]{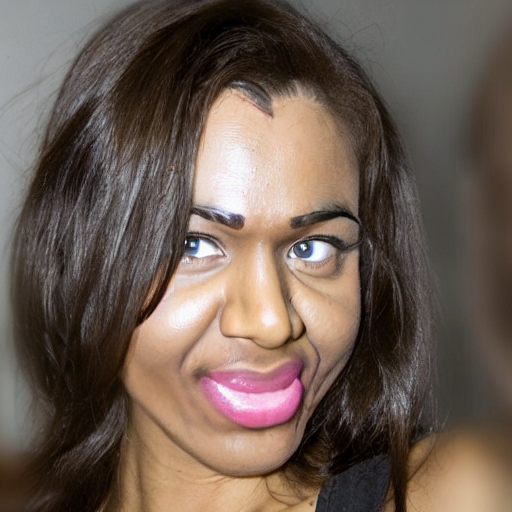} \\ \vspace{-10pt}
            \includegraphics[height=\linewidth]{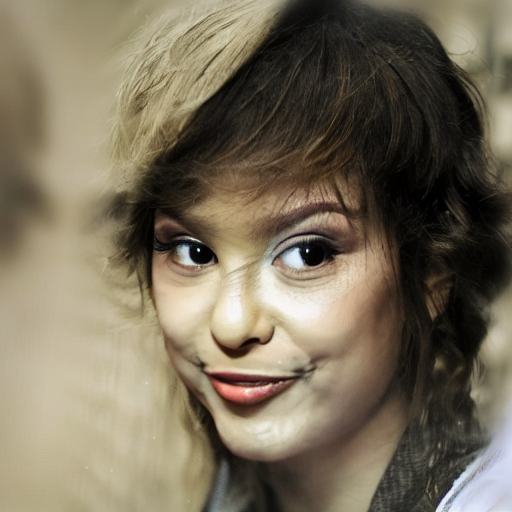} \\ \vspace{-10pt}
            \includegraphics[height=\linewidth]{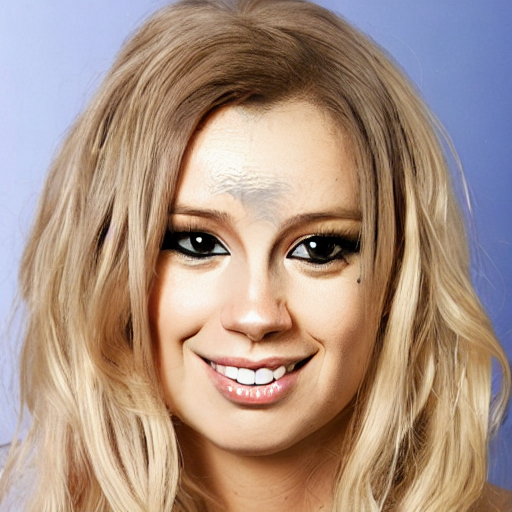} \\ \vspace{-10pt}
        \end{minipage}
    }
    \hspace{-9pt}
    \subfloat[Ours]{
        \begin{minipage}{0.120\linewidth}
            \includegraphics[height=\linewidth]{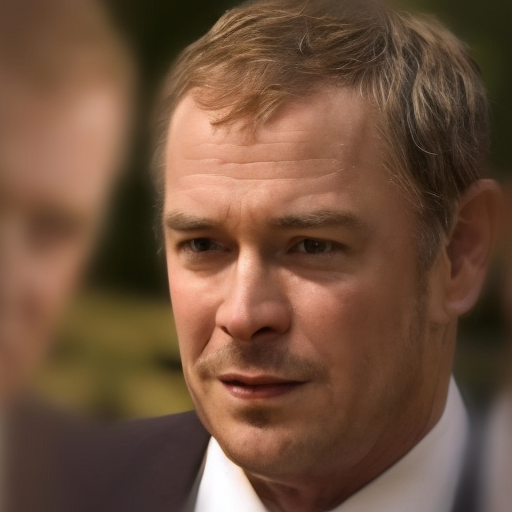} \\ \vspace{-10pt}
            \includegraphics[height=\linewidth]{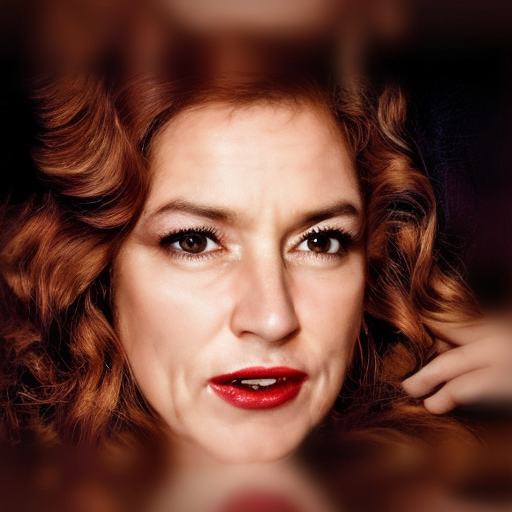} \\ \vspace{-10pt}
            \includegraphics[height=\linewidth]{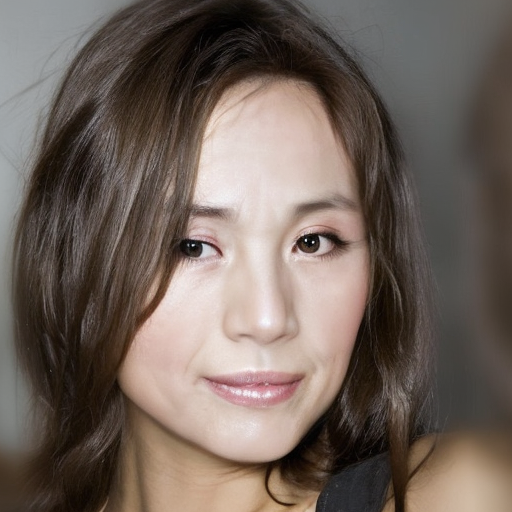} \\ \vspace{-10pt}
            \includegraphics[height=\linewidth]{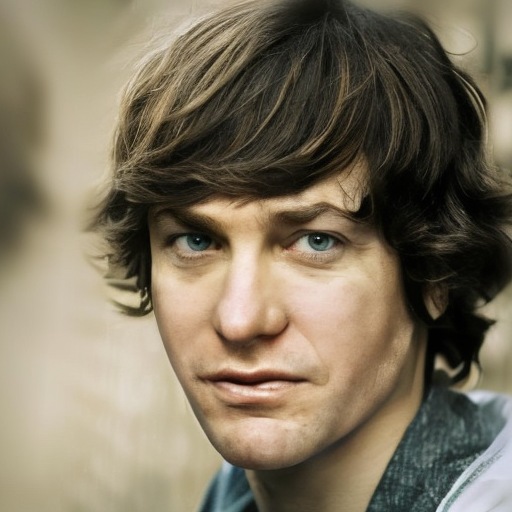} \\ \vspace{-10pt}
            \includegraphics[height=\linewidth]{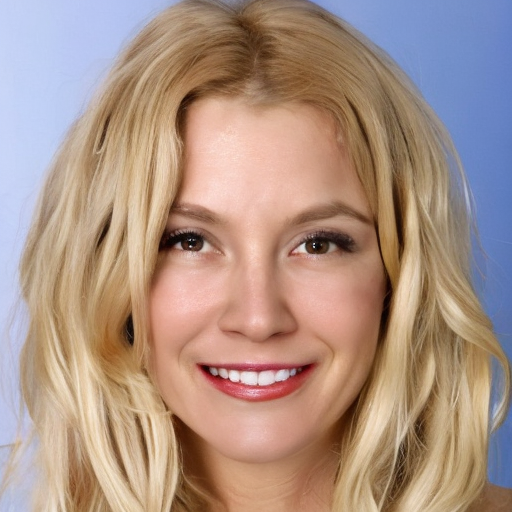} \\ \vspace{-10pt}
        \end{minipage}
    }
    \caption{
        Qualitative comparison of face anonymization and recovery among different methods on the CelebA-HQ dataset~\cite{CelebAHQ}. (a)-(h) are the original face images, the anonymization results of RiDDLE~\cite{RiDDLE}, G$^2$Face~\cite{yang2024g}, AIDPro~\cite{AIDPro}, DiffPrivacy~\cite{DiffPrivacy}, FAMS~\cite{FAMS}, NullFace~\cite{NullFace} and our method, respectively. Note that RiDDLE, G$^2$Face, and AIDPro are GAN-based methods, while the others are diffusion-based methods. Our method achieves superior anonymization and image quality compared to the SOTA methods.}
     \label{fig:visual_comparison}
\end{figure*}

\subsubsection{Metrics} 
We evaluate the performance of our method using three categories of metrics.
\textit{(i) Identity Removal:} To measure the effectiveness of identity anonymization, we report Top-1 and Top-5 identity retrieval accuracies, mean Average Precision (mAP), and the cosine similarity of identity embeddings. These metrics are computed using three representative face recognition models: ArcFace~\cite{deng2019arcface}, AdaFace~\cite{kim2022adaface}, and TopoFR~\cite{dan2024topofr}. Lower values across these metrics correspond to stronger anonymization. 
\textit{(ii) Attribute Preservation:} We evaluate the consistency of identity-independent attributes by measuring the L2 distance between the anonymized and original images across multiple facial properties: (a) facial landmarks (68 keypoints) using MTCNN~\cite{zhang2016joint}; (b) head pose using HopeNet~\cite{ruiz2018fine}; (c) facial expression using FECNet~\cite{vemulapalli2019compact}; and (d) gaze direction using L2CS-Net~\cite{abdelrahman2023l2cs}. Lower distance values indicate better preservation of semantic and perceptual features unrelated to identity. 
\textit{(iii) Image Quality:} We assess the visual quality of generated images using MUSIQ~\cite{ke2021musiq} for perceptual assessment and FID~\cite{heusel2017gans} for distributional similarity to real images. Higher MUSIQ scores and lower FID values reflect superior image fidelity and realism.

\subsection{Anonymization Comparison with State-of-the-art Methods}
\subsubsection{Qualitative Comparison}
To validate the effectiveness of our proposed method in face anonymization and high-fidelity image generation, we conduct a visual comparison on the CelebA-HQ~\cite{CelebAHQ} dataset against six state-of-the-art (SOTA) approaches, including three GAN-based methods (RiDDLE~\cite{RiDDLE}, G$^2$Face~\cite{yang2024g}, and AIDPro~\cite{AIDPro}) and three diffusion-based methods (DiffPrivacy~\cite{DiffPrivacy}, FAMS~\cite{FAMS}, and NullFace~\cite{NullFace}). As illustrated in Fig.~\ref{fig:visual_comparison}, our approach delivers markedly superior image quality compared to the GAN-based methods~\cite{RiDDLE,AIDPro,yang2024g}, benefiting from the inherent generative strength of diffusion-based architectures. More importantly, compared to existing diffusion-based methods~\cite{DiffPrivacy,FAMS,NullFace}, our model strikes a better balance between effective identity obfuscation and visual realism, yielding anonymized images that are not only natural and semantically consistent but also more suitable for downstream tasks. This performance advantage arises from our inference-intervention-free framework design that enables explicit identity control during training, thereby avoiding the distribution shifts and quality degradation associated with inference-time manipulations employed by prior methods.
Furthermore, the proposed IGLH module enables effective decoupling and integration of identity-related and identity-irrelevant features across multiple spatial scales, thereby enhancing both the anonymization strength and the visual fidelity of the generated outputs.

\begin{table*}[t!]
    \centering
    \caption{Identity anonymization comparison of our method with SOTA methods on CelebA-HQ and FFHQ datasets. The best results are in \textbf{bold}, the second best are underlined. (ID-retrieval: Top-1 identity retrieval accuracy / Top-5 identity retrieval accuracy; mAP: mean Average Precision; ID-Sim: Identity Similarity).}
    \resizebox{\linewidth}{!}{
    \renewcommand{\arraystretch}{1.2}
    \begin{tabular}{l|l|ccc|ccc|ccc}
        \bottomrule
        \multirow{2}{*}{Dataset}   & \multirow{2}{*}{Method}            & \multicolumn{3}{c|}{ArcFace~\cite{deng2019arcface}}          & \multicolumn{3}{c|}{AdaFace~\cite{kim2022adaface}}           & \multicolumn{3}{c}{TopoFR~\cite{dan2024topofr}}              \\ \cline{3-11} 
                                   &                                    & ID-retrieval ↓           & mAP ↓           & ID-Sim. ↓       & ID-retrieval ↓           & mAP ↓           & ID-Sim. ↓       & ID-retrieval ↓           & mAP ↓           & ID-Sim. ↓       \\ \hline
        \multirow{6}{*}{CelebA-HQ} & RiDDLE~\cite{RiDDLE}               & 0.0126 / 0.0374          & 0.0304          & 0.1233          & 0.0288 / 0.0818          & 0.0604          & 0.1268          & 0.0246 / 0.0752          & 0.0560          & 0.1321          \\
                                   & G$^2$Face~\cite{yang2024g}         & 0.0148 / 0.0325          & 0.0259          & 0.1102          & 0.0216 / 0.0498          & 0.0373          & 0.0925          & 0.0178 / 0.0396          & 0.0307          & 0.0803          \\
                                   & AIDPro~\cite{AIDPro}               & 0.0125 / 0.0254          & 0.0221          & 0.1207          & 0.0147 / 0.0395          & 0.0196          & 0.0935          & 0.0183 / 0.0327          & 0.0219          & 0.0775          \\
                                   & DiffPrivacy~\cite{DiffPrivacy}     & 0.0172 / 0.0372          & 0.0301          & 0.1260          & 0.1598 / 0.2834          & 0.2216          & 0.1999          & 0.1238 / 0.2450          & 0.1833          & 0.1947          \\
                                   & FAMS~\cite{FAMS}                   & 0.0172 / 0.0372          & 0.0301          & 0.1260          & 0.0504 / 0.1050          & 0.0808          & 0.1380          & 0.0286 / 0.0628          & 0.0483          & 0.1175          \\
                                   & NullFace~\cite{NullFace}           & {\ul 0.0138   / 0.0224}  & {\ul 0.0211}    & {\ul 0.0856}    & {\ul 0.0112 / 0.0316}    & {\ul 0.0115}    & {\ul 0.0857}    & {\ul 0.0056 / 0.0138}    & {\ul 0.0108}    & {\ul 0.0672}    \\
                                   \rowcolor{gray!20}
                                   & Ours                               & \textbf{0.0000 / 0.0002} & \textbf{0.0003} & \textbf{0.0045} & \textbf{0.0040 / 0.0100} & \textbf{0.0092} & \textbf{0.0738} & \textbf{0.0036 / 0.0094} & \textbf{0.0080} & \textbf{0.0583} \\ \hline
        \multirow{6}{*}{FFHQ}      & RiDDLE~\cite{RiDDLE}               & 0.0182 / 0.0568          & 0.0428          & 0.1286          & 0.0344 / 0.0984          & 0.0724          & 0.1219          & 0.0402 / 0.1038          & 0.0781          & 0.1355          \\
                                   & G$^2$Face~\cite{yang2024g}         & 0.0146 / 0.0375          & 0.0301          & 0.1152          & 0.0246 / 0.0615          & 0.0453          & 0.0987          & 0.0264 / 0.0642          & 0.0491          & 0.0921          \\
                                   & AIDPro~\cite{AIDPro}               & {\ul 0.0092   / 0.0315}  & {\ul 0.0227}    & {\ul 0.0531}    & {\ul 0.0174 / 0.0580}    & {\ul 0.0419}    & {\ul 0.0961}    & 0.0260 / 0.0838          & 0.0424          & 0.1049          \\
                                   & DiffPrivacy~\cite{DiffPrivacy}     & 0.2944 / 0.4336          & 0.3643          & 0.2289          & 0.2770 / 0.4280          & 0.3526          & 0.2007          & 0.2118 / 0.3322          & 0.2743          & 0.1946          \\
                                   & FAMSe~\cite{FAMS}                  & 0.1588 / 0.2620          & 0.2137          & 0.2065          & 0.2770 / 0.4280          & 0.3526          & 0.2007          & 0.2118 / 0.3322          & 0.2743          & 0.1946          \\
                                   & NullFace~\cite{NullFace}           & 0.0128 / 0.0528          & 0.0241          & 0.0758          & 0.0230 / 0.0451          & 0.0543          & 0.1247          & {\ul 0.0154 / 0.0430}    & {\ul 0.0324}    & {\ul 0.0872}    \\
                                   \rowcolor{gray!20}
                                   & Ours                               & \textbf{0.0000 / 0.0018} & \textbf{0.0022} & \textbf{0.0173} & \textbf{0.0120 / 0.0420} & \textbf{0.0324} & \textbf{0.0838} & \textbf{0.0122 / 0.0348} & \textbf{0.0281} & \textbf{0.0692} \\ \toprule
        \end{tabular}
    }
    \label{tab:identity}
\end{table*}

\begin{table}[t!]
    \centering
    \caption{Attribute and image quality comparison of our method with SOTA methods on CelebA-HQ and FFHQ datasets.}
    \resizebox{\linewidth}{!}{
    \renewcommand{\arraystretch}{1.2}
    \setlength{\tabcolsep}{2pt}
    \begin{tabular}{l|l|cccc|cc}
    \bottomrule
    \multirow{2}{*}{Dataset}    & \multirow{2}{*}{Method}               & \multicolumn{4}{c|}{Attribute}                                        & \multicolumn{2}{c}{Image quality}   \\ \cline{3-8} 
                                &                                       & LM.↓            & Pose↓           & Exp.↓           & Gaze↓           & MUSIQ↑           & FID↓             \\ \hline
    \multirow{6}{*}{CelebA-HQ}  & RiDDLE~\cite{RiDDLE}                  & 10.2153         & 6.3365          & 0.3635          & 0.2672          & 56.9412          & 68.3889          \\
                                & G$^2$Face~\cite{yang2024g}            & 7.4321          & 4.1234          & 0.2567          & 0.2404          & 64.0499          & 12.6789          \\
                                & AIDPro~\cite{AIDPro}                  & 13.9803         & {\ul 3.5822}    & 0.3242          & 0.2513          & 55.6198          & 14.5528          \\
                                & DiffPrivacy~\cite{DiffPrivacy}        & 13.1254         & 5.8128          & 0.2849          & 0.3070          & 72.9146          & 22.5528          \\
                                & FAMS~\cite{FAMS}                      & {\ul 6.1096}    & 3.9854          & {\ul 0.2375}    & {\ul 0.2466}    & 71.0268          & 14.2307          \\
                                & NullFace~\cite{NullFace}              & 6.5369          & 4.1009          & 0.2481          & 0.2604          & \textbf{73.7165} & \textbf{10.8548} \\
                                \rowcolor{gray!20}
                                & Ours                                  & \textbf{5.5728} & \textbf{3.2009} & \textbf{0.2166} & \textbf{0.2377} & {\ul 72.9912}    & {\ul 11.2434}    \\ \hline
    \multirow{6}{*}{FFHQ}       & RiDDLE~\cite{RiDDLE}                  & 10.2978         & 6.9163          & 0.3732          & 0.3529          & 63.5916          & 60.9246          \\
                                & G$^2$Face~\cite{yang2024g}            & 7.8345          & 4.5123          & 0.2678          & 0.2990          & 64.1207          & 14.9962          \\
                                & AIDPro~\cite{AIDPro}                  & 11.9469         & 9.1822          & 0.3379          & 0.3229          & 63.0842          & 26.0405          \\
                                & DiffPrivacy~\cite{DiffPrivacy}        & 17.8886         & 6.4208          & 0.2918          & 0.3529          & 71.8727          & 17.4138          \\
                                & FAMS~\cite{FAMS}                      & {\ul 6.2041}    & {\ul 3.6263}    & {\ul 0.2290}    & {\ul 0.2995}    & 71.4544          & \textbf{9.1436}  \\
                                & NullFace~\cite{NullFace}              & 6.8986          & 4.1189          & 0.2395          & 0.3238          & {\ul 73.3516}    & 11.4206          \\
                                \rowcolor{gray!20}
                                & Ours                                  & \textbf{5.8560} & \textbf{3.6027} & \textbf{0.2216} & \textbf{0.2989} & \textbf{73.8899} & {\ul 9.6959}     \\ \toprule
    \end{tabular}
    }
    \label{tab:attribute}
\end{table}

\subsubsection{Quantitative Comparison}
We further conducted a detailed quantitative evaluation on two benchmark datasets, CelebA-HQ~\cite{CelebAHQ} and FFHQ~\cite{FFHQ}, to substantiate the superiority of our proposed approach over existing SOTA methods. The results are presented in Tables~\ref{tab:identity} and \ref{tab:attribute}. We assess performance across three categories of metrics. The first category focuses on identity removal, including Top-1 and Top-5 identity retrieval accuracy, mAP, and cosine similarity of identity embeddings. The second category evaluates the preservation of attributes unrelated to identity, measured by the L2 distance between the anonymized and original images across facial landmarks, head pose, expression, and gaze direction. The third category concerns image generation quality, assessed using MUSIQ~\cite{ke2021musiq} and FID scores~\cite{heusel2017gans}.

As evidenced by the table, our method consistently outperforms competing approaches across both identity-related and attribute-preservation metrics on both datasets. These results confirm that our method effectively eliminates identity information while maintaining fidelity in identity-agnostic facial features. This strong performance is primarily attributed to the proposed IDLR, which explicitly disentangles identity and non-identity features during training, thereby enhancing the model's capability to control identity generation and non-identity feature preservation independently. In addition, the IDLR module, enhanced with landmark fusion and the identity-mask diffusion training scheme, enables the model to retain rich non-identity-related information, such as facial structure, expressions, and pose. 
Furthermore, during inference, our OIM strategy reliably generates control vectors that are maximally disentangled from the original identity embeddings, leading to robust identity obfuscation.
Our method also ranks first or second in image quality metrics, indicating that the generated images exhibit high visual fidelity. This can be largely attributed to the powerful pretraining and inherent stability of the diffusion model employed in our framework.

\begin{figure}[t]
    \centering
    \subfloat[DiffPrivacy~\cite{DiffPrivacy}]{
        \vspace{-20pt}
        \begin{minipage}{0.4\linewidth}
            \includegraphics[height=\linewidth]{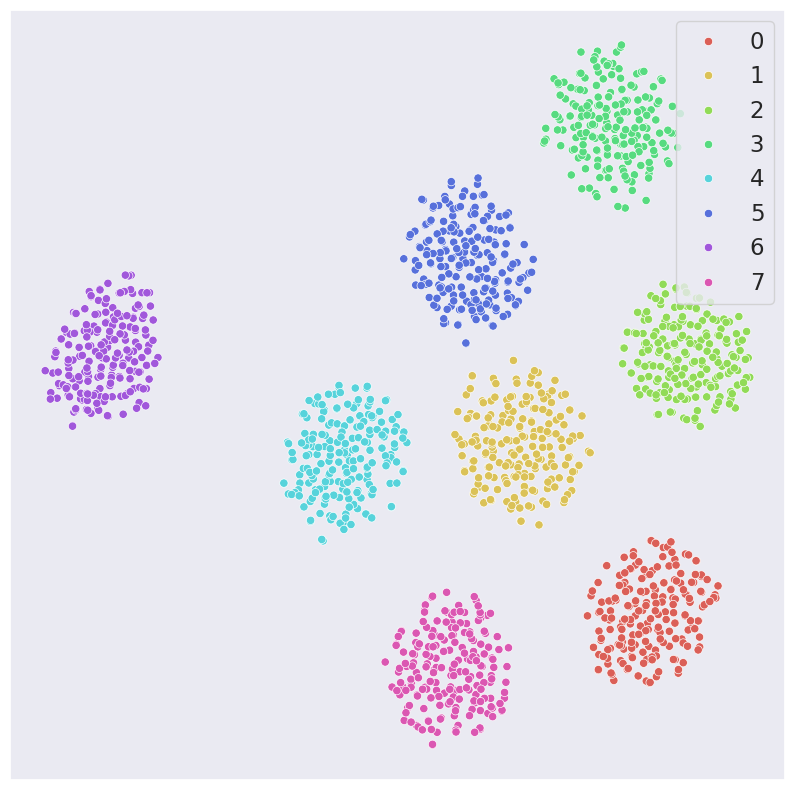} \\ \vspace{-10pt}
        \end{minipage}
    }
    \subfloat[FAMS~\cite{FAMS}]{
        \vspace{-20pt}
        \begin{minipage}{0.4\linewidth}
            \includegraphics[height=\linewidth]{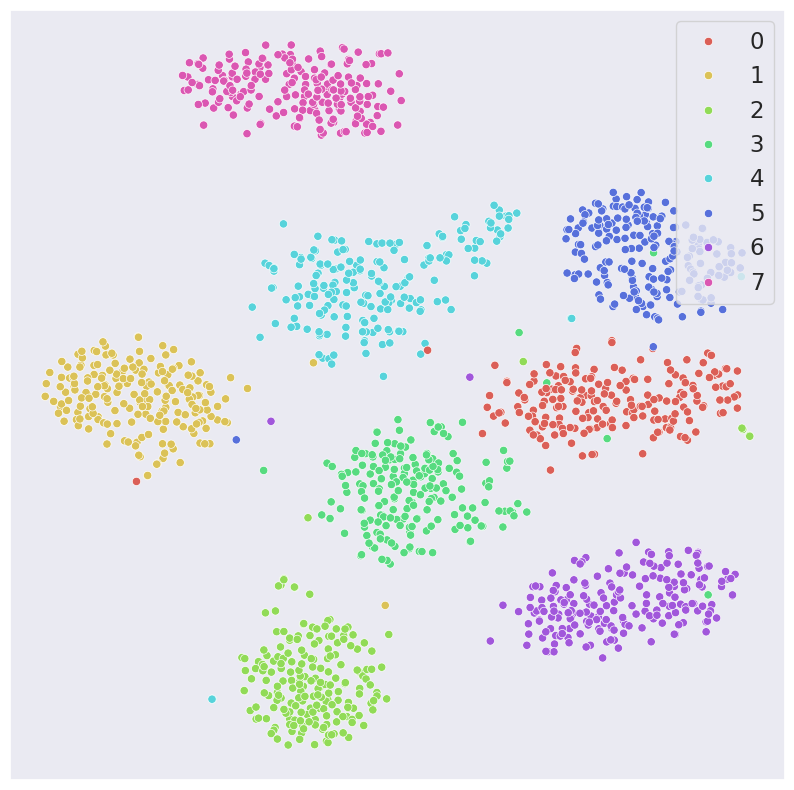} \\ \vspace{-10pt}
        \end{minipage}
    }
    \vspace{-6pt}
    \subfloat[NullFace~\cite{NullFace}]{
        \vspace{-12pt}
        \begin{minipage}{0.4\linewidth}
            \includegraphics[height=\linewidth]{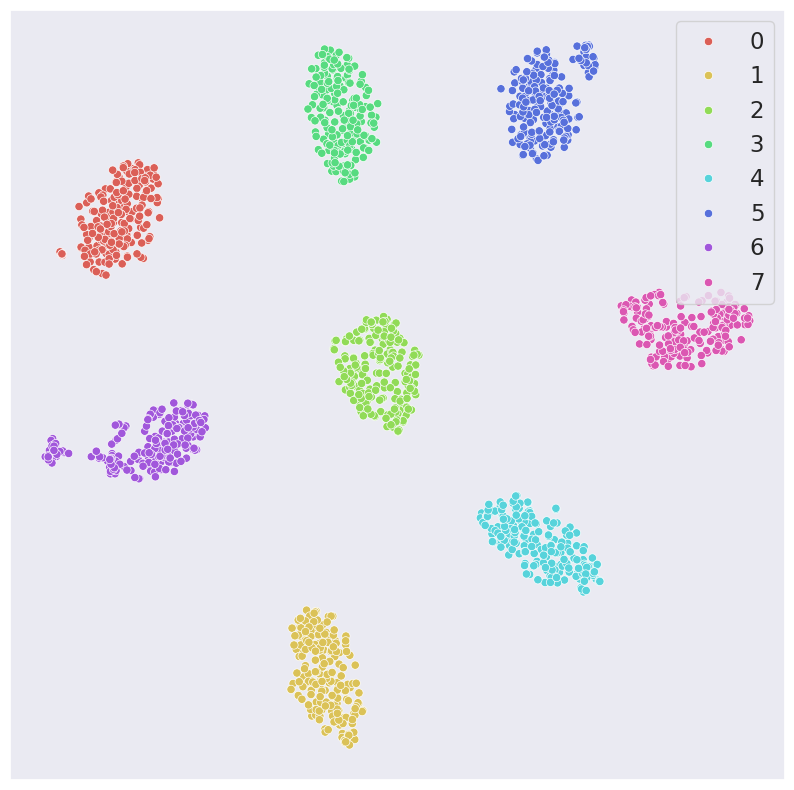} \\ \vspace{-10pt}
        \end{minipage}
    }
    \subfloat[Ours]{
        \vspace{-12pt}
        \begin{minipage}{0.4\linewidth}
            \includegraphics[height=\linewidth]{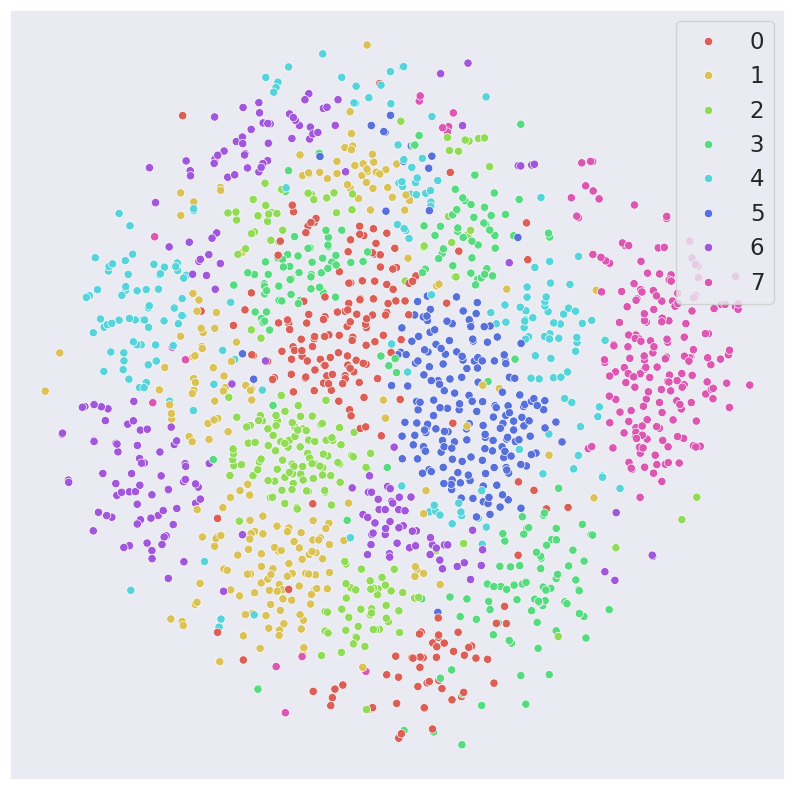} \\ \vspace{-10pt}
        \end{minipage}
    }
    \caption{
        Visualization of t-SNE results on identity embeddings from anonymized images generated by different diffusion-based methods.}
      \label{fig:tsne}
\end{figure}

\begin{figure}[t]
    \centering
    \subfloat[Input]{
        \begin{minipage}{0.18\linewidth}
            \includegraphics[height=\linewidth]{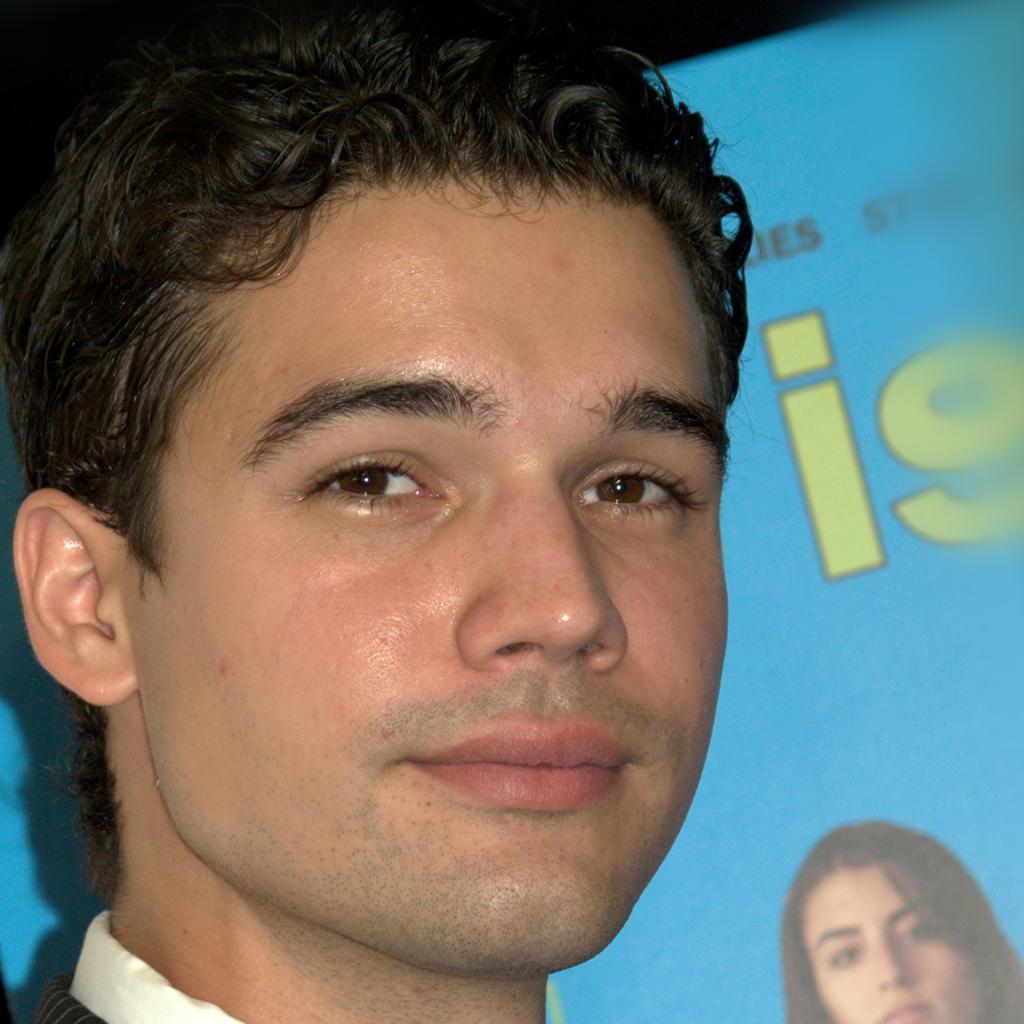} \\
            \includegraphics[height=\linewidth]{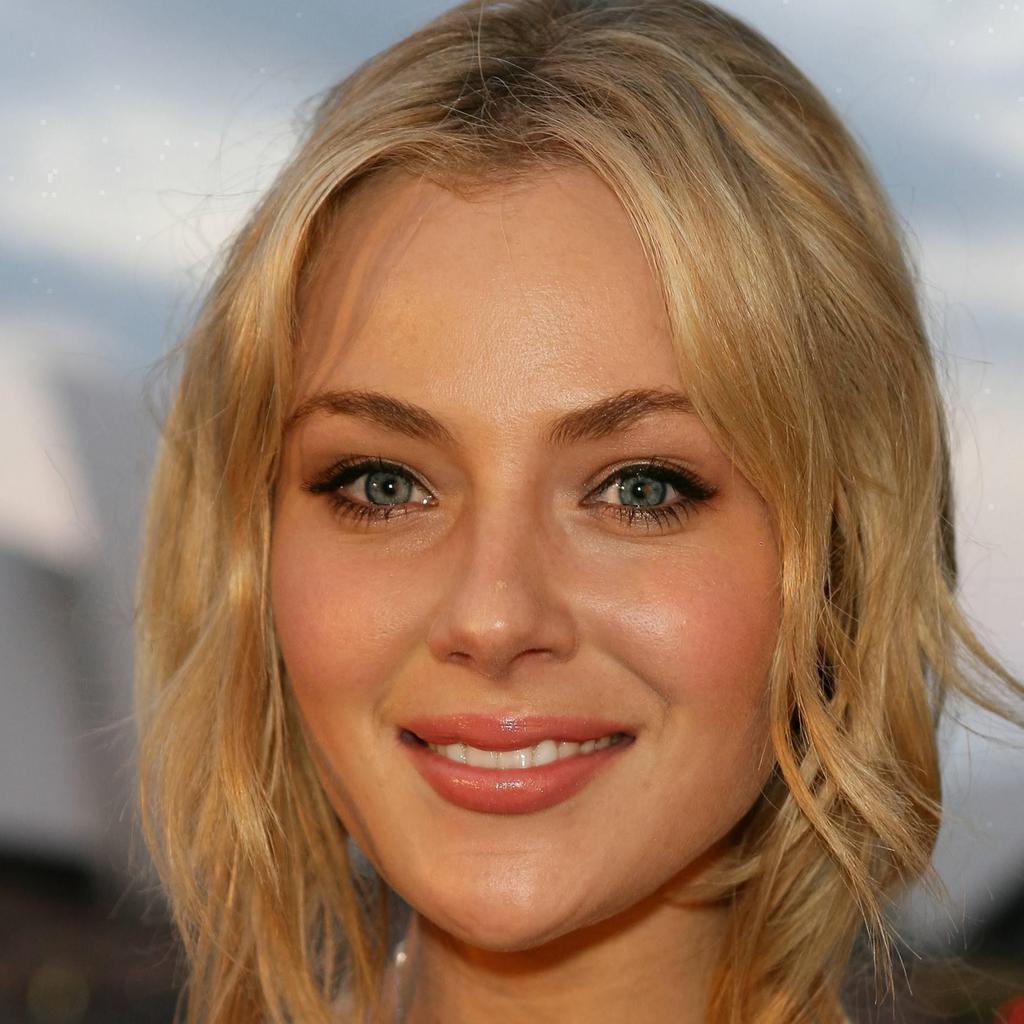} \\
            \includegraphics[height=\linewidth]{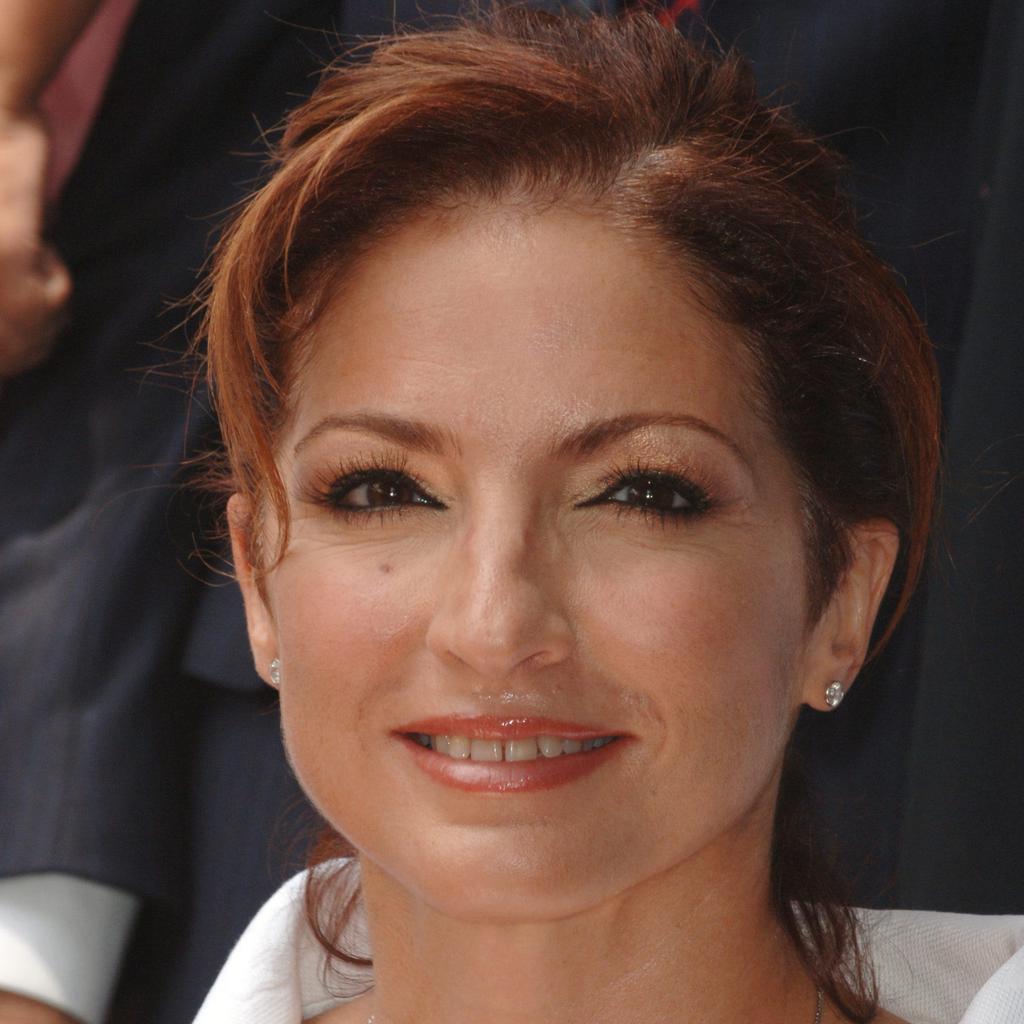} \\ \vspace{-10pt}
        \end{minipage}
    }
    \hspace{-12pt}
    \subfloat[$Y_1$]{
        \begin{minipage}{0.18\linewidth}
            \includegraphics[height=\linewidth]{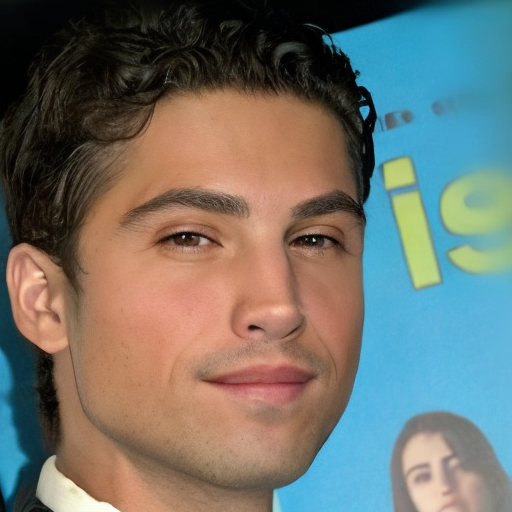} \\
            \includegraphics[height=\linewidth]{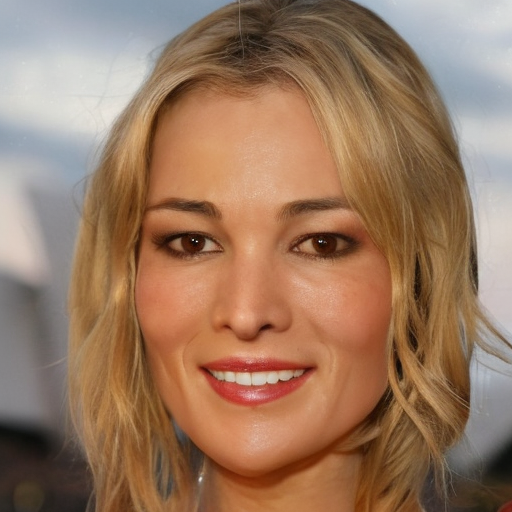} \\
            \includegraphics[height=\linewidth]{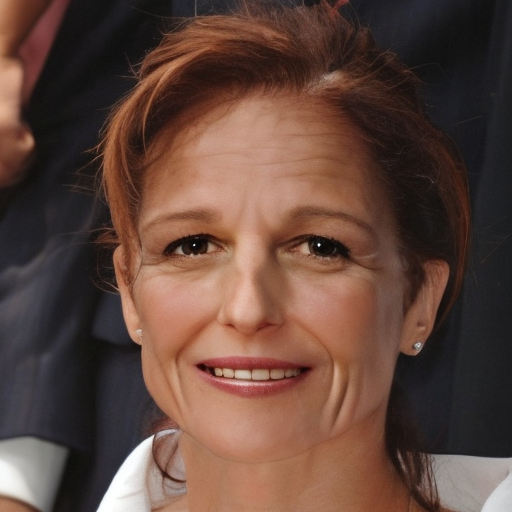} \\ \vspace{-10pt}
        \end{minipage}
    }
    \hspace{-12pt}
    \subfloat[$Y_2$]{
        \begin{minipage}{0.18\linewidth}
            \includegraphics[height=\linewidth]{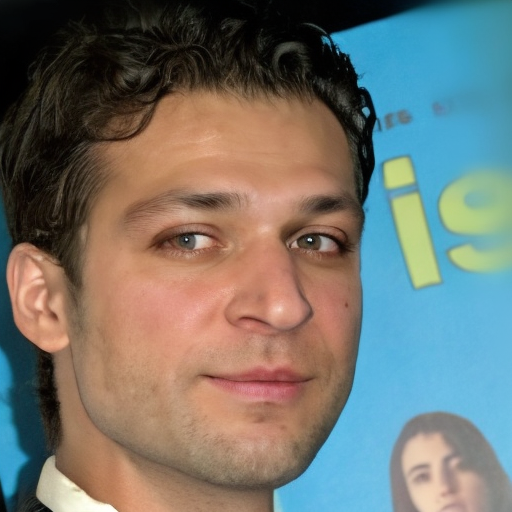} \\
            \includegraphics[height=\linewidth]{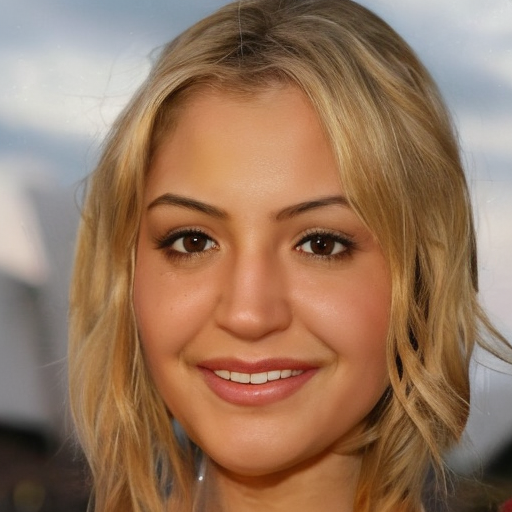} \\
            \includegraphics[height=\linewidth]{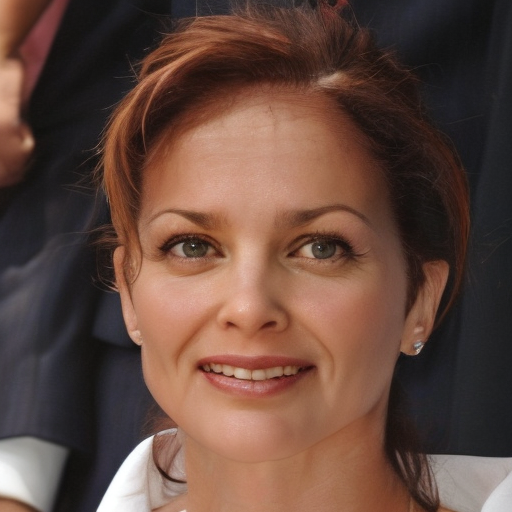} \\ \vspace{-10pt}
        \end{minipage}
    }
    \hspace{-12pt}
    \subfloat[$Y_3$]{
        \begin{minipage}{0.18\linewidth}
            \includegraphics[height=\linewidth]{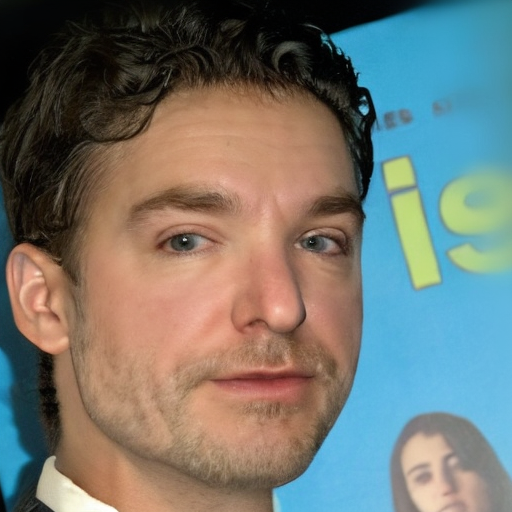} \\
            \includegraphics[height=\linewidth]{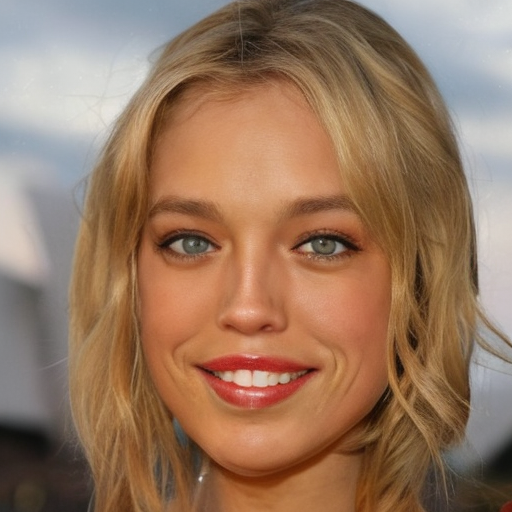} \\
            \includegraphics[height=\linewidth]{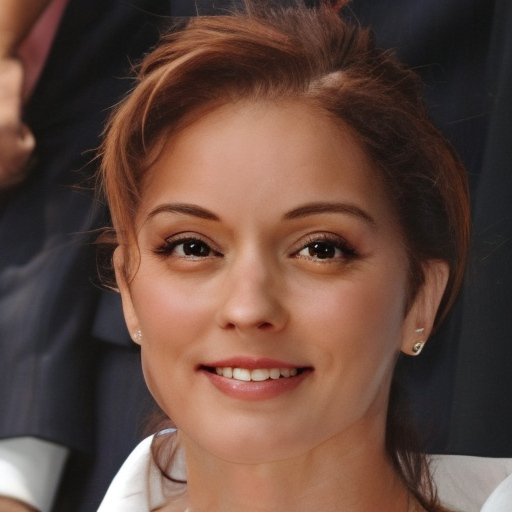} \\ \vspace{-10pt}
        \end{minipage}
    }
    \hspace{-12pt}
    \subfloat[Zoom in]{
        \begin{minipage}{0.27\linewidth}
            \includegraphics[height=\linewidth]{figs/subfigs/multi/multi_0_0.jpg} \\
            \includegraphics[height=\linewidth]{figs/subfigs/multi/multi_0_1.png} \\ \vspace{-10pt}
        \end{minipage}
    }
    \caption{
        Visual results of diverse anonymized counterparts for the same input face given different identity embeddings on the CelebA-HQ dataset. Zoom in for better visualization.}
    \label{fig:diverse_samples}
\end{figure}

\subsection{Anonymization Diversity Analysis}
In face anonymization, the diversity of generated outputs is crucial for both privacy protection and practical utility. To evaluate this aspect, we compare our method with representative diffusion-based anonymization approaches~\cite{DiffPrivacy,FAMS,NullFace} by visualizing the distribution of identity embeddings using t-SNE~\cite{van2008visualizing}. Specifically, we randomly select eight face images from the CelebA-HQ test set, generate 200 anonymized samples per image, extract their identity embeddings using ArcFace~\cite{deng2019arcface}, and project them into a two-dimensional space via t-SNE (Fig.~\ref{fig:tsne}).

The visualization demonstrates that our method produces significantly more diverse and widely distributed identity embeddings compared with existing diffusion-based methods. Samples derived from the same source image are scattered across the embedding space and frequently intermingle with those from other sources, making re-identification difficult. This indicates that our approach simultaneously achieves strong anonymization and high identity diversity.
The improvement stems from two key factors. First, our unified training framework explicitly learns to disentangle identity and non-identity features, yielding a model with stable and precise control over identity generation. In contrast, prior diffusion-based methods~\cite{DiffPrivacy,FAMS,NullFace} rely on post-inference optimization, which often causes mode collapse and limited diversity. Second, our ID-VAE combined with the proposed OIM strategy enables the generation of an unbounded variety of identities that remain decorrelated from the original embedding, ensuring robust anonymization while preserving diversity. Existing approaches, by contrast, depend on a single inference-time condition, inherently restricting variability.

Finally, Fig.~\ref{fig:diverse_samples} illustrates multiple anonymized outputs from the same source image. The results confirm that our method effectively removes identity-specific cues while preserving identity-irrelevant features such as facial structure and expression, and produces a rich variety of plausible anonymized faces.

\subsection{Ablation Study}

\begin{figure*}[t]
    \centering
    \captionsetup[subfloat]{font=scriptsize}
    \subfloat[Input]{
        \begin{minipage}{0.115\linewidth}
            \includegraphics[height=\linewidth]{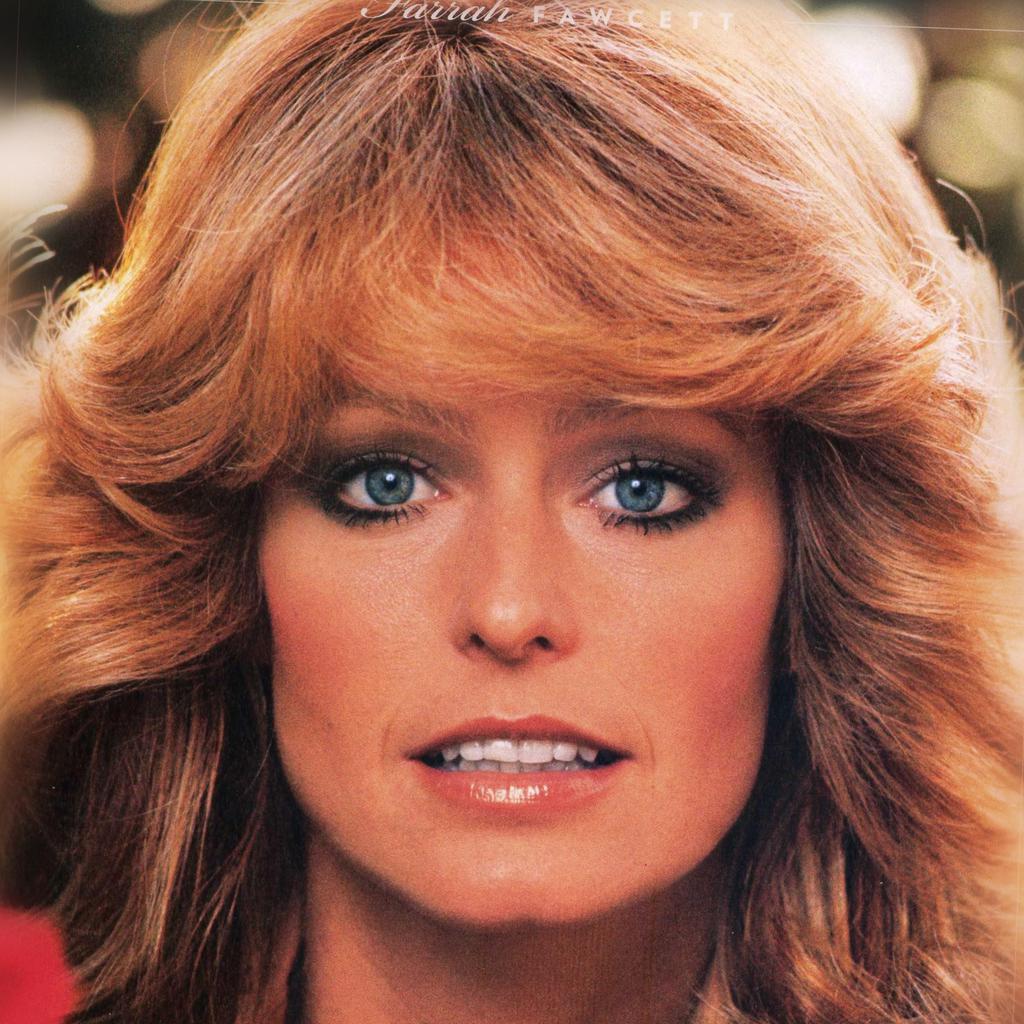} \\ \vspace{-9pt}
            \includegraphics[height=\linewidth]{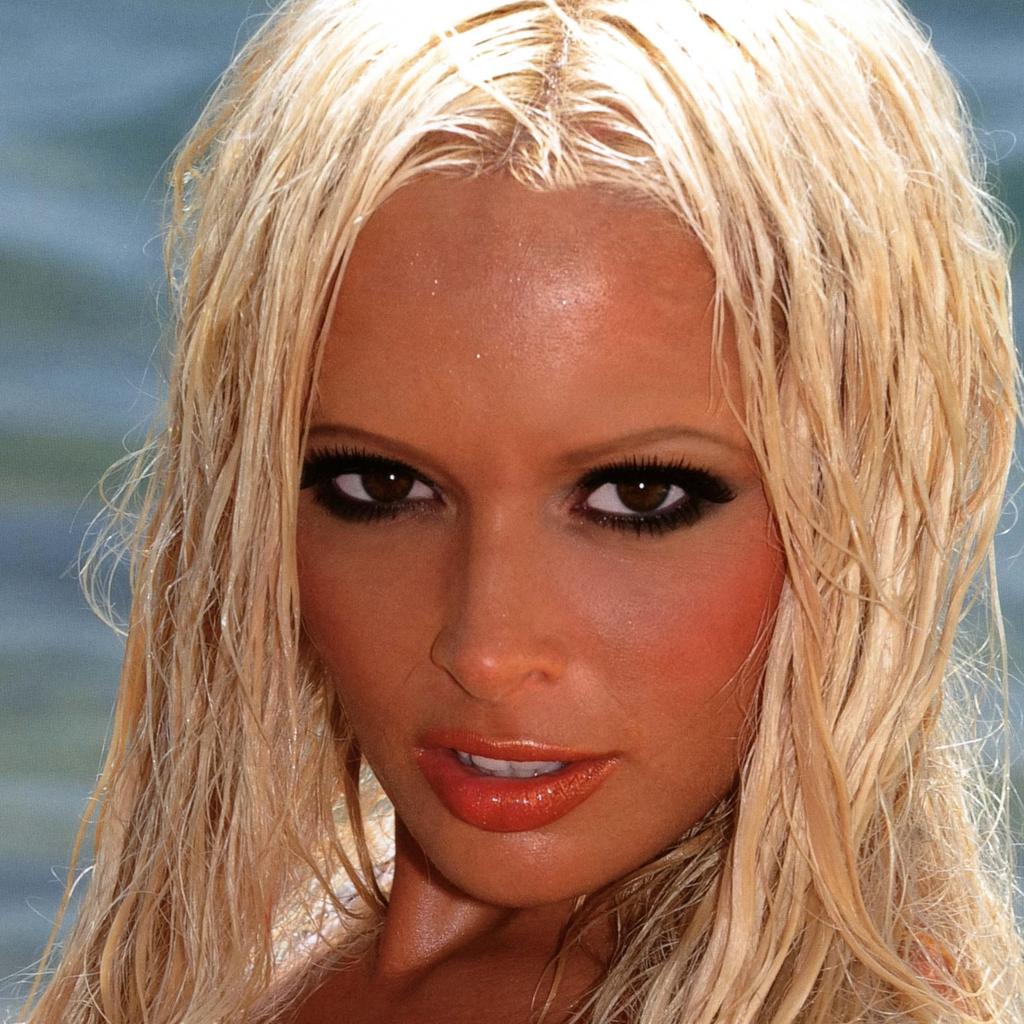} \\ \vspace{-12pt}
        \end{minipage}
    }
    \hspace{-8pt}
    \subfloat[w/o pair]{
        \begin{minipage}{0.115\linewidth}
            \includegraphics[height=\linewidth]{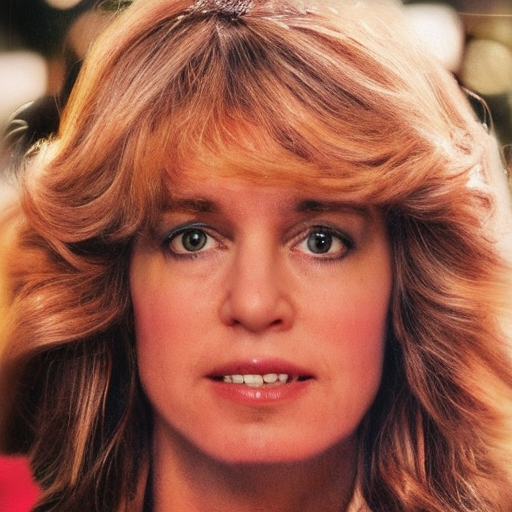} \\ \vspace{-9pt}
            \includegraphics[height=\linewidth]{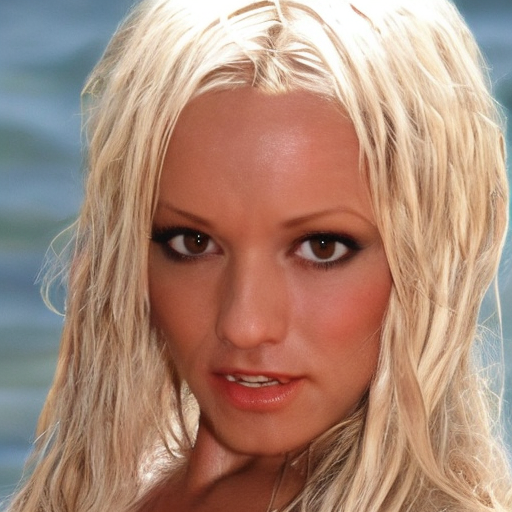} \\ \vspace{-12pt}
        \end{minipage}
    }
    \hspace{-8pt}
    \subfloat[w/o masking]{
        \begin{minipage}{0.115\linewidth}
            \includegraphics[height=\linewidth]{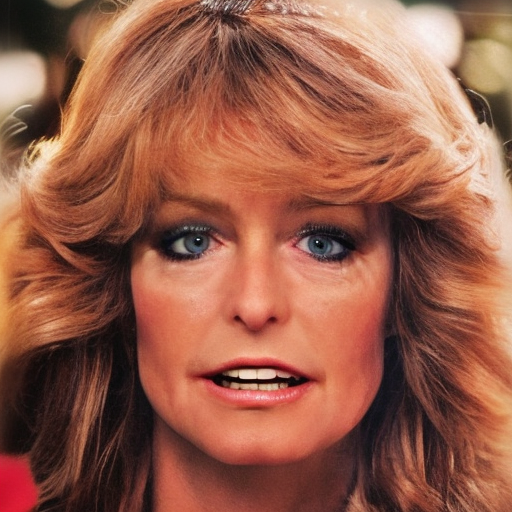} \\ \vspace{-9pt}
            \includegraphics[height=\linewidth]{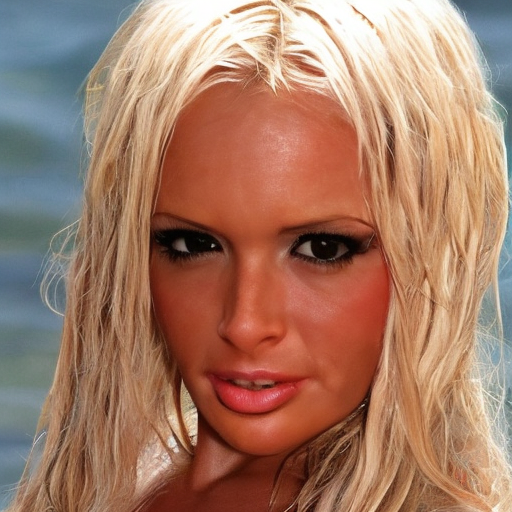} \\ \vspace{-12pt}
        \end{minipage}
    }
    \hspace{-8pt}
    \subfloat[w/o IGLH]{
        \begin{minipage}{0.115\linewidth}
            \includegraphics[height=\linewidth]{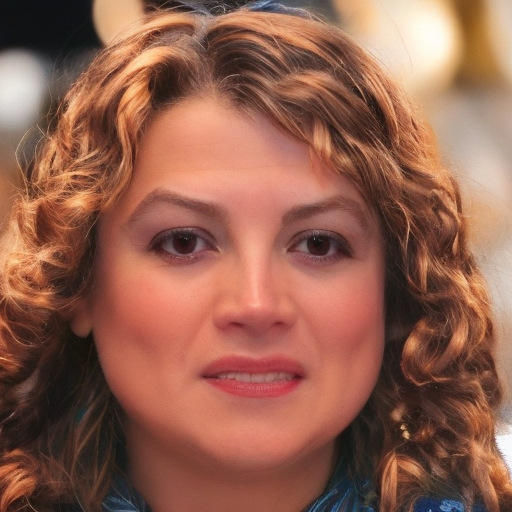} \\ \vspace{-9pt}
            \includegraphics[height=\linewidth]{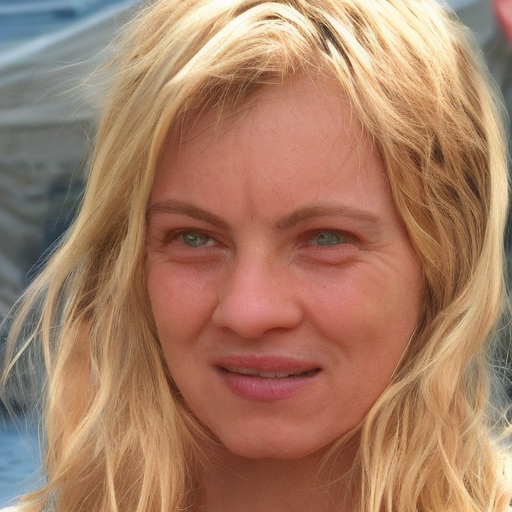} \\ \vspace{-12pt}
        \end{minipage}
    }
    \hspace{-8pt}
    \subfloat[w/o $\mathcal{L}_\text{diff-recon}$]{
        \begin{minipage}{0.115\linewidth}
            \includegraphics[height=\linewidth]{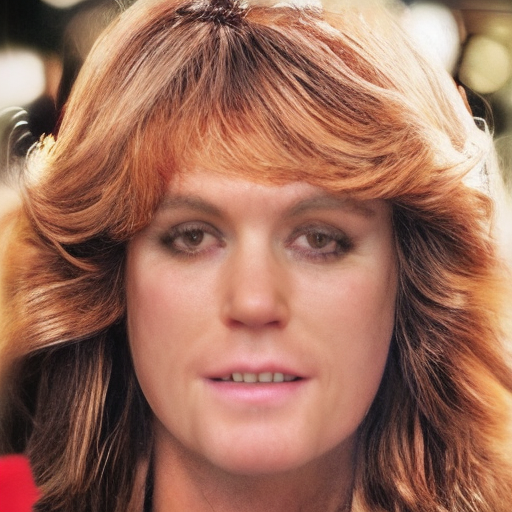} \\ \vspace{-9pt}
            \includegraphics[height=\linewidth]{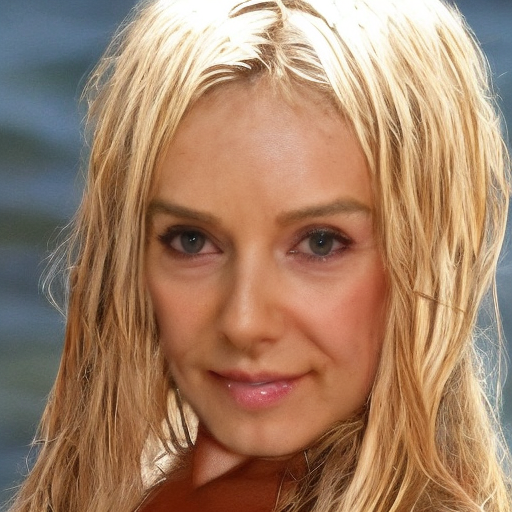} \\ \vspace{-12pt}
        \end{minipage}
    }
    \hspace{-8pt}
    \subfloat[w/o $\mathcal{L}_\text{id-region}$]{
        \begin{minipage}{0.115\linewidth}
            \includegraphics[height=\linewidth]{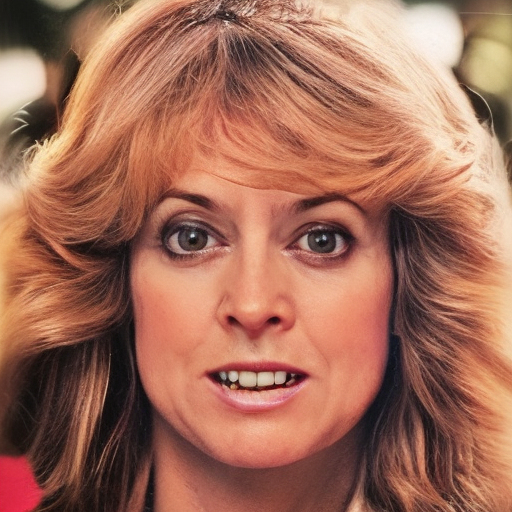} \\ \vspace{-9pt}
            \includegraphics[height=\linewidth]{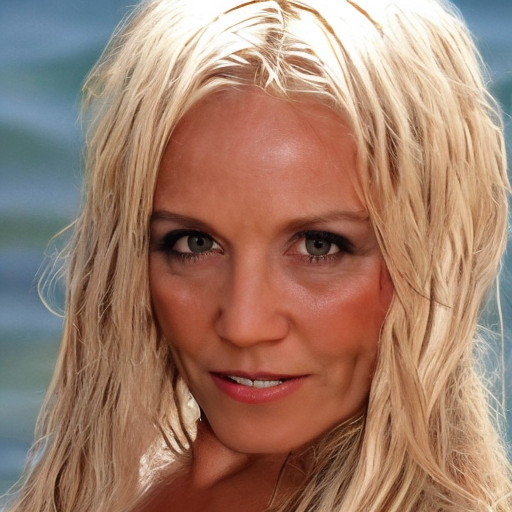} \\ \vspace{-12pt}
        \end{minipage}
    }
    \hspace{-8pt}
    \subfloat[w/o OIM]{
        \begin{minipage}{0.115\linewidth}
            \includegraphics[height=\linewidth]{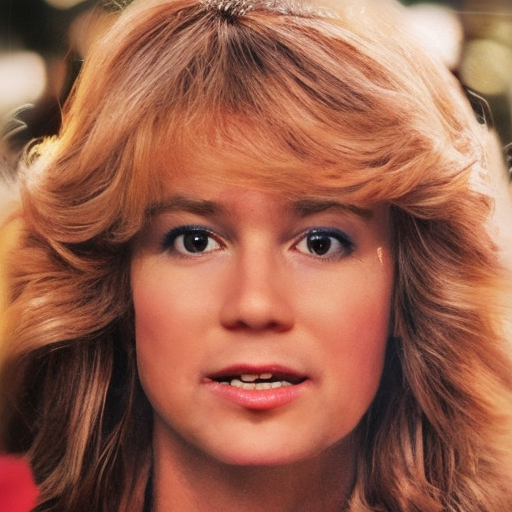} \\ \vspace{-9pt}
            \includegraphics[height=\linewidth]{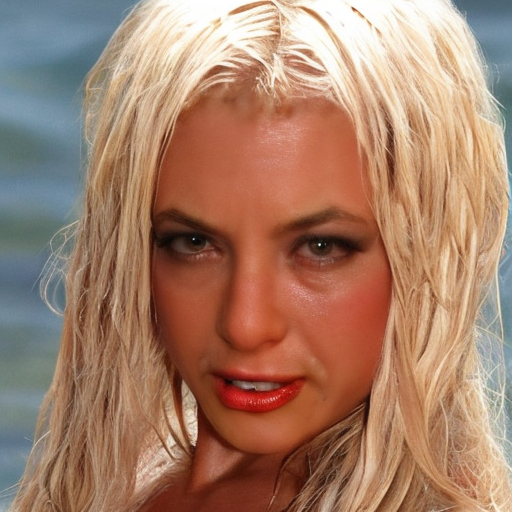} \\ \vspace{-12pt}
        \end{minipage}
    }
    \hspace{-8pt}
    \subfloat[Full (Ours)]{
        \begin{minipage}{0.115\linewidth}
            \includegraphics[height=\linewidth]{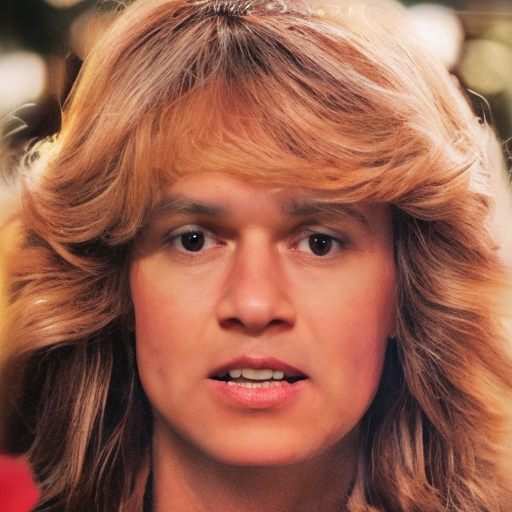} \\ \vspace{-9pt}
            \includegraphics[height=\linewidth]{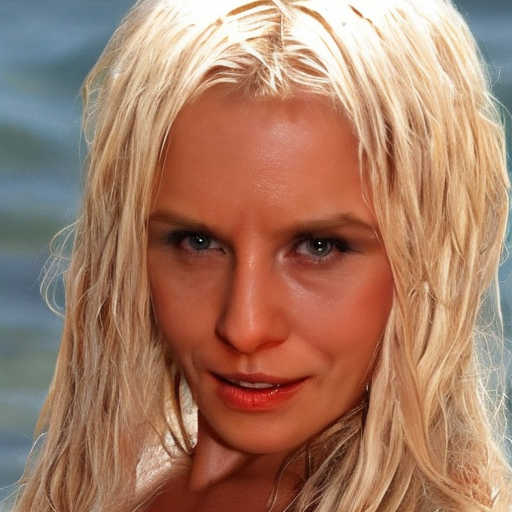} \\ \vspace{-12pt}
        \end{minipage}
    }
    \caption{
        Qualitative ablation study on the CelebA-HQ dataset.}
     \label{fig:ablation}
\end{figure*}

\begin{table}[t!]
    \centering
    \caption{Ablation study on CelebA-HQ dataset. (Retri.: Top-1 identity retrieval accuracy)}
    \resizebox{\linewidth}{!}{
    \renewcommand{\arraystretch}{1.2}
    \setlength{\tabcolsep}{2.5pt}
    \begin{tabular}{l|ccc|ccc|c}
    \bottomrule
    \multirow{2}{*}{Method}               & \multicolumn{3}{c|}{Identity}                       & \multicolumn{3}{c|}{Attribute}                      & Quality    \\ \cline{2-8} 
                                             & Retri.↓         & mAP↓            & Sim.↓           & LM.↓            & Pose↓           & Exp.↓           & MUSIQ↑           \\ \hline
    w/o pair                                 & 0.2151          & 0.1792          & 0.1937          & {\ul 5.4924}    & {\ul 2.9187}    & 0.2177          & 73.1141          \\
    w/o masking                          & 0.4120          & 0.4998          & 0.3711          & \textbf{5.4625} & \textbf{2.7995} & \textbf{0.1959} & 71.8064          \\
    w/o IGLH                                 & 0.0136          & 0.0351          & 0.0614          & 11.0791         & 7.5702          & 0.3175          & 72.2935          \\
    w/o $\mathcal{L}_\text{diff-recon}$      & {\ul 0.0007}    & {\ul 0.0005}    & {\ul 0.0046}    & 5.6690          & 4.2476          & 0.2635          & 65.6009          \\
    w/o $\mathcal{L}_\text{id-region}$       & 0.0131          & 0.0317          & 0.0594          & 5.8703          & 3.3385          & 0.2376          & 73.3418          \\
    w/o OIM                                  & 0.0125          & 0.0314          & 0.0523          & 5.7647          & 3.2252          & 0.2234          & {\ul 73.5466}    \\
    \rowcolor{gray!20}
    Full (Ours)                              & \textbf{0.0000} & \textbf{0.0003} & \textbf{0.0045} & 5.5728          & 3.2009          & {\ul 0.2166}    & \textbf{73.8899} \\ \toprule
    \end{tabular}
    }
    \label{tab:ablation}
    \end{table}
In this section, we conduct a comprehensive ablation analysis of the key components introduced in this paper. The ablation studies include the evaluation of our learning strategy, individual modules, loss function design, and the OIM strategy during inference. The corresponding visualization results are provided in Fig.~\ref{fig:ablation} and Table~\ref{tab:ablation}.

\subsubsection{Identity-masked Learning Strategy}
Our identity-masked learning strategy involves two key operations: 1) the construction of identity-matched pairs for training, and 2) the masking of facial regions in the images. As demonstrated in the figure and table (w/o pairing and w/o masking), the omission of these strategies leads to significantly poorer anonymization performance. While the model is able to retain identity-agnostic attributes with greater fidelity, it struggles with anonymization. Without the pairing strategy, the identity conditions correspond too closely to the original image, which diminishes the diversity and generalizability of the model’s identity control capabilities. Furthermore, without the facial masking strategy, the model tends to focus on learning the original identity features more easily, as it can rely on the multi-level CLIP features, which naturally preserve these attributes. These results underscore the importance of both strategies for improving the model's ability to effectively anonymize faces.

\subsubsection{Module Ablation}
In this ablation study, we focus on evaluating the impact of our proposed IGLH module. By replacing the IGLH with the original cross-attention module, where the non-identity token and identity token are concatenated and input as the key and value, we assess the effect of this modification. The results, shown in the figures and tables (denoted as ``w/o IGLH”), indicate that, without the IGLH module, the model struggles to preserve identity-agnostic features from the original image, thereby compromising the usability of the anonymized faces. This comparison validates the effectiveness of the IGLH module in facilitating the integration of both identity and non-identity features, thereby enhancing the anonymization quality while preserving identity-irrelevant attributes.

\subsubsection{Loss Function Ablation}
For the loss function ablation, we specifically evaluate the contributions of the $\mathcal{L}_\text{diff-recon}$ and $\mathcal{L}_\text{id-region}$ loss terms, as the other terms are fundamental to the model's operation. As seen in the results, removing the $\mathcal{L}_\text{diff-recon}$ loss leads to a significant degradation in image quality, both visually and in terms of the MUSIQ metric. This highlights the critical role of the $\mathcal{L}_\text{diff-recon}$ loss in ensuring high-quality image generation. On the other hand, omitting the $\mathcal{L}_\text{id-region}$ loss primarily affects the fidelity of the facial region. The absence of this loss term leads to unnatural results, as the model lacks explicit guidance in separating identity and non-identity regions. This confusion results in poor image fidelity in identity-related regions. Thus, the $\mathcal{L}_\text{id-region}$ loss proves essential for maintaining the quality and naturalness of the generated faces.

\subsubsection{OIM Strategy}
Finally, we examine the impact of our orthogonal identity sampling strategy during inference. As shown in the ``w/o OIM'' in the figures and tables, this strategy significantly influences the similarity of the anonymized face to the original identity. When the OIM strategy is applied, the anonymization effect improves without affecting identity-agnostic features, thereby enhancing the overall anonymization process. This further demonstrates the effectiveness of our proposed strategy in enhancing face anonymization while preserving non-identity features.




\section{Conclusion}

In this paper, we presented ID\textsuperscript{2}Face, a novel training-centric diffusion-based face anonymization approach that eliminates the need for inference-time intervention by disentangling identity and non-identity attributes directly during training. Through identity-masked diffusion learning, the model achieves explicit disentanglement of identity and non-identity representations. This is realized by the Identity-Decoupled Latent Recomposer (IDLR), which separates identity features via variational encoding and extracts non-identity information from intra-identity variation. The disentangled factors are then integrated by the Identity-Guided Latent Harmonizer (IGLH), which performs gated, spatially-aware fusion to preserve both structural and semantic consistency in the output.
To further enhance privacy protection, we introduce an Orthogonal Identity Mapping (OIM) strategy that ensures the sampled identity vectors remain orthogonal to the source identity, effectively suppressing residual identity leakage without compromising image quality.
Extensive experiments on CelebA-HQ and FFHQ demonstrate that our method achieves state-of-the-art anonymization performance, producing visually realistic outputs with superior identity removal and strong preservation of downstream-relevant attributes.

\bibliographystyle{IEEEtran}
\bibliography{refs}

\begin{thebibliography}{10}
\providecommand{\url}[1]{#1}
\csname url@samestyle\endcsname
\providecommand{\newblock}{\relax}
\providecommand{\bibinfo}[2]{#2}
\providecommand{\BIBentrySTDinterwordspacing}{\spaceskip=0pt\relax}
\providecommand{\BIBentryALTinterwordstretchfactor}{4}
\providecommand{\BIBentryALTinterwordspacing}{\spaceskip=\fontdimen2\font plus
\BIBentryALTinterwordstretchfactor\fontdimen3\font minus
  \fontdimen4\font\relax}
\providecommand{\BIBforeignlanguage}[2]{{%
\expandafter\ifx\csname l@#1\endcsname\relax
\typeout{** WARNING: IEEEtran.bst: No hyphenation pattern has been}%
\typeout{** loaded for the language `#1'. Using the pattern for}%
\typeout{** the default language instead.}%
\else
\language=\csname l@#1\endcsname
\fi
#2}}
\providecommand{\BIBdecl}{\relax}
\BIBdecl

\bibitem{neustaedter2006blur}
C.~Neustaedter, S.~Greenberg, and M.~Boyle, ``Blur filtration fails to preserve
  privacy for home-based video conferencing,'' \emph{ACM Transactions on
  Computer-Human Interaction (TOCHI)}, vol.~13, no.~1, pp. 1--36, 2006.

\bibitem{vishwamitra2017blur}
N.~Vishwamitra, B.~Knijnenburg, H.~Hu, Y.~P. Kelly~Caine \emph{et~al.}, ``Blur
  vs. block: Investigating the effectiveness of privacy-enhancing obfuscation
  for images,'' in \emph{IEEE Conf. Comput. Vis. Pattern Recog. Worksh.}, 2017,
  pp. 39--47.

\bibitem{hukkelaas2019deepprivacy}
H.~Hukkel{\aa}s, R.~Mester, and F.~Lindseth, ``Deepprivacy: A generative
  adversarial network for face anonymization,'' in \emph{International
  symposium on visual computing}, 2019, pp. 565--578.

\bibitem{maximov2020ciagan}
M.~Maximov, I.~Elezi, and L.~Leal-Taix{\'e}, ``Ciagan: Conditional identity
  anonymization generative adversarial networks,'' in \emph{IEEE Conf. Comput.
  Vis. Pattern Recog.}, 2020, pp. 5447--5456.

\bibitem{kuang2021effective}
Z.~Kuang, H.~Liu, J.~Yu, A.~Tian, L.~Wang, J.~Fan, and N.~Babaguchi,
  ``Effective de-identification generative adversarial network for face
  anonymization,'' in \emph{ACM Int. Conf. Multimedia}, 2021, pp. 3182--3191.

\bibitem{RiDDLE}
D.~Li, W.~Wang, K.~Zhao, J.~Dong, and T.~Tan, ``Riddle: Reversible and
  diversified de-identification with latent encryptor,'' in \emph{IEEE Conf.
  Comput. Vis. Pattern Recog.}, 2023, pp. 8093--8102.

\bibitem{yang2024g}
H.~Yang, X.~Xu, C.~Xu, H.~Zhang, J.~Qin, Y.~Wang, P.-A. Heng, and S.~He,
  ``G2face: High-fidelity reversible face anonymization via generative and
  geometric priors,'' \emph{IEEE Trans. Inf. Forensics Secur.}, 2024.

\bibitem{AIDPro}
T.~Wang, W.~Wen, X.~Xiao, Z.~Hua, Y.~Zhang, and Y.~Fang, ``Beyond privacy:
  Generating privacy-preserving faces supporting robust image authentication,''
  \emph{IEEE Trans. Inf. Forensics Secur.}, vol.~20, pp. 2564--2576, 2025.

\bibitem{barattin2023attribute}
S.~Barattin, C.~Tzelepis, I.~Patras, and N.~Sebe, ``Attribute-preserving face
  dataset anonymization via latent code optimization,'' in \emph{IEEE Conf.
  Comput. Vis. Pattern Recog.}, 2023, pp. 8001--8010.

\bibitem{laishram2025toward}
L.~Laishram, M.~Shaheryar, J.~T. Lee, and S.~K. Jung, ``Toward a
  privacy-preserving face recognition system: A survey of leakages and
  solutions,'' \emph{ACM Computing Surveys}, vol.~57, no.~6, pp. 1--38, 2025.

\bibitem{proencca2021uu}
H.~Proen{\c{c}}a, ``The uu-net: Reversible face de-identification for visual
  surveillance video footage,'' \emph{IEEE Trans. Circuit Syst. Video
  Technol.}, vol.~32, no.~2, pp. 496--509, 2021.

\bibitem{ye2024securereid}
M.~Ye, W.~Shen, J.~Zhang, Y.~Yang, and B.~Du, ``Securereid: Privacy-preserving
  anonymization for person re-identification,'' \emph{IEEE Trans. Inf.
  Forensics Secur.}, vol.~19, pp. 2840--2853, 2024.

\bibitem{ciftci2023my}
U.~A. Ciftci, G.~Yuksek, and I.~Demir, ``My face my choice: Privacy enhancing
  deepfakes for social media anonymization,'' in \emph{IEEE/CVF Winter
  Conference on Applications of Computer Vision}, 2023, pp. 1369--1379.

\bibitem{DDPM}
J.~Ho, A.~Jain, and P.~Abbeel, ``Denoising diffusion probabilistic models,''
  \emph{Adv. Neural Inform. Process. Syst.}, vol.~33, pp. 6840--6851, 2020.

\bibitem{DDIM}
J.~Song, C.~Meng, and S.~Ermon, ``Denoising diffusion implicit models,''
  \emph{arXiv preprint arXiv:2010.02502}, 2020.

\bibitem{ldm}
R.~Rombach, A.~Blattmann, D.~Lorenz, P.~Esser, and B.~Ommer, ``High-resolution
  image synthesis with latent diffusion models,'' in \emph{IEEE Conf. Comput.
  Vis. Pattern Recog.}, 2022, pp. 10\,684--10\,695.

\bibitem{DiffPrivacy}
X.~He, M.~Zhu, D.~Chen, N.~Wang, and X.~Gao, ``Diff-privacy: Diffusion-based
  face privacy protection,'' \emph{IEEE Trans. Circuit Syst. Video Technol.},
  2024.

\bibitem{FAMS}
H.-W. Kung, T.~Varanka, S.~Saha, T.~Sim, and N.~Sebe, ``Face anonymization made
  simple,'' in \emph{IEEE/CVF Winter Conference on Applications of Computer
  Vision}, 2025, pp. 1040--1050.

\bibitem{NullFace}
H.-W. Kung, T.~Varanka, T.~Sim, and N.~Sebe, ``Nullface: Training-free
  localized face anonymization,'' \emph{arXiv preprint arXiv:2503.08478}, 2025.

\bibitem{goodfellow2020generative}
I.~Goodfellow, J.~Pouget-Abadie, M.~Mirza, B.~Xu, D.~Warde-Farley, S.~Ozair,
  A.~Courville, and Y.~Bengio, ``Generative adversarial networks,''
  \emph{Communications of the ACM}, vol.~63, no.~11, pp. 139--144, 2020.

\bibitem{mou2024t2i}
C.~Mou, X.~Wang, L.~Xie, Y.~Wu, J.~Zhang, Z.~Qi, and Y.~Shan, ``T2i-adapter:
  Learning adapters to dig out more controllable ability for text-to-image
  diffusion models,'' in \emph{AAAI Conf. Artif. Intell.}, vol.~38, no.~5,
  2024, pp. 4296--4304.

\bibitem{yu2024beyond}
Y.~Yu, B.~Liu, C.~Zheng, X.~Xu, H.~Zhang, and S.~He, ``Beyond textual
  constraints: Learning novel diffusion conditions with fewer examples,'' in
  \emph{IEEE Conf. Comput. Vis. Pattern Recog.}, 2024, pp. 7109--7118.

\bibitem{kang2025sita}
J.~Kang, H.~Yang, Y.~Cai, H.~Zhang, X.~Xu, Y.~Du, and S.~He, ``Sita:
  Structurally imperceptible and transferable adversarial attacks for stylized
  image generation,'' \emph{IEEE Trans. Inf. Forensics Secur.}, 2025.

\bibitem{gal2022image}
R.~Gal, Y.~Alaluf, Y.~Atzmon, O.~Patashnik, A.~H. Bermano, G.~Chechik, and
  D.~Cohen-Or, ``An image is worth one word: Personalizing text-to-image
  generation using textual inversion,'' \emph{arXiv preprint arXiv:2208.01618},
  2022.

\bibitem{ruiz2023dreambooth}
N.~Ruiz, Y.~Li, V.~Jampani, Y.~Pritch, M.~Rubinstein, and K.~Aberman,
  ``Dreambooth: Fine tuning text-to-image diffusion models for subject-driven
  generation,'' in \emph{IEEE Conf. Comput. Vis. Pattern Recog.}, 2023, pp.
  22\,500--22\,510.

\bibitem{peng2024portraitbooth}
X.~Peng, J.~Zhu, B.~Jiang, Y.~Tai, D.~Luo, J.~Zhang, W.~Lin, T.~Jin, C.~Wang,
  and R.~Ji, ``Portraitbooth: A versatile portrait model for fast
  identity-preserved personalization,'' in \emph{IEEE Conf. Comput. Vis.
  Pattern Recog.}, 2024, pp. 27\,080--27\,090.

\bibitem{ye2023ip}
H.~Ye, J.~Zhang, S.~Liu, X.~Han, and W.~Yang, ``Ip-adapter: Text compatible
  image prompt adapter for text-to-image diffusion models,'' \emph{arXiv
  preprint arXiv:2308.06721}, 2023.

\bibitem{kawar2023imagic}
B.~Kawar, S.~Zada, O.~Lang, O.~Tov, H.~Chang, T.~Dekel, I.~Mosseri, and
  M.~Irani, ``Imagic: Text-based real image editing with diffusion models,'' in
  \emph{IEEE Conf. Comput. Vis. Pattern Recog.}, 2023, pp. 6007--6017.

\bibitem{huang2025diffusion}
Y.~Huang, J.~Huang, Y.~Liu, M.~Yan, J.~Lv, J.~Liu, W.~Xiong, H.~Zhang, L.~Cao,
  and S.~Chen, ``Diffusion model-based image editing: A survey,'' \emph{IEEE
  Trans. Pattern Anal. Mach. Intell.}, 2025.

\bibitem{wen2023divide}
Y.~Wen, B.~Liu, J.~Cao, R.~Xie, and L.~Song, ``Divide and conquer: a two-step
  method for high quality face de-identification with model explainability,''
  in \emph{Int. Conf. Comput. Vis.}, 2023, pp. 5148--5157.

\bibitem{yuan2022pro}
L.~Yuan, L.~Liu, X.~Pu, Z.~Li, H.~Li, and X.~Gao, ``Pro-face: A generic
  framework for privacy-preserving recognizable obfuscation of face images,''
  in \emph{ACM Int. Conf. Multimedia}, 2022, pp. 1661--1669.

\bibitem{yuan2024pro}
L.~Yuan, W.~Chen, X.~Pu, Y.~Zhang, H.~Li, Y.~Zhang, X.~Gao, and T.~Ebrahimi,
  ``Pro-face c: Privacy-preserving recognition of obfuscated face via feature
  compensation,'' \emph{IEEE Trans. Inf. Forensics Secur.}, 2024.

\bibitem{karras2019style}
T.~Karras, S.~Laine, and T.~Aila, ``A style-based generator architecture for
  generative adversarial networks,'' in \emph{IEEE Conf. Comput. Vis. Pattern
  Recog.}, 2019, pp. 4401--4410.

\bibitem{zheng2022general}
Y.~Zheng, H.~Yang, T.~Zhang, J.~Bao, D.~Chen, Y.~Huang, L.~Yuan, D.~Chen,
  M.~Zeng, and F.~Wen, ``General facial representation learning in a
  visual-linguistic manner,'' in \emph{IEEE Conf. Comput. Vis. Pattern Recog.},
  2022, pp. 18\,697--18\,709.

\bibitem{wang2021towards}
X.~Wang, Y.~Li, H.~Zhang, and Y.~Shan, ``Towards real-world blind face
  restoration with generative facial prior,'' in \emph{IEEE Conf. Comput. Vis.
  Pattern Recog.}, 2021, pp. 9168--9178.

\bibitem{radford2021learning}
A.~Radford, J.~W. Kim, C.~Hallacy, A.~Ramesh, G.~Goh, S.~Agarwal, G.~Sastry,
  A.~Askell, P.~Mishkin, J.~Clark \emph{et~al.}, ``Learning transferable visual
  models from natural language supervision,'' in \emph{Int. Conf. Mach.
  Learn.}\hskip 1em plus 0.5em minus 0.4em\relax PMLR, 2021, pp. 8748--8763.

\bibitem{liu2022convnet}
Z.~Liu, H.~Mao, C.-Y. Wu, C.~Feichtenhofer, T.~Darrell, and S.~Xie, ``A convnet
  for the 2020s,'' in \emph{IEEE Conf. Comput. Vis. Pattern Recog.}, 2022, pp.
  11\,976--11\,986.

\bibitem{li2023gligen}
Y.~Li, H.~Liu, Q.~Wu, F.~Mu, J.~Yang, J.~Gao, C.~Li, and Y.~J. Lee, ``Gligen:
  Open-set grounded text-to-image generation,'' in \emph{IEEE Conf. Comput.
  Vis. Pattern Recog.}, 2023, pp. 22\,511--22\,521.

\bibitem{deng2019arcface}
J.~Deng, J.~Guo, N.~Xue, and S.~Zafeiriou, ``Arcface: Additive angular margin
  loss for deep face recognition,'' in \emph{IEEE Conf. Comput. Vis. Pattern
  Recog.}, 2019, pp. 4690--4699.

\bibitem{kingma2013auto}
D.~P. Kingma and M.~Welling, ``Auto-encoding variational bayes,'' \emph{arXiv
  preprint arXiv:1312.6114}, 2013.

\bibitem{kingma2014adam}
D.~P. Kingma and J.~Ba, ``Adam: A method for stochastic optimization,''
  \emph{arXiv preprint arXiv:1412.6980}, 2014.

\bibitem{chen2023simswap}
X.~Chen, B.~Ni, Y.~Liu, N.~Liu, Z.~Zeng, and H.~Wang, ``Simswap++: Towards
  faster and high-quality identity swapping,'' \emph{IEEE Trans. Pattern Anal.
  Mach. Intell.}, vol.~46, no.~1, pp. 576--592, 2023.

\bibitem{FFHQ}
T.~Karras, S.~Laine, and T.~Aila, ``A style-based generator architecture for
  generative adversarial networks,'' in \emph{IEEE Conf. Comput. Vis. Pattern
  Recog.}, 2019, pp. 4401--4410.

\bibitem{CelebAHQ}
T.~Karras, T.~Aila, S.~Laine, and J.~Lehtinen, ``Progressive growing of {GAN}s
  for improved quality, stability, and variation,'' in \emph{Int. Conf. Learn.
  Represent.}, 2018.

\bibitem{kim2022adaface}
M.~Kim, A.~K. Jain, and X.~Liu, ``Adaface: Quality adaptive margin for face
  recognition,'' in \emph{IEEE Conf. Comput. Vis. Pattern Recog.}, 2022, pp.
  18\,750--18\,759.

\bibitem{dan2024topofr}
J.~Dan, Y.~Liu, J.~Deng, H.~Xie, S.~Li, B.~Sun, and S.~Luo, ``Topofr: A closer
  look at topology alignment on face recognition,'' \emph{Adv. Neural Inform.
  Process. Syst.}, vol.~37, pp. 37\,213--37\,240, 2024.

\bibitem{zhang2016joint}
K.~Zhang, Z.~Zhang, Z.~Li, and Y.~Qiao, ``Joint face detection and alignment
  using multitask cascaded convolutional networks,'' \emph{IEEE Sign. Process.
  Letters}, vol.~23, no.~10, pp. 1499--1503, 2016.

\bibitem{ruiz2018fine}
N.~Ruiz, E.~Chong, and J.~M. Rehg, ``Fine-grained head pose estimation without
  keypoints,'' in \emph{IEEE Conf. Comput. Vis. Pattern Recog. Worksh.}, 2018,
  pp. 2074--2083.

\bibitem{vemulapalli2019compact}
R.~Vemulapalli and A.~Agarwala, ``A compact embedding for facial expression
  similarity,'' in \emph{IEEE Conf. Comput. Vis. Pattern Recog.}, 2019, pp.
  5683--5692.

\bibitem{abdelrahman2023l2cs}
A.~A. Abdelrahman, T.~Hempel, A.~Khalifa, A.~Al-Hamadi, and L.~Dinges,
  ``L2cs-net: Fine-grained gaze estimation in unconstrained environments,'' in
  \emph{2023 8th International Conference on Frontiers of Signal Processing
  (ICFSP)}, 2023, pp. 98--102.

\bibitem{ke2021musiq}
J.~Ke, Q.~Wang, Y.~Wang, P.~Milanfar, and F.~Yang, ``Musiq: Multi-scale image
  quality transformer,'' in \emph{Int. Conf. Comput. Vis.}, 2021, pp.
  5148--5157.

\bibitem{heusel2017gans}
M.~Heusel, H.~Ramsauer, T.~Unterthiner, B.~Nessler, and S.~Hochreiter, ``Gans
  trained by a two time-scale update rule converge to a local nash
  equilibrium,'' \emph{Adv. Neural Inform. Process. Syst.}, vol.~30, 2017.

\bibitem{van2008visualizing}
L.~v.~d. Maaten and G.~Hinton, ``Visualizing data using t-sne,'' \emph{Journal
  of machine learning research}, vol.~9, no. Nov, pp. 2579--2605, 2008.

\bibitem{songscore}
Y.~Song, J.~Sohl-Dickstein, D.~P. Kingma, A.~Kumar, S.~Ermon, and B.~Poole,
  ``Score-based generative modeling through stochastic differential
  equations,'' in \emph{Int. Conf. Learn. Represent.}, 2021.

\bibitem{lu2022dpm}
C.~Lu, Y.~Zhou, F.~Bao, J.~Chen, C.~Li, and J.~Zhu, ``Dpm-solver: A fast ode
  solver for diffusion probabilistic model sampling in around 10 steps,''
  \emph{Adv. Neural Inform. Process. Syst.}, vol.~35, pp. 5775--5787, 2022.

\bibitem{sauer2024adversarial}
A.~Sauer, D.~Lorenz, A.~Blattmann, and R.~Rombach, ``Adversarial diffusion
  distillation,'' in \emph{Eur. Conf. Comput. Vis.}\hskip 1em plus 0.5em minus
  0.4em\relax Springer, 2024, pp. 87--103.

\bibitem{yin2024one}
T.~Yin, M.~Gharbi, R.~Zhang, E.~Shechtman, F.~Durand, W.~T. Freeman, and
  T.~Park, ``One-step diffusion with distribution matching distillation,'' in
  \emph{IEEE Conf. Comput. Vis. Pattern Recog.}, 2024, pp. 6613--6623.

\end{thebibliography}


\begin{IEEEbiography}[{\includegraphics[width=1in,height=1.25in,clip,keepaspectratio]{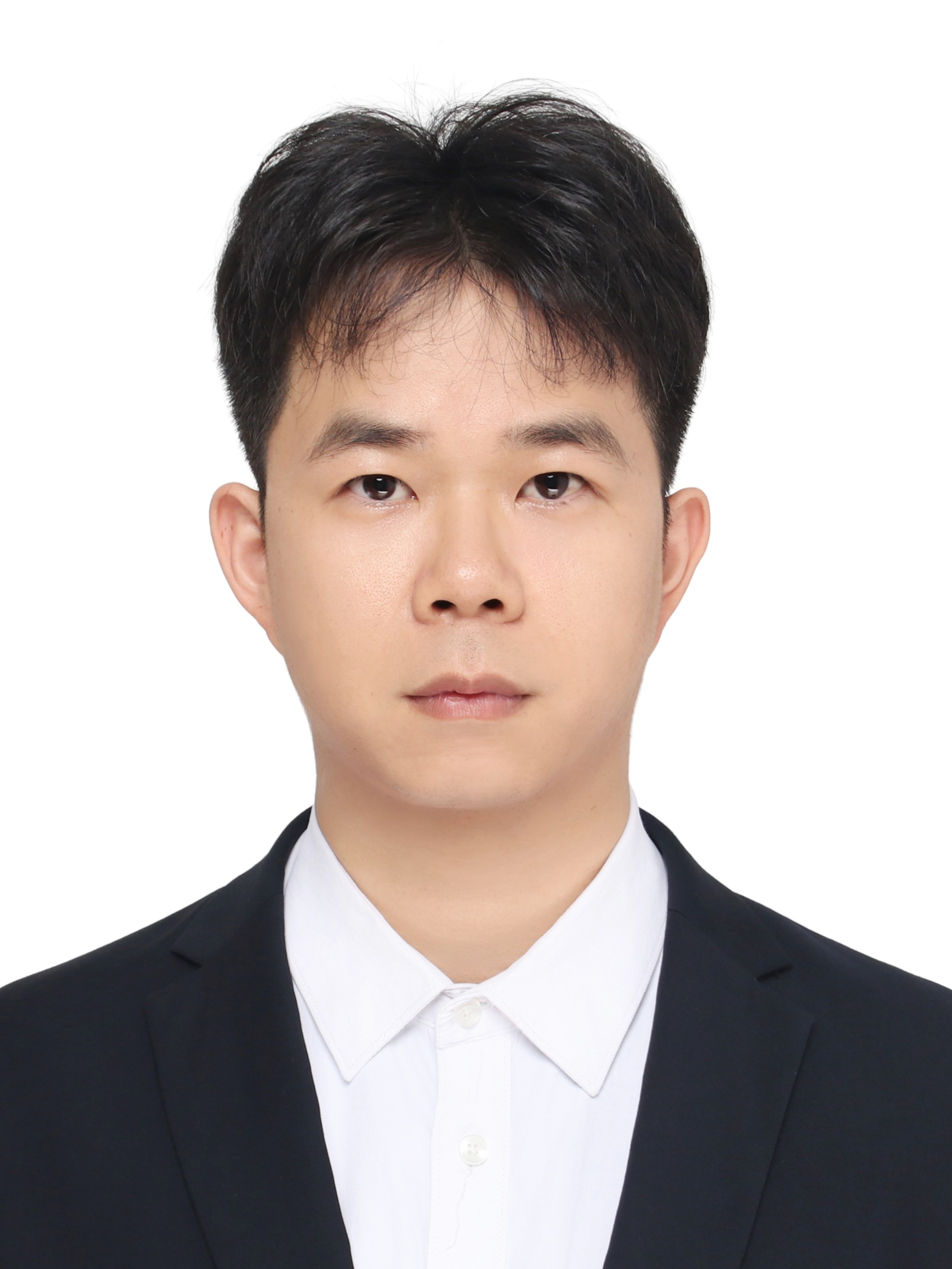}}]
    {Haoxin Yang} is a Ph.D. student at the School of Computer Science \& Engineering, South China University of Technology. He obtained his B.Sc. and M.Sc. degrees from South China Agricultural University and Shenzhen University in 2019 and 2022, respectively. His research interests include privacy and security in computer vision and AIGC.
   \end{IEEEbiography}
   
   \begin{IEEEbiography}[{\includegraphics[width=1in,height=1.25in,clip,keepaspectratio]{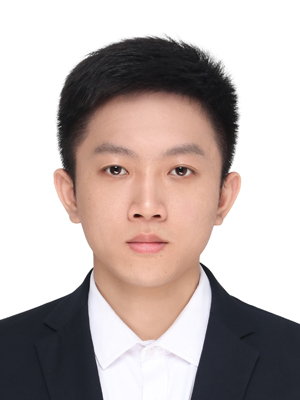}}]
    {Yihong Lin} is currently working toward the Ph.D degree with the School of Computer Science and Engineering, South China University of Technology. His research interests include 3D reconstruction and generation.
     \end{IEEEbiography}
   
   \begin{IEEEbiography}[{\includegraphics[width=1in,height=1.25in,clip,keepaspectratio]{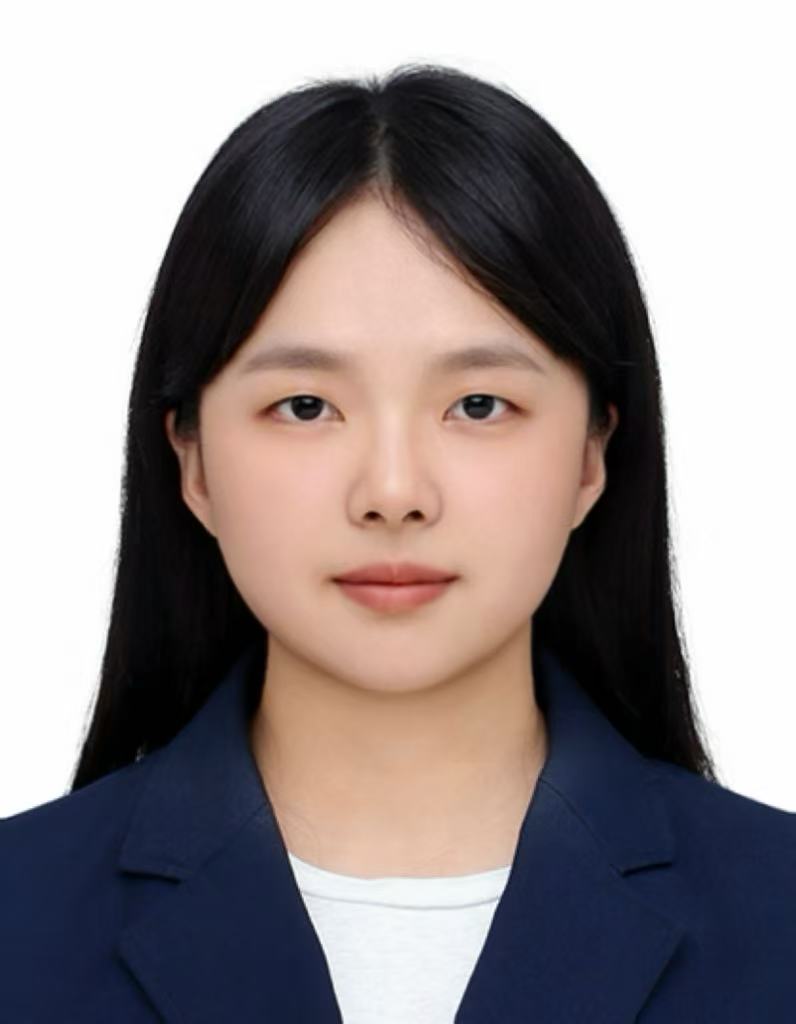}}]
    {Jingdan Kang} is a graduate student at the School of Future Technology, South China University of Technology. She received her B.Sc. degree from South China University of Technology in 2023. Her research interests include privacy and security in computer vision.
   \end{IEEEbiography}
   
   \begin{IEEEbiography}[{\includegraphics[width=1in,height=1.25in,clip,keepaspectratio]{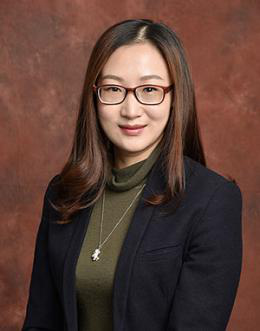}}]
    {Xuemiao Xu} received her B.S. and M.S. degrees in Computer Science and Engineering from South China University of Technology in 2002 and 2005, respectively, and her Ph.D. degree in Computer Science and Engineering from The Chinese University of Hong Kong in 2009. She is currently a professor in the School of Computer Science and Engineering at South China University of Technology. Her research interests include object detection, tracking, recognition, understanding, and synthesis of images and videos, particularly their applications in intelligent transportation.
   \end{IEEEbiography}

   \begin{IEEEbiography}[{\includegraphics[width=1in,height=1.25in,clip,keepaspectratio]{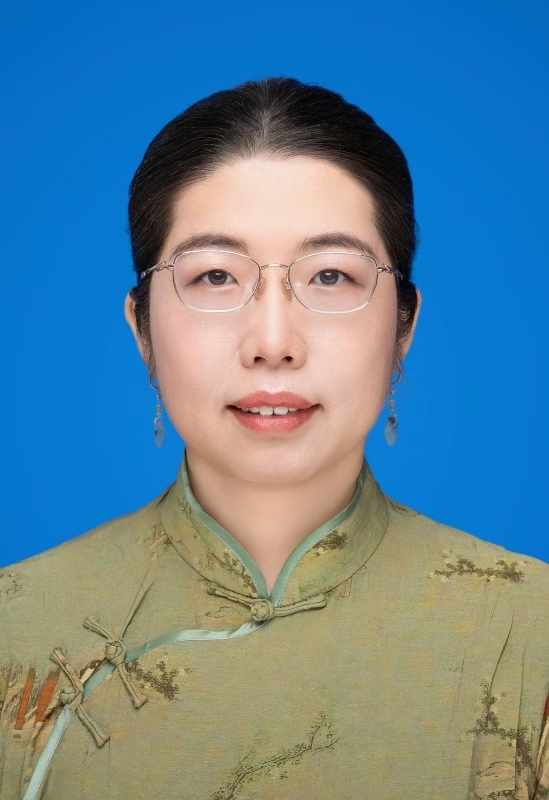}}]
   {Yue Li} is currently an Associate Professor in the School of Computer Science and Engineering at South China University of Technology. She received her B.S. and M.S. degrees from South China University of Technology, and her Ph.D. degree from Tsinghua University. Her research interests include artificial intelligence, data mining, and computer science popularization.
    \end{IEEEbiography}

   \begin{IEEEbiography}[{\includegraphics[width=1in,height=1.25in,clip,keepaspectratio]{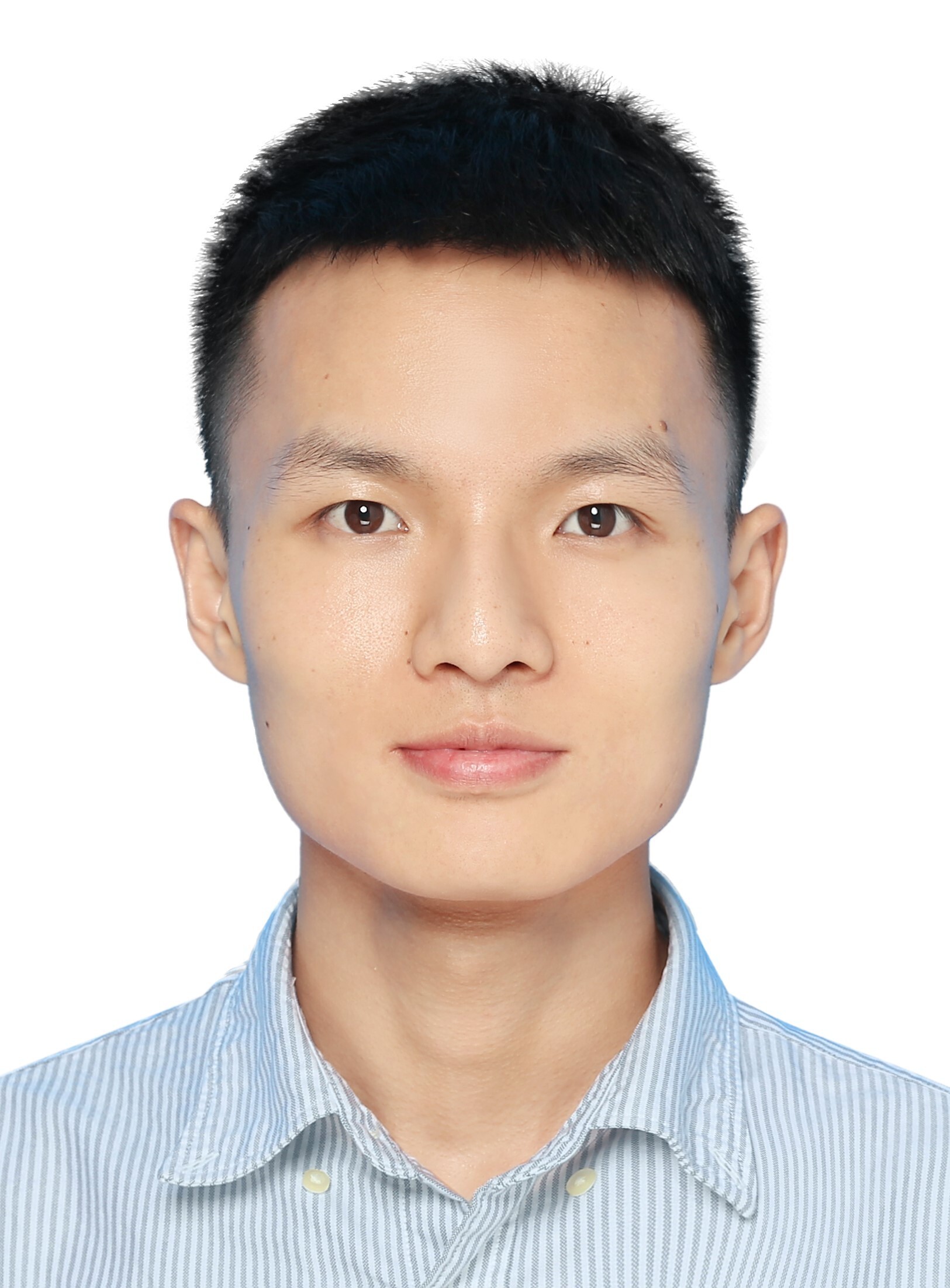}}]
    {Cheng Xu} received his Ph.D. in Computer Science and Technology from South China University of Technology in 2023. He is currently a Research Scientist at Singapore Management University. Prior to this, he was a Post-Doctoral Fellow at The Hong Kong Polytechnic University from 2023 to 2025. His research interests primarily include human-centric AIGC and medical image analysis.
   \end{IEEEbiography}
   
   \begin{IEEEbiography}[{\includegraphics[width=1in,height=1.25in,clip,keepaspectratio]{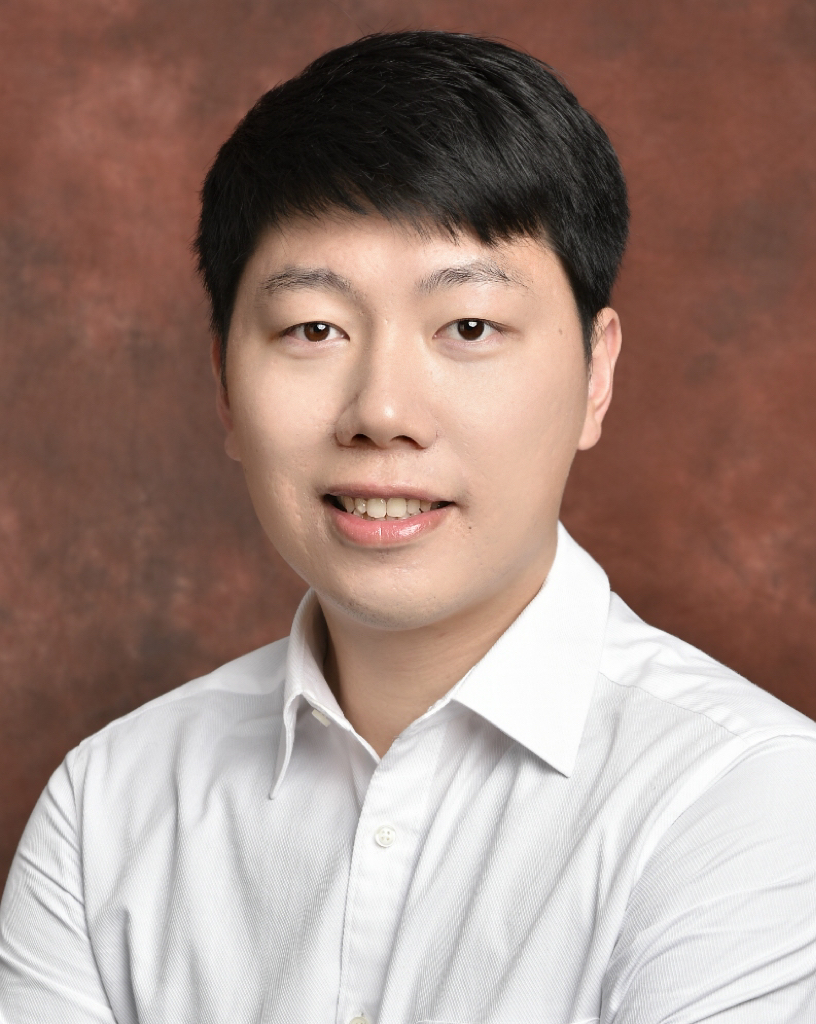}}]{Shengfeng He (Senior Member, IEEE)} is an associate professor in the School of Computing and Information Systems at Singapore Management University. He was a faculty member at South China University of Technology (2016–2022). He earned his B.Sc. and M.Sc. from Macau University of Science and Technology (2009, 2011) and a Ph.D. from City University of Hong Kong (2015). His research focuses on computer vision and generative models. He has received awards including the Google Research Award, PerCom 2024 Best Paper Award, and the Lee Kong Chian Fellowship. He is a senior IEEE member and a distinguished CCF member. He serves as lead guest editor for IJCV and associate editor for IEEE TPAMI, IEEE TNNLS, IEEE TCSVT, Visual Intelligence, and Neurocomputing. He is an area chair/senior PC member for CVPR, NeurIPS, ICLR, ICML, AAAI, IJCAI, BMVC, and the Conference Chair of Pacific Graphics 2026.\end{IEEEbiography} 
\newpage
\begin{strip}
\centering\huge{$\star$--------------------Supplementary Material--------------------$\star$}
\vspace{30pt}
\end{strip}

\section{Analysis of Reconstruction and Anonymization Conflicts in Diffusion Models}

Diffusion models are fundamentally designed to reconstruct data with high fidelity, which inevitably preserves identity information. In contrast, anonymization requires the suppression of identity while retaining identity-irrelevant attributes. This inherent objective mismatch poses a significant challenge when directly incorporating anonymization objectives into the diffusion process to construct an effective anonymization framework. In this section, we provide a theoretical analysis of this conflict.
\subsubsection{Identity Subspace Decomposition}  
We first represent any face image $X$ as a composition of two orthogonal components:  
\begin{equation} 
X = P_{s}X + P_{u}X, 
\end{equation}  
where $P_{s}$ projects onto the \emph{identity subspace}, and $P_{u} = \textbf{\text{I}} - P_{s}$ projects onto the complementary \emph{utility subspace} (e.g., pose, expression, background). This decomposition provides a conceptual basis for disentangling identity and utility factors.  

\subsubsection{Diffusion Denoising Objective}  
The denoising diffusion probabilistic model (DDPM)~\cite{DDPM} is trained to predict Gaussian noise injected into $x$:  
\begin{equation} 
\mathcal{L}_{\text{diff}}(\theta) = \mathbb{E}_{x, t, \epsilon} \left[ \| \epsilon - \epsilon_\theta(x_t, t) \|^2 \right], 
\end{equation}  
where $x_t = \sqrt{\bar\alpha_t}x + \sqrt{1-\bar\alpha_t}\epsilon$, $\alpha_t$ denotes the variance schedule, $t$ is the time step, and $\epsilon \sim \mathcal{N}(0, \textbf{I})$. This objective is equivalent~\cite{DDPM,songscore} to minimizing a reconstruction loss (ignoring time-dependent weights for clarity):  
\begin{equation} 
\mathcal{L}_{\text{diff}}(\theta) = \| x - \hat x \|^2. 
\end{equation}  

Under the subspace decomposition, optimizing $\mathcal{L}_{\text{diff}}$ simultaneously minimizes reconstruction error in both subspaces:  
\begin{equation} 
\min \mathcal{L}_{\text{diff}}(\theta) \iff \min \| P_{s}(x - \hat x) \|^2 + \| P_{u}(x - \hat x) \|^2. 
\end{equation}  
Thus, diffusion training inherently drives $\hat x$ toward $x$ along both identity and utility directions. In practice, however, identity features are highly discriminative and therefore disproportionately reinforced, leading to strong identity preservation.  

\subsubsection{Anonymization Objective}  
Anonymization, by contrast, seeks to suppress identity while preserving utility. A typical formulation is:  
\begin{equation} 
\mathcal{L}_{\rm anony}(\theta) = \mathbb{E}\left[d(P_{u}x, P_{u}\hat x)\right] + \lambda I(\hat x; c), 
\end{equation}  
where $d(\cdot,\cdot)$ measures distortion in the utility subspace (ensuring $P_u \hat{x} \approx P_u x$), $I(\hat x;c)$ is the mutual information between the anonymized output $\hat x$ and the identity label $c$ of $x$, and $\lambda$ balances the two terms. Since $c$ is fully determined by $P_s x$, minimizing $I(\hat x;c)$ requires decorrelating $P_s \hat{x}$ from $P_s x$:  
\begin{equation} 
\min I(\hat x; c) \ \Rightarrow \ P_s \hat{x} \ \bot \ P_s x. 
\end{equation}  
Hence, anonymization explicitly enforces deviation from the original identity, in direct opposition to diffusion’s reconstruction-driven preservation.  

\begin{figure}[t!]
    \centering
    \includegraphics[width=\linewidth]{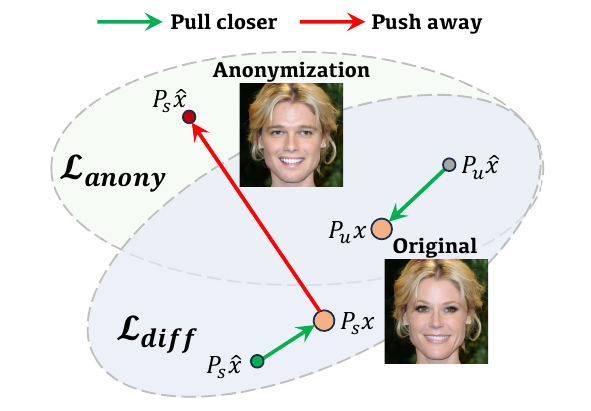}
    \caption{Optimization directions of $\mathcal{L}_\text{diff}$ and $\mathcal{L}_\text{anony}$ in the identity ($P_s$) and utility ($P_u$) subspaces. While $\mathcal{L}_\text{diff}$ pulls the generated sample toward the input $x$ in the identity subspace (the green $P_s \hat{x}$), $\mathcal{L}_\text{anony}$ pushes it away (the red $P_s \hat{x}$), leading to conflicting optimization directions.}
    \label{fig:conflict}
    \end{figure}

\subsubsection{Conflict in the Identity Subspace}  
As illustrated in Fig.~\ref{fig:conflict}, the two objectives are inherently antagonistic:  
\begin{align*} 
\min \mathcal{L}_{\text{diff}} &\supset P_{s}(x - \hat x) && \text{(identity preservation)}, \\ 
\min \mathcal{L}_{\text{anony}} &\supset P_s \hat{x} \ \bot \ P_s x && \text{(identity suppression)}. 
\end{align*}  

From an information-theoretic perspective, diffusion training maximizes mutual information with the original data:  
\begin{equation} 
\max \ I(\hat x;x), 
\end{equation}  
whereas anonymization minimizes mutual information with identity:  
\begin{equation} 
\min \ I(\hat x;c). 
\end{equation}  

Since $c$ is strongly correlated with $x$, faithful reconstruction inevitably risks identity leakage. Therefore, to enable effective anonymization within diffusion frameworks, this conflict must be explicitly resolved during both training and inference. The central challenge lies in the principled decoupling and targeted processing of the identity subspace $P_s$.

\subsubsection{Proof} 

\textbf{Assumptions.}  
We assume the following:  

i) Each image $X$ can be decomposed into two complementary and independent subspaces: an identity subspace $P_s$ and a utility subspace $P_u$, such that
\[
X = P_s X + P_u X.
\]  

ii) The identity label $C$ only depends on the identity-related features. That is, there exists a bijection function $f$ such that
\[
C = f(P_s X).
\]

Given an original face image $x$ and its anonymized counterpart $\hat{x}$, our goal is to show that minimizing the mutual information $I(\hat{x}; c)$ with $c=f(P_s x)$ is equivalent to enforcing statistical independence between the identity subspaces of $x$ and $\hat{x}$, i.e., $P_s \hat{x} \perp P_s x$.

\textbf{Step 1. Rewrite the mutual information.}  

Since $c = f(P_s x)$, we have
\begin{equation}
I(\hat{x}; c) = I(P_s \hat{x}, P_u \hat{x}; f(P_s x)).
\end{equation}
This follows because $\hat{x}$ and $(P_s \hat{x}, P_u \hat{x})$ are in one-to-one correspondence.

By the chain rule of mutual information,
\begin{equation}
\begin{aligned}
& I(P_s \hat{x}, P_u \hat{x}; f(P_s x)) \\
& = I(P_u \hat{x}; f(P_s x)) + I(P_s \hat{x}; f(P_s x) \mid P_u \hat{x}).
\end{aligned}
\end{equation}

Using Assumptions 1 and 2, the first term vanishes, i.e.,
\[
I(P_u \hat{x}; f(P_s x)) = 0,
\]
thus
\begin{equation}
I(\hat{x}; c) = I(P_s \hat{x}; f(P_s x) \mid P_u \hat{x}).
\end{equation}

If we additionally assume that $P_s \hat{x}$ and $P_u \hat{x}$ are statistically independent, the conditional term reduces to
\begin{equation}
I(\hat{x}; c) = I(P_s \hat{x}; f(P_s x)).
\end{equation}

\textbf{Step 2. Mutual information and entropy.}  

By the definition of mutual information,
\begin{equation}
I(P_s \hat{x}; f(P_s x)) = H(f(P_s x)) - H(f(P_s x) \mid P_s \hat{x}).
\end{equation}
Since $H(f(P_s x))$ only depends on the data distribution, minimizing $I(\hat{x}; c)$ is equivalent to maximizing the conditional entropy $H(f(P_s x) \mid P_s \hat{x})$, i.e., making $P_s \hat{x}$ carry as little identity information as possible.

\textbf{Step 3. The case of invertible $f$.}  

If $f$ is invertible, the invariance of mutual information under bijective transformations yields
\begin{equation}
I(P_s \hat{x}; f(P_s x)) = I(P_s \hat{x}; P_s x).
\end{equation}
Therefore, minimizing $I(\hat{x}; c)$ is equivalent to minimizing $I(P_s \hat{x}; P_s x)$.

In the extreme case where the mutual information vanishes,
\[
I(P_s \hat{x}; P_s x) = 0,
\]
we obtain statistical independence:
\[
P_s \hat{x} \perp P_s x.
\]

\textbf{Conclusion.}  
Under the above assumptions, minimizing $I(\hat{x}; c)$ is equivalent to enforcing that the generated identity subspace $P_s \hat{x}$ is statistically independent of the original identity subspace $P_s x$, thereby ensuring no identity information is leaked.

\section{Failure Cases and Limitations}

\begin{figure}[t]
    \centering
    \subfloat[]{
        \begin{minipage}{0.22\linewidth}
            \includegraphics[height=\linewidth]{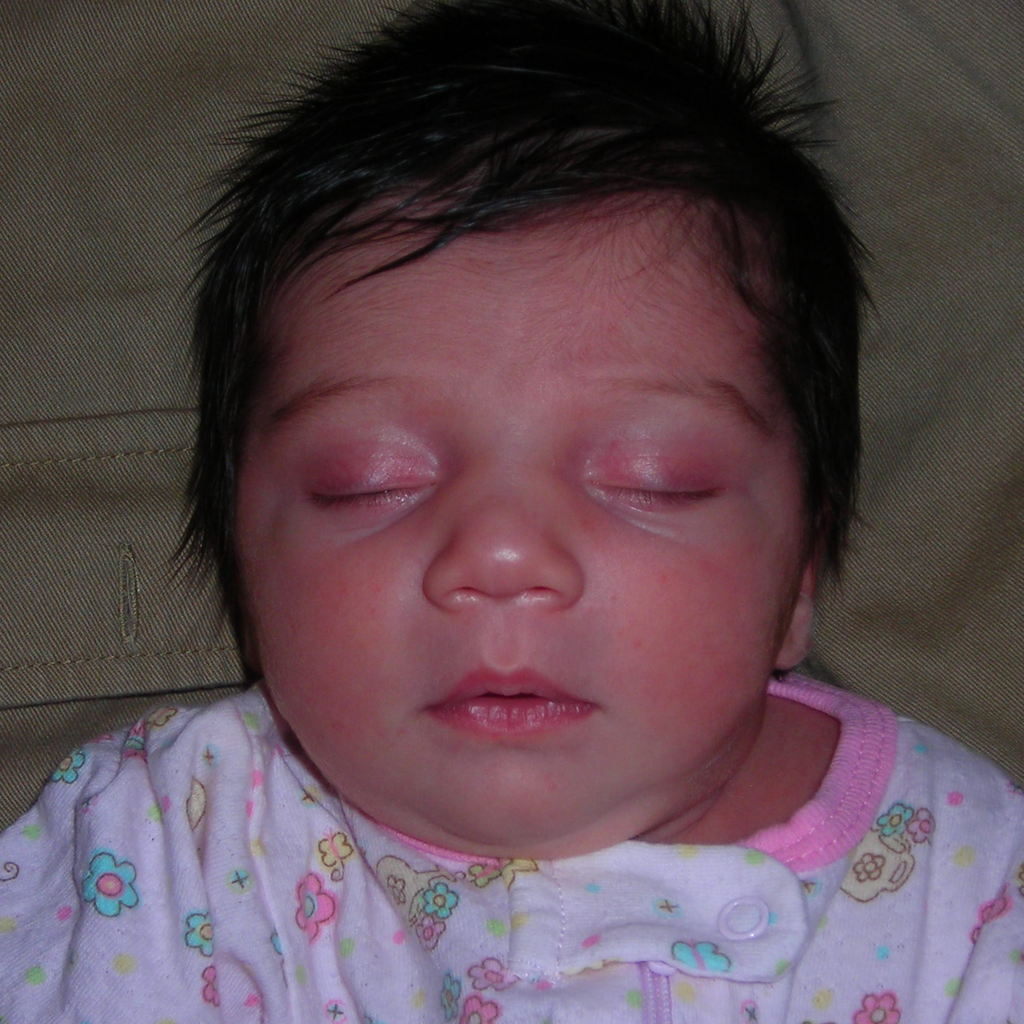} \\ \vspace{-8pt}
            \includegraphics[height=\linewidth]{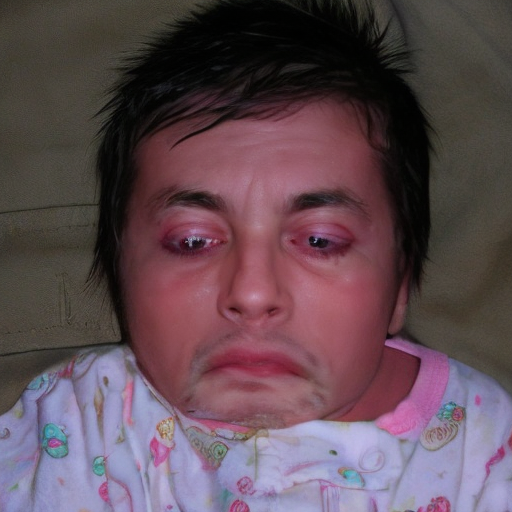} \\ \vspace{-12pt}
        \end{minipage}
    }
    \hspace{-6pt}
    \subfloat[]{
        \begin{minipage}{0.22\linewidth}
            \includegraphics[height=\linewidth]{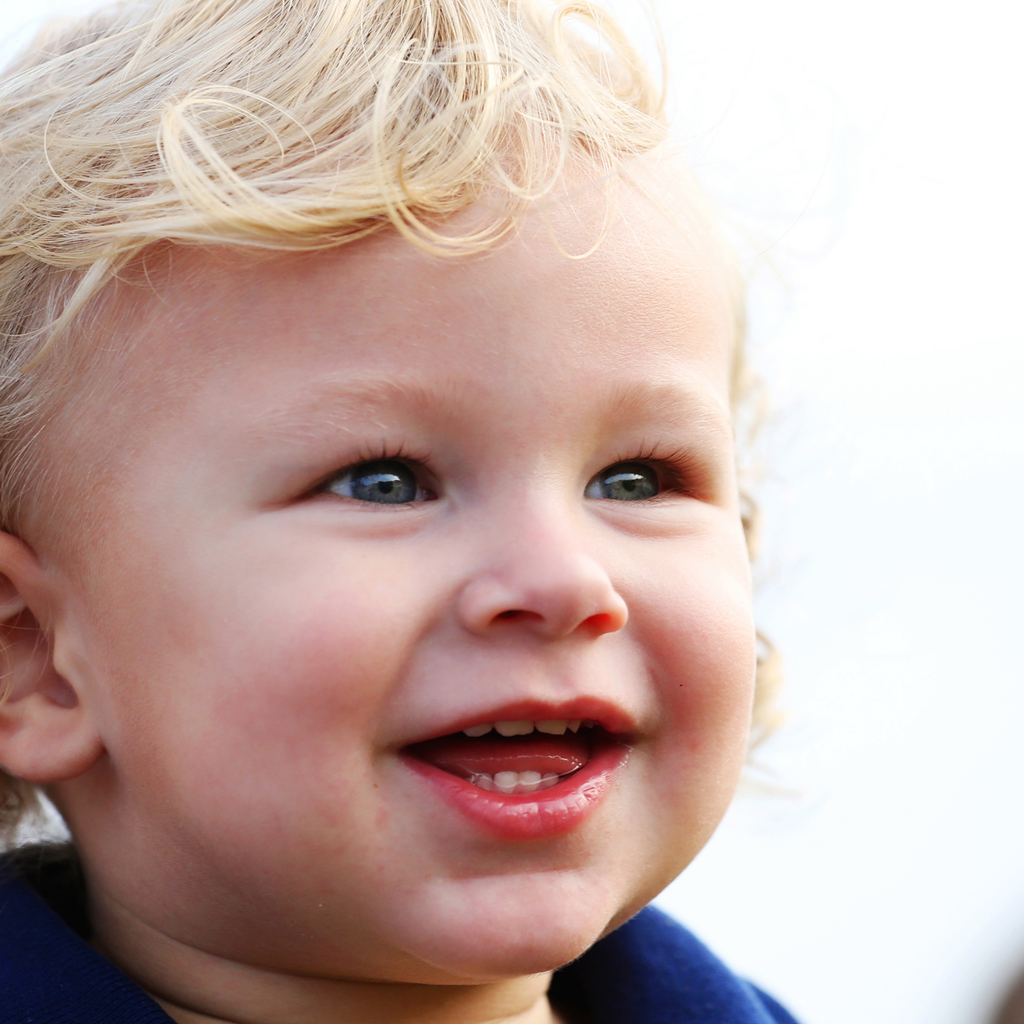} \\ \vspace{-8pt}
            \includegraphics[height=\linewidth]{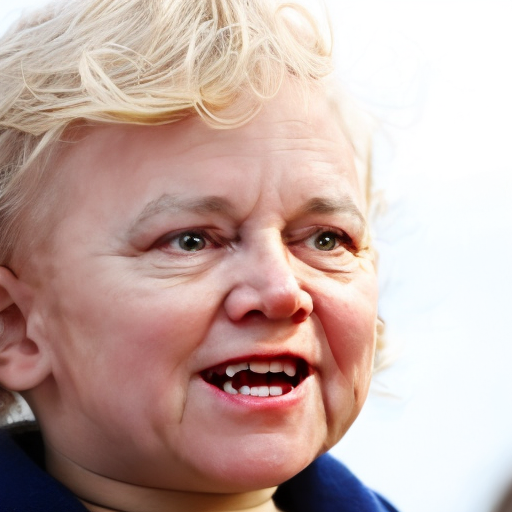} \\ \vspace{-12pt}
        \end{minipage}
    }
    \hspace{-6pt}
    \subfloat[]{
        \begin{minipage}{0.22\linewidth}
            \includegraphics[height=\linewidth]{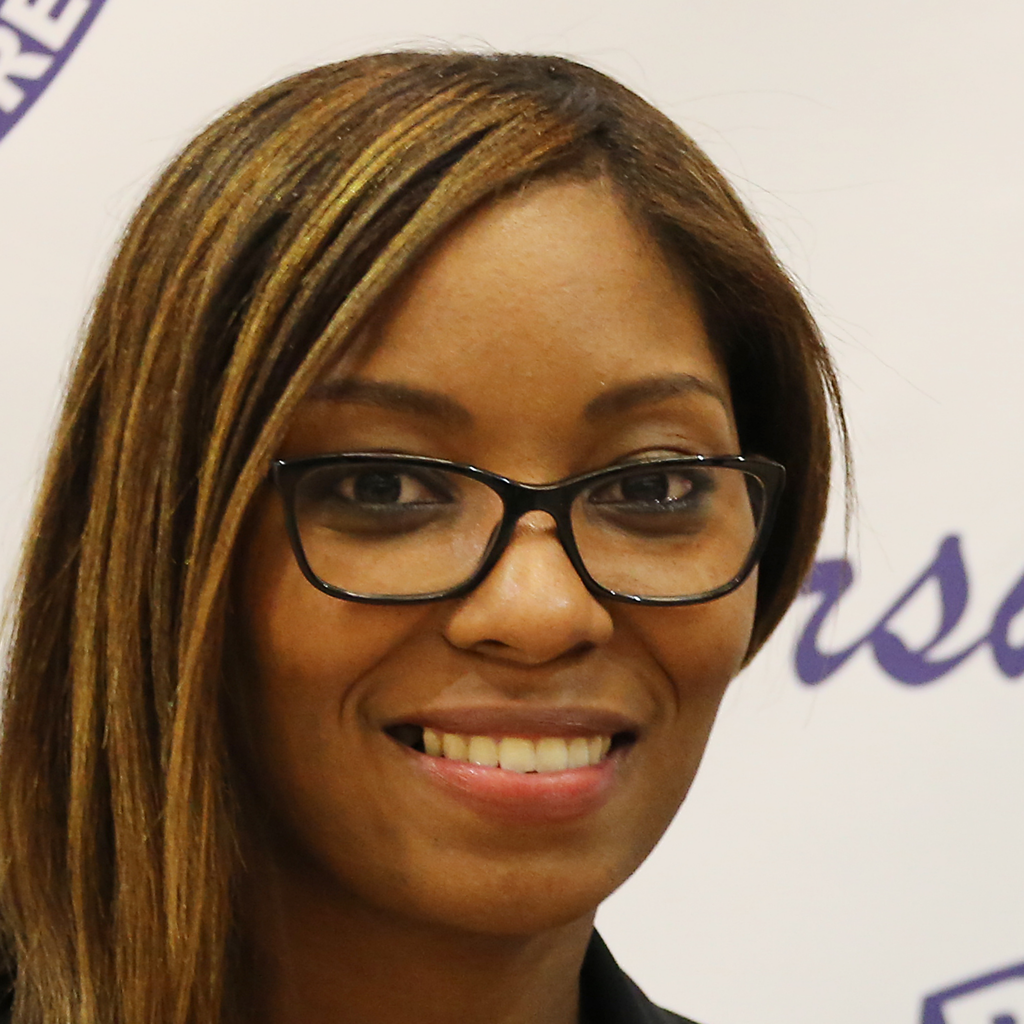} \\ \vspace{-8pt}
            \includegraphics[height=\linewidth]{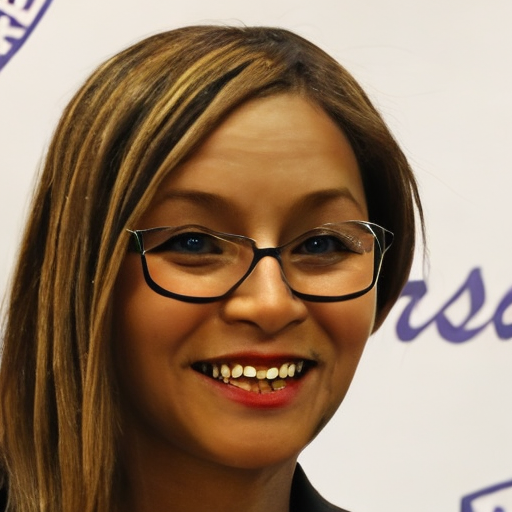} \\ \vspace{-12pt}
        \end{minipage}
    }
    \hspace{-6pt}
    \subfloat[]{
        \begin{minipage}{0.22\linewidth}
            \includegraphics[height=\linewidth]{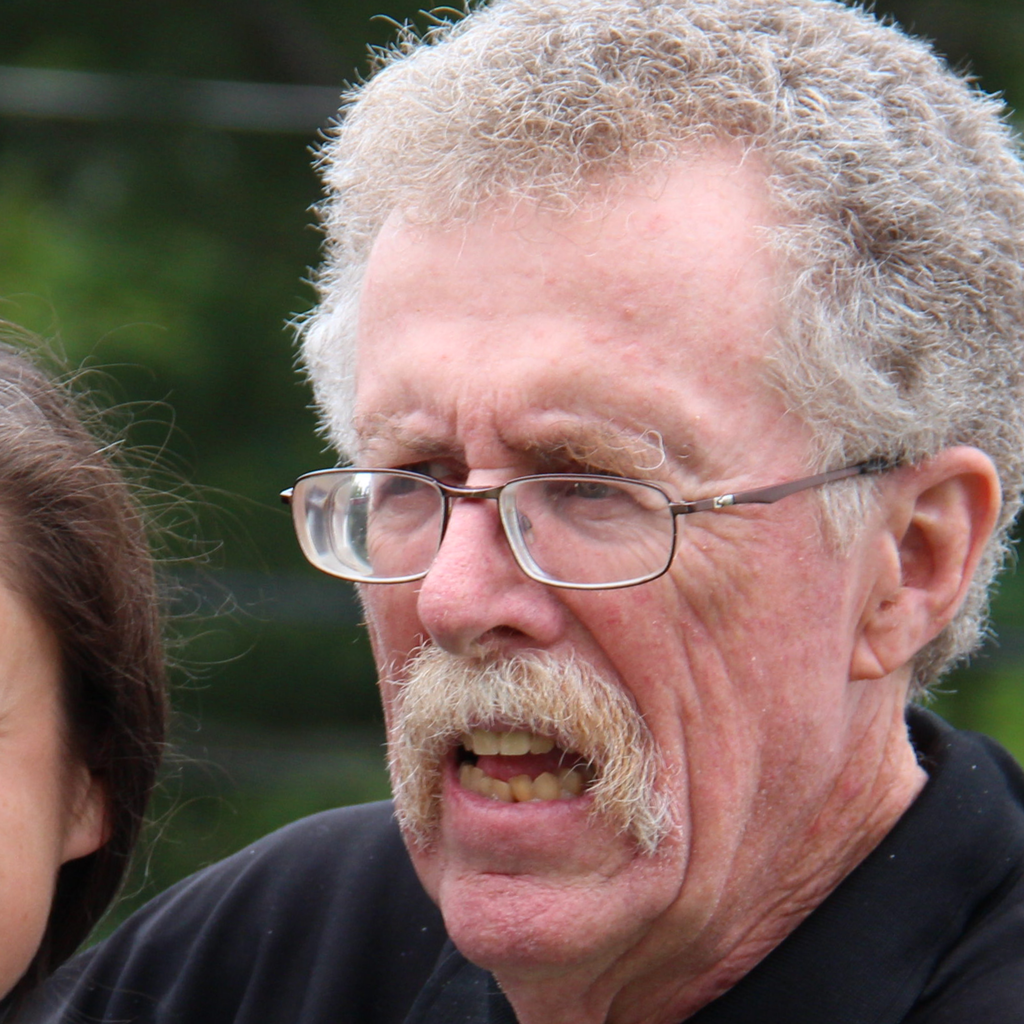} \\ \vspace{-8pt}
            \includegraphics[height=\linewidth]{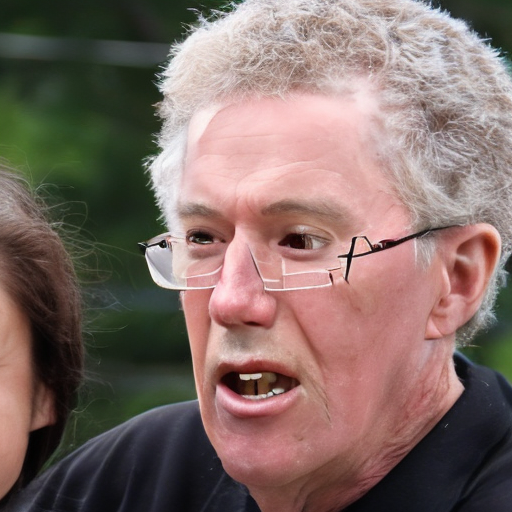} \\ \vspace{-12pt}
        \end{minipage}
    }
    \caption{
        Failure cases of our model.}
    \label{fig:limitation}
\end{figure}

\begin{table}[t!]
    \centering
    \caption{Computational complexity of ID\textsuperscript{2}Face.}
    \begin{tabular}{cccc}
    \toprule
                              & Params.   & Memory     & Latency \\ \midrule
    ID\textsuperscript{2}Face & 1,128.08M & 7,237.37MB & 4.55s   \\ \bottomrule
    \end{tabular}
    \label{tab:efficiency}
    \end{table}

While our method achieves strong anonymization performance in the majority of cases, several limitations remain, as illustrated in Fig.~\ref{fig:limitation}. First, when the input depicts baby faces, the anonymized outputs may appear less natural. This is largely attributable to the training data, which is dominated by adult faces, making it challenging for the model to learn realistic baby-specific features. Likewise, for images of individuals wearing glasses, the anonymized results may occasionally contain artifacts. This issue stems from our degradation strategy, which blurs glasses during preprocessing, combined with the limited representation of glasses in the dataset, reducing the model’s ability to reconstruct such cases convincingly.

In addition, while our approach benefits from the stability, controllability, and image quality inherent to diffusion models, these advantages come with computational trade-offs. Compared to GAN-based methods, diffusion models typically require more parameters and longer inference times, as shown in Table~\ref{tab:efficiency}. Although recent accelerated sampling techniques~\cite{lu2022dpm,sauer2024adversarial,yin2024one} can partially alleviate this issue, inference speed remains an area for further improvement.

\end{document}